\documentclass[11pt]{amsart}

\usepackage[foot]{amsaddr}

\usepackage[margin=1in]{geometry}
\usepackage{booktabs}

\usepackage{amsmath,amssymb,amsfonts,mathtools,color,xcolor,bm}
\usepackage[hidelinks]{hyperref}
\hypersetup{
    pdftitle={Clustering by a truncated estimator},
    pdfsubject={Spectral clustering for imbalanced isotropic Gaussian mixtures}
}
\usepackage{float}
\usepackage[style=nature]{biblatex}
\usepackage{wasysym}
\usepackage{amsmath,amssymb}
\usepackage{xcolor,tikz}
\usetikzlibrary{matrix,positioning,calc,arrows.meta,fit,decorations.pathreplacing}
\usetikzlibrary{patterns}

\newcommand{\mathbbm}[1]{\mathbf{#1}}

\DeclareMathOperator{\E}{\mathbb{E}}
\DeclareMathOperator{\Pb}{\mathbb{P}}

\DeclareMathOperator{\bbR}{\mathbb{R}}

\DeclareMathOperator{\bbP}{\mathbb{P}}

\DeclareMathOperator*{\diag}{diag}
\DeclareMathOperator*{\Span}{span}
\DeclareMathOperator*{\range}{range}
\DeclareMathOperator{\supp}{supp}

\DeclareMathOperator{\rank}{rank}

\DeclareMathOperator{\argmin}{argmin}

\DeclareMathOperator{\cov}{Cov}

\newcommand{\one}{\bm{1}}

\newcommand{\e}{\varepsilon}

\DeclarePairedDelimiter{\norm}{\lVert}{\rVert}
\DeclarePairedDelimiter{\ip}{\langle}{\rangle}

\newcommand{\bone}{\mathbf{1}}
\newcommand{\bzero}{\mathbf{0}}

\newcommand{\op}{\mathrm{op}}

\let\hat\widehat

\let\bar\overline

\newtheorem{theorem}{Theorem}[section]
\newtheorem{proposition}[theorem]{Proposition}

\newtheorem{lemma}[theorem]{Lemma}

\theoremstyle{remark}
\newtheorem{remark}[theorem]{Remark}

\theoremstyle{definition}
\newtheorem{definition}[theorem]{Definition}

\newcounter{algorithm}
\renewcommand{\thealgorithm}{\arabic{algorithm}}
\newenvironment{algorithm}[1]{%
    \refstepcounter{algorithm}%
    \par\medskip
    \noindent\begin{minipage}{\textwidth}
    \hrule\medskip
    \textbf{Algorithm \thealgorithm. #1}\par\medskip
}{%
    \medskip\hrule
    \end{minipage}
    \par\medskip
}

\begin{document}
	
	\title[Kernelised covariance matrices 
    ]{Kernelised covariance matrices, dimension reduction  and imbalanced spectral clustering*}
	\author{Xiaoyu Dong$^{\dagger,**}$
    }
\email{xdong@nus.edu.sg}
\author{Subhro Ghosh$^\dagger$}

\email{subhrowork@gmail.com}
\author{Soumendu Sundar Mukherjee$^\ddagger$}
\email{ssmukherjee@isical.ac.in}
\author{Himasish Talukdar$^{\ddagger,**}$
}

\email{talukdar.himasish@gmail.com}
\thanks{$^*$All the authors are listed alphabetically.}
\address{$^\dagger$National University of Singapore, Singapore}
\address{$^\ddagger$Indian Statistical Institute, Kolkata, India}
\thanks{$^{**}$X.D. and H.T. are corresponding authors. }

	\date{}

\begin{abstract}
The Gaussian Mixture Model ({\it abbrv.} GMM) is a classic generative model for clustered data, for which spectral clustering based on dimension reduction via PCA has been an effective framework. 
In this work, we investigate kernelised covariance matrices as an alternative mechanism for dimension reduction and spectral clustering, focusing on  truncation kernels that select pairs of data points with suitable separation constraints. The structure of these matrices is motivated by considerations from statistical mechanics of repulsive particle systems. Our approach is particularly relevant for the setting of imbalanced clusters, where the sizes of the largest and smallest clusters have a diverging ratio with the problem size, and existing PCA-based methods turn out to be of limited effectiveness.

We demonstrate efficient dimension reduction and clustering performance of these modified covariance matrices, substantiated by quantitative bounds. In particular, they achieve superior dimension reduction guarantees compared to PCA  for a wide range of parameters in the regime of imbalanced clusters. For spectral clustering, our approach exhibits improvements over PCA in the imbalanced setting,  in particular guaranteeing accurate clustering in a much larger regime of intra-cluster separations. Our theoretical guarantees reflected in a range of simulation experiments that exhibit these improvements for unbalanced clustering problems. Our analytical approach, based on kernel Laplacians, provide a toolbox as well as a relevant conceptual ingredient for clustering problems at large. We believe that our work will motivate the application of such kernelised covariance matrices for related classes of statistical problems, as well as spur new investigations in the theory of random matrices.
\end{abstract}

	\maketitle

\tableofcontents

    \section{Introduction}

\subsection{Clustering and the Gaussian mixture model}

The classical problem of clustering asks for a partition of a data set
into groups of similar points. Clustering is foundational to machine
learning and statistics, both classical and modern, with a wide array
of applications such as lossy compression \cite{lloyd,gray-neuhoff},
image classification in cryo-EM \cite{van-heel,zhao-singer},
inference of genetic population structure \cite{pritchard},
graphon estimation \cite{gao-lu-zhou}, and the discovery of cancer
subtypes from gene-expression data \cite{golub1999,perou2000}.

The Gaussian Mixture Model (GMM) is a typical model for the generation of clustered data, going back to
Pearson \cite{pearson}, where the components of the GMM are the clusters. Clustering studies that use GMM include direct estimation \cite{banfield-raftery,fraley-raftery}, likelihood and $k$-means  \cite{bishop,kulis-jordan}, learning theoretic aspects
\cite{dasgupta,vempala-wang,moitra-valiant}, and many more. To wit, one
observes $X_1,\dots,X_n$ in $\mathbb R^p$ with
\begin{equation}\label{eq:intro-model}
  X_i = \theta^*_{z^*_i} + \varepsilon_i, \quad \varepsilon_i \sim \mathcal N(0,\sigma^2 I_p),
\end{equation}
the centres $\theta^*_1,\dots,\theta^*_k$ and the labels $z^* \in [k]^n$ both unknown,
and one wants to estimate $z^*$. Difficulty is governed by the minimum separation
$\Delta_* := \min_{a \neq b} \|\theta^*_a - \theta^*_b\|_2$, and the question is how
small $\Delta_*$ can be before recovery fails.

Broadly speaking, three
families of methodological devices are in common use for the analysis of clustered data. The first approach entails direct minimisation of the so-called $k$-means objective, usually through iterative
algorithms \cite{lloyd,macqueen,hartigan-wong}.
The second approach concerns likelihood fitting of a
mixture, usually by EM \cite{dempster,mclachlan-peel}. Finally, the third and perhaps the theoretically best substantiated approach entails spectral methods, which
project the data onto a subspace read off from the spectrum of a data matrix and cluster
the projected data \cite{vempala-wang,LZZ2021,zhang-zhou2024,donath-hoffman,fiedler,shi-malik,ng-jordan-weiss,vonluxburg}.

In \cite{LZZ2021}, L\"offler, Zhang, and Zhou  showed
that plain spectral clustering, PCA followed by $k$-means,  attains the  minimax optimal
misclustering rate, with no iterative refinement needed. Abbe, Fan, and Wang
\cite{abbe-fan-wang} developed an $\ell_p$ analysis of PCA and spectral
clustering. Later Zhang and Zhou \cite{zhang-zhou2024} obtained
spectral guarantees under weaker separation conditions and extended
the analysis to sub-Gaussian mixtures. Lu and Zhou \cite{lu-zhou} gave
matching guarantees for Lloyd's algorithm from a good initialisation. Exact recovery and
its thresholds were studied by Giraud and Verzelen \cite{giraud-verzelen} and by Ndaoud
\cite{ndaoud}.
Chen and Zhang \cite{chen-zhang} studied anisotropic Gaussian mixtures
using a version of Lloyd's algorithm that estimates and uses the
covariance matrices.

\subsection{Segmenting data with imbalanced clusters}

Data with an imbalanced clustering structure, e.g., the size of one cluster dominates,  is of growing importance in a host of real-world applications. Principal instances include medical and biological data, where the healthy patients form one
overwhelming group and the sub-population of interest is small
\cite{he-garcia,chandola,zong}, and rare cell populations must
be distinguished from more abundant cell types \cite{grun2015}. In addition, fraudulent transactions run to a few per thousand
\cite{bolton-hand}, communities in a large network are small relative to the graph
\cite{leskovec}, and species abundances and word frequencies are heavy-tailed
\cite{fisher-corbet-williams,newman}. 

In fact, average misclustering error is then a poor yardstick in the setting of imbalanced clusters,
since it can be small while a minority cluster is lost entirely. In this work, we will therefore also investigate the worst rate over clusters to provide a more comprehensive picture.

Much of the existing theoretical literature, particularly in the context of the optimality of known rates, applies to the setting of {\it balanced clusters}. Roughly speaking, this entails that relative sizes of the biggest and smallest clusters stays bounded (e.g., in terms of their ratio) as the problem sizes grows to infinity. In \cite{zhang-zhou2024} (and also other works such as \cite{LZZ2021}), the authors quantify the smallest cluster size by
\[
  \beta:=\frac{k n_{\min}}{n}\in(0,1],
  \quad n_{\min}:=\min_{a\in[k]}n_a
\]
and the guarantees established therein are minimax optimal when the separation satisfy condition $(\Delta/\sigma) \gtrsim\frac{1}{\sqrt{\beta}}$, which is most meaningful for fixed $\beta$. In particular, if $\beta \to 0$ rapidly (e.g., $\beta \asymp \frac{1}{\sqrt{n}}$), a natural setting for data with {\it imbalanced clusters}, the separation condition 
becomes very restrictive. 
Therefore, these existing guarantees are of limited effectiveness in addressing clustered data with imbalanced clusters, which calls forth new ways of approaching the clustering problem. This is the problem we principally aim to address in this work. More precisely, our truncation based clustering method works for both balanced clusters and imbalanced clusters, and in particular, our method significantly improves the separation requirements when one cluster dominates and number of observations is much larger than the dimension (see Remark \ref{wb:rem:clustering-comparison} for detailed discussion).

\subsection{Kernelised covariance matrices and statistical physics}
\subsubsection{Clustering and Gaussian Determinantal Processes}
Our approach to clustering passes through kernelisations of covariance matrices, motivated by recent works leveraging ideas from statistical mechanics. In particular, Ghosh and Rigollet \cite{ghosh-rigollet-pnas} introduced a family of truncated covariance-type matrices in the course of their investigations on Gaussian Determinantal Processes (GDP) as a parametric model for repulsion in point sets and directionality phenomena in datasets. More specifically, Ghosh and Rigollet \cite{ghosh-rigollet-pnas} proposes a matrix-valued test statistic that involves the truncated covariance matrix
\begin{equation}\label{eq:intro-gdp}
  \widehat\Sigma_{\rm GDP} \;\propto\; \sum_{i,j} \mathbf 1\{\|X_i - X_j\|_2 < r\}\,
  (X_i - X_j)(X_i - X_j)^\top
\end{equation}
for a cutoff $r$.  The determinantal model was an ansatz there, in the spirit of the Wigner surmise, rather than a hypothesis about the  nature of the data; and on benchmark clustering applications such as  Fisher's iris data and the Wisconsin breast cancer data the resulting spectral subspaces exhibited strong performance in distinguishing the known clusters. More generally, it was demonstrated in \cite{ghosh-rigollet-pnas} that the principal eigen-directions of $aI-b\widehat\Sigma_{\rm GDP}$, with appropriate coefficients $a$ and $b$, maybe deployed as a device for uncovering statistically significant directions in a dataset. For instance, in the simple setting of a two-cluster dataset, the canonical direction of interest is the line joining the centers of the two clusters.

In this work, we will work with the matrix-valued statistic that is of the form \eqref{eq:intro-gdp}, with the inequality turned around:
\begin{equation}\label{eq:intro-ours}
  \sum_{i,j} \mathbf 1\{\|X_i - X_j\|_2^2 \ge u_0\}\,(X_i - X_j)(X_i - X_j)^\top ,
\end{equation}
with a threshold $u_0$ fixed suitably.

\
For contrast, in the setting of spectral clustering, we recall that the sample covariance matrix has the representation
\begin{equation}\label{eq:intro-gram}
  \frac{1}{n}\sum_i (X_i - \bar{X})(X_i - \bar{X})^\top \;=\; \frac{1}{2n^2} \sum_{i,j=1}^n (X_i - X_j)(X_i - X_j)^\top .
\end{equation}
where $\bar X:=n^{-1}\sum_{i=1}^nX_i$.
Spectral clustering proceeds with projections on to the leading eigenspaces of \eqref{eq:intro-gram}. Now fix $i$
and $j$ in the same cluster. Then $X_i - X_j = \varepsilon_i - \varepsilon_j$, so the
corresponding summand carries no information about the centres at all. With clusters of
comparable size such pairs are a constant fraction of the sum. However, with one dominant cluster
they are almost all of it, outlining a central difficulty that would emerge in the setting of imbalanced clustering.
It is therefore of interest to focus on the pairs that straddle two clusters and drop the rest. This provides a natural heuristic in support of the structure of the matrix-valued statistic \eqref{eq:intro-ours} leveraged in this work. 

\subsubsection{The Widom-Rowlinson Model as a mechanism for cluster formation}
A direct view towards clustering phenomena is given by the related Widom-Rowlinson model \cite{widom-rowlinson} of repulsive particle systems from statistical mechanics. To briefly understand this model, let us consider two species of particles occupying a box, with unlike particles
repelling and like particles not interacting at all. Ruelle \cite{ruelle} showed that at high activity it
separates into phases. Confining the particles in a potential,  one expects the equilibrium state to be a
pair of blobs, one per species, pushed apart by the inter-species repulsion and held in by the potential.

While the Widom--Rowlinson model does provide a simple physical model for clustering phenomena, its direct applicability is constrained by the fact that its properties are not amenable to simple formulae and closed form analysis \cite{ruelle-book,chayes-chayes-kotecky,georgii-haggstrom}. This is, in fact,
typical of interacting particle systems, and  determinantal processes are the exception:
every $k$-point function is an explicit determinant, the pair correlation is in closed
form, linear statistics have exact first and second moments, existence follows from a
spectral condition on the kernel \cite{macchi,hkpv}, and there is an exact sampling
algorithm. A parametric subfamily like the Gaussian Determinantal Processes carries an interpretable matrix-valued
parameter, which is what made \eqref{eq:intro-gdp} usable. It may be emphasized that the analysis in this paper is
carried out entirely under the Gaussian mixture \eqref{eq:intro-model}, and the statistical mechanical considerations are purely for motivational purposes, along the lines of \cite{ghosh-rigollet-pnas}.

\subsubsection{Kernelised and truncated covariance matrices}
In fact, the statistic \eqref{eq:intro-ours} admits a natural generalization to a class of kernelised covariance matrices given by 
\begin{equation}\label{eq:intro-kernel}
  \sum_{i,j} \Psi(X_i,X_j ; r)\,(X_i - X_j)(X_i - X_j)^\top ,
\end{equation}
where $\Psi$ is an interaction kernel between the data points $X_i$ and $X_j$, modulated by the parameter $r$. The kernelised statistic \eqref{eq:intro-kernel} can be further specialised to the translation invariant form 
\begin{equation}\label{eq:intro-kernel-inv}
  \sum_{i,j} \mathbf \Psi(X_i-X_j ; r)\,(X_i - X_j)(X_i - X_j)^\top ,
\end{equation}
which includes the important Gaussian kernel as a special case, accorded by the choice of $\Psi(X_i-X_j ; r) =  1 - \exp\left(-\|X_i-X_j\|^2/r^2 \right)$. The companion paper \cite{ghosh_mukherjee_talukdar_2024_bulk} analyses the spectral properties of kernelised covariance matrices as above, establishing novel limiting spectral distributions and generalisations of the Marcenko-Pastur law.

It is indeed of interest to consider such general kernel classes for particular applications, such as clustering. For our purposes of imbalanced clustering, however, it turns out that the far-neighbour indicator kernel as in \eqref{eq:intro-ours} is a more appropriate choice than other possibilities (such as the Gaussian kernel). For a discussion of these matters, we refer the reader to Section \ref{sec: thresholdchoice}.

	\section{Setup and background}
    \subsection{Notation}
	For an integer $m\ge 1$, write $[m]:=\{1,2,\dots,m\}$. We use the notation $\log_+(x)$ for $\max\{\log(x),0\}$. For a vector $v\in\mathbb{R}^m$, we write $v_i$ for its $i$th entry and $\|v\|_2$ for its Euclidean norm. For a matrix $M\in\mathbb{R}^{a\times b}$, we write $M_{i,\cdot}$ for the $i$th row, $M_{\cdot,j}$ for the $j$th column, $\|M\|_{\mathrm{op}}$ for the operator norm, and $\|M\|_{\mathrm{F}}$ for the Frobenius norm. The $n\times n$ identity matrix is denoted by $I_n$. For a subset $S\subseteq [n]$, define its indicator vector $\one_S\in\bbR^n$ by $(\one_S)_i=1$ if $i\in S$ and $(\one_S)_i=0$ otherwise. The all-zeros and all-ones vectors in $\mathbb{R}^m$ are denoted by $\bzero_m$ and $\mathbf{1}_{[m]}$ respectively. Further define $J_{m \times m'}:= \bone_{[m]} \bone_{[m']}^\top$, the $m\times m'$ (possibly rectangular) matrix consisting only of $1$'s. For scalars $y_1,\dots,y_m$, $\mathrm{diag}(y_1,\dots,y_m)$ is the diagonal matrix with diagonal entries $(y_1,\dots,y_m)$. For a symmetric positive semidefinite matrix $A\succeq 0$, the notation $A^{1/2}$ refers to the unique symmetric positive semidefinite square root. For an subspace $\mathcal U$ in $\bbR^n$, we denote the orthogonal projection onto $\mathcal U$ by $\Pi_{\mathcal U}$.
    
    The indicator of an event $E$ is written as $\mathbbm{1}_{\{E\}}$. For $x>0$, write $\log_+x:=\max\{0,\log x\}$.
	The symbols $\chi_p^2$ and $\chi_p^2(\lambda)$ denote, respectively,
	the central chi-square distribution with $p$ degrees of freedom and
	the noncentral chi-square distribution with $p$ degrees of freedom and
	noncentrality parameter $\lambda\ge0$.

Throughout the paper, the symbols $c_{\ref{wb:thm:dimension-reduction}.1}$,
$c_{\ref{wb:thm:dimension-reduction}.2}$,
$c_1,c_2,\ldots,$ $C$ denote positive absolute constants.

    \subsection{Gaussian Mixture Model}
	Fix the number of clusters $k\ge 2$. Let
	\[
	\theta_1^\ast,\dots,\theta_k^\ast \in \mathbb{R}^p
	\]
	be the (unknown) cluster centers, and let
	\[
	z^\ast=(z_1^\ast,\dots,z_n^\ast)\in [k]^n
	\]
	be the (unknown) cluster assignment vector. For each $l\in[k]$, define the true cluster index set and its size by
	\[
	B_l := \{ i\in[n] : z_i^\ast = l \},\quad n_l := |B_l|.
	\]
	Throughout this paper, we assume that every cluster is nonempty, i.e.
	\[
	n_l\ge 1\quad\text{for all }l\in[k].
	\]
    
    Let
    \begin{equation}\label{wb:eq:onebig}
        n_{\max}:=\max_{1\le a\le k}n_a,
        \text{ and }
        m:=n-n_{\max}.
    \end{equation}
In other words, all clusters outside a largest cluster contain a total of $m$ observations. 
Without loss of generality, we may relabel the clusters so that
\begin{equation}\label{wb:eq:onebig-convention}
        n_1=n_{\max}=n-m,
        \quad
        \sum_{a=2}^k n_a=m.
\end{equation}
Since every cluster is nonempty, \eqref{wb:eq:onebig} also implies $m\ge k-1$.

We also use the notation
\begin{equation}\label{wb:eq:nmin}
        n_{\min}:=\min_{1\le a\le k} n_a .
\end{equation}
for the size of the smallest cluster.
	
	We observe independent random vectors $X_1,\dots,X_n\in\mathbb{R}^p$ generated according to an isotropic Gaussian mixture model:
	\begin{equation}
		X_i = \theta^\ast_{z_i^\ast} + \varepsilon_i,\quad \varepsilon_i \sim \mathcal{N}(0,\sigma^2 I_p),\quad i\in[n],
		\label{eq:gmm}
	\end{equation}
	where $\{\varepsilon_i\}_{i=1}^n$ are independent. The goal of clustering is to estimate $z^\ast$ from the observed data $\{X_i\}_{i=1}^n$.
	
	It is convenient to collect the data into a matrix
	\[
	X := \bigl[ X_1,\dots,X_n \bigr] \in \mathbb{R}^{p\times n}.
	\]
	Define the signal (cluster center) matrix
	\[
	P := \bigl[ \theta^\ast_{z_1^\ast},\dots,\theta^\ast_{z_n^\ast} \bigr] \in \mathbb{R}^{p\times n}
	\]
	and the noise matrix
	\[
	E := \bigl[ \varepsilon_1,\dots,\varepsilon_n \bigr] \in \mathbb{R}^{p\times n}.
	\]
	Then the model \eqref{eq:gmm} can be written compactly as
	\[
	X = P + E.
	\]
	More formally, we have the following definition.

	\begin{definition}
		Let $p\ge 1$ and $k\ge 2$. Let
		\[
		\Theta^*=(\theta_1^\ast,\dots,\theta_k^\ast) \in \mathbb{R}^{p \times k}
		\]
		and let $n \ge 2$. Let
		\[
		z^\ast=(z_1^\ast,\dots,z_n^\ast)\in [k]^n
		\]
		be a cluster assignment vector such that each label in $[k]$ appears at least once. Let $\sigma>0$, and let the random vectors $X_1,\ldots,X_n$ satisfy
		\begin{equation}
			X_i = \theta^\ast_{z_i^\ast} + \varepsilon_i,\quad \varepsilon_i \sim \mathcal{N}(0,\sigma^2 I_p),\quad i\in[n],
			\label{eq:gmm-def}
		\end{equation}
		where $\{\varepsilon_i\}_{i=1}^n$ are independent. Define
		\[
		X := \bigl[ X_1,\dots,X_n \bigr] \in \mathbb{R}^{p\times n},
		\]
		and define
		\[
		P := \bigl[ \theta^\ast_{z_1^\ast},\dots,\theta^\ast_{z_n^\ast} \bigr] \in \mathbb{R}^{p\times n}
		\]
		and
		\[
		E := \bigl[ \varepsilon_1,\dots,\varepsilon_n \bigr] \in \mathbb{R}^{p\times n}.
		\]
		Then $(X,P,E)$ is called a Gaussian mixture observation model for the true cluster configuration $(\Theta^*,z^*)$ and noise level $\sigma$.
	\end{definition}
	
	The difficulty of clustering depends significantly on the geometric separation of the centers
	\[
	\min_{\substack{j,\ell\in[k]\\ j\neq \ell}} \bigl\| \theta_j^\ast - \theta_\ell^\ast \bigr\|_2.
	\]
	Intuitively, when $\bigl\| \theta_j^\ast - \theta_\ell^\ast \bigr\|_2 \approx0$, two clusters have almost identical centers and can hardly be distinguished from samples, while as $\min_{\substack{j,\ell\in[k], j\neq \ell}} \bigl\| \theta_j^\ast - \theta_\ell^\ast \bigr\|_2$ gets larger, the clustering problem becomes easier. 
    More formally, by norm estimate of high dimensional vector (see Lemma~\ref{wb:thm:gaussian-vector-norm}), we know that $\|\varepsilon_i\|_2$ is concentrated around $\sigma\sqrt p$.  Therefore in the strong-separation regime when
\[
        \Delta\gg \sigma\sqrt p,
\]
the distance between centers dominates the norms of random noise vectors, and individual observations are geometrically well separated. As a result, this regime is comparatively easy for dimension reduction and clustering.  The more interesting and harder regime  is the weak-separation regime
\begin{equation}\label{eq:weak-separation}
        \Delta\ll \sigma\sqrt p,
        \quad\text{equivalently}\quad
        \Delta^2/\sigma^2\ll p,
\end{equation}
where the separation between two centers is smaller than the typical noise length of one observation. So, in this regime, pairwise distances still contain aggregate information.

  In practice, it may not be possible to obtain the exact minimum separation $\min_{\substack{j,\ell\in[k]\\ j\neq \ell}} \bigl\| \theta_j^\ast - \theta_\ell^\ast \bigr\|_2$. But it is reasonable to posit we have access to a lower bound 
\begin{equation}\label{eq:min_sep}
	0< \Delta \le \min_{\substack{j,\ell\in[k]\\ j\neq \ell}} \bigl\| \theta_j^\ast - \theta_\ell^\ast \bigr\|_2.
\end{equation}
Our dimension reduction and clustering algorithm only uses this lower bound. And similar assumptions are made in previous studies (see, e.g., \cite{BelkinSinha2010,BubeckMeilaVonLuxburg2012,SrivastavaSarkarHanasusanto2023}). Indeed, in practice, the separation threshold may be
partly thought of as a choice made the practitioner, based on domain knowledge or
the purpose of the analysis, since data from two clusters with sufficiently close centers
may not be reliably distinguished from a single cluster using
the available data, by the information theoretic lower bounds~(see, e.g., \cite{Tsybakov2009,LaurentMarteauMaugisRabusseau2016,lu-zhou,zhang-zhou2024}).

   Let $h:=1+\log_+(n/(p+m))$ and $r_\ast:=\min\{k-1,p\}$. Our dimension reduction result assumes $\Delta^2/\sigma^2\gtrsim h+\sqrt{ph}$, and the clustering result assumes the analogous condition with $h$ replaced by $h+r_\ast(p+m+\log n)/n_{\min}$.  In practical application it is reasonable to assume $h$ is $o(p)$, so our method is designed to work in the harder weak-separation regime.

    \subsection{Classical PCA based dimension reduction and clustering}
    \label{wb:subsec:classical-spectral-pipeline}

    We first recall the centered PCA based dimension-reduction and clustering methods as a starting point.  Standard PCA based clustering (e.g., L\"offler, Zhang, and Zhou~\cite[Algorithm~1]{LZZ2021}) directly use the uncentered data matrix $X$. But to motivate our method, here we center the data to obtain  translation invariant data.  More precisely, let $\one_{[n]}$ be the all ones vector and let
    \[
        H_n:=I_n-\frac1n\one_{[n]}\one_{[n]}^\top
    \]
    be transformation that centers the data, i.e., move the average of all samples to the zero vector. And define the centered data, signals, and noise
    \[
        X_c:=XH_n,
        \quad
        P_c:=PH_n,
        \quad
        E_c:=EH_n.
    \]
    Then $X_c=P_c+E_c$.  Write a singular value decomposition as
    \[
        X_c
        =
        \sum_{j=1}^{\min\{p,n\}}
        \widehat\sigma^{\rm PCA}_j\,
        \widehat u^{\rm PCA}_j
        (\widehat v^{\rm PCA}_j)^\top,
        \quad
        \widehat\sigma^{\rm PCA}_1
        \ge \widehat\sigma^{\rm PCA}_2
        \ge\cdots\ge0.
    \]
    For $1\le r\le\min\{p,n\}$, the classical PCA projects the data onto the leading $r$-dimensional subspace spanned by
    \[
        \widehat U_{{\rm PCA},r}
        :=
        \bigl[
            \widehat u^{\rm PCA}_1,\ldots,
            \widehat u^{\rm PCA}_r
        \bigr].
    \]
    Equivalently, we can also write the eigenvalue decomposition
    \[
         \sum_{i=1}^n (X_i - \bar{X})(X_i - \bar{X})^\top= X_cX_c^\top
        =
        \sum_{j=1}^{\min\{p,n\}}
        (\widehat\sigma^{\rm PCA}_j)^2
        \widehat u^{\rm PCA}_j
        (\widehat u^{\rm PCA}_j)^\top,
    \]
    where $\bar{X}:=\frac{1}{n}\sum_{i=1}^n X_i$ is the sample mean.
    So, $\widehat U_{{\rm PCA},r}$ may be obtained by taking the eigenspaces corresponding to the $r$ largest eigenvalues of $X_cX_c^\top$.  In the special case in which the observations are already sample centered, $X\one_{[n]}=0$, one has $X_c=X$, and the same subspace is obtained directly from the eigendecomposition of $XX^\top$.  
    A common choice of $r$ is
    \[
        r_0:=\rank(P_c),
    \]
    and let $U_0\in\mathbb R^{p\times r_0}$ be any orthonormal basis for $\operatorname{col}(P_c)$.  When $r_0\ge1$, the natural measurement of estimation error is the distance
    \[
        \left\|
            \widehat U_{{\rm PCA},r_0}
            \widehat U_{{\rm PCA},r_0}^\top
            -
            U_0U_0^\top
        \right\|_{\rm op}.
    \]
    After dimension reduction, the classical PCA based spectral clustering applies $k$-means to the columns of
    \[
        \widehat P_{\rm PCA}
        :=
        \widehat U_{{\rm PCA},r}^\top X_c
        \in\mathbb R^{r\times n}.
    \]
    The $i$th column $\widehat P_{{\rm PCA},\cdot,i}=\widehat U_{{\rm PCA},r}^\top(X_c)_{\cdot,i}\in\mathbb R^r$ contains an observation of reduce dimension.
    Indeed, if $\widehat V_{{\rm PCA},r}$ and $\widehat\Sigma_{{\rm PCA},r}$ denote the corresponding right singular-vector and singular-value matrices, then
    \[
        \widehat P_{\rm PCA}
        =
        \widehat\Sigma_{{\rm PCA},r}
        \widehat V_{{\rm PCA},r}^\top.
    \]
    In other words, the classical PCA based approach projects the observations onto an estimated low dimensional subspace, and then do clustering using data with reduced dimension.

\section{Key ingredients and approach}

    The aim of our approach is to improve the estimation of the low dimensional subspace, leading to improvements in clustering accuracy in the imbalanced setting. To explain the motivation, we start from the following representation
    \begin{equation}
    \label{wb:eq:raw-centered-pairwise-covariance}
        X_cX_c^\top
        =
        XH_nX^\top
        =
        \frac1{2n}
        \sum_{i,j=1}^n
        (X_i-X_j)(X_i-X_j)^\top.
    \end{equation}
    The sum in \eqref{wb:eq:raw-centered-pairwise-covariance} treats all pairs equally.  For a within-cluster pair $z_i^*=z_j^*$, however,
    \[
        X_i-X_j=\varepsilon_i-\varepsilon_j,
    \]
    so the corresponding term contains no between-center signal.

      This observation motivates a reduced-noise version of $X_cX_c^\top$: keep the pairwise terms that carry information about differences between cluster centers while removing within-cluster terms.  The objective is to construct an estimator $\widehat U_{r_0}$ that is closer to $U_0$. This idea leads to the kernelised covariance matrix
    \begin{equation*}
  \frac{1}{2n}\sum_{i,j} \mathbf \Psi(X_i-X_j ; r)\,(X_i - X_j)(X_i - X_j)^\top 
\end{equation*}
discussed in the introduction, which would ideally shrink the pure noise terms
\[
        X_i-X_j=\varepsilon_i-\varepsilon_j
    \]
However, this kernelised covariance matrix has dimension $p \times p$, so it does not include sufficient information to determine cluster assignment for $n$ data points. To overcome this, we observe that we can write
 \begin{equation*}
  \frac{1}{2n}\sum_{i,j} \mathbf \Psi(X_i-X_j ; r)\,(X_i - X_j)(X_i - X_j)^\top = X\frac{\widehat L}{n}X^\top
\end{equation*}
for a suitable Laplacian matrix ${\widehat L}$ (defined formally below in Section \ref{sec: truncation}).  This suggests the transformed data
\[Y=X\left(\frac{\widehat L}{n}\right)^{1/2} \in \mathbb{R}^{p \times n}\]
which has the correct dimensions and also its covariance matrix 
 \begin{equation}
 \label{eq:k1}
 Y Y^\top = X\frac{\widehat L}{n}X^\top= \frac{1}{2n}\sum_{i,j} \mathbf \Psi(X_i-X_j ; r)\,(X_i - X_j)(X_i - X_j)^\top 
\end{equation}is kernelised by design.

	\subsection{A truncated covariance matrix and truncation matrices}\label{sec: truncation} In this section, we define suitable ${\widehat L}$ such that
    \eqref{eq:k1} holds.
    It turns out that an effective  choice (for the purpose of dimension reduction and clustering)  of the kernel $\Psi$ in \eqref{eq:k1} is a truncation kernel. This leads to the consideration of truncated covariance matrices and the terminology of truncation matrices used in this section.
    The suitability of the truncation kernel, cast against other possibilities, 
    {is taken up for discussion in Section \ref{sec: thresholdchoice}}.
	\subsubsection{Data truncation matrices}
\begin{definition}[Degree and Laplacian operators]\label{def:degree-laplacian-operators}
    Let $q\ge 1$ and let $H\in\mathbb{R}^{q\times q}$. Define the degree operator by
    \[
        \mathcal{D}(H):=\diag(H\one_{[q]})
    \]
    and the Laplacian operator by
    \[
        \mathcal{L}(H):=\mathcal{D}(H)-H
        =\diag(H\one_{[q]})-H.
    \]
\end{definition}

\begin{remark}
        If $H$ is symmetric with nonnegative entries, then $\mathcal{L}(H)$ is symmetric positive semidefinite. Also its value is  not affected by the diagonal entries of $H$. 
\end{remark}

	\begin{definition}[Data truncation matrices]\label{def:datatrunc}
		Let $X\in\bbR^{p\times n}$ be an observed data matrix. Let $\phi: [0,+\infty) \to [0,1]$ be a Borel measurable function. Let
		\[
		\widehat A_{ij} :=
		\phi(\|X_i-X_j\|_2^2)
		\]
		be the kernel matrix. Let
		\[
		\widehat D := \mathcal{D}(\widehat A)
		\]
		be the data degree matrix. Let
		\[
		\widehat L := \mathcal{L}(\widehat A)
		\]
		be the data Laplacian matrix. Then $(\widehat A,\widehat D, \widehat L)$ is called the data truncation matrices for the data $X$ using threshold function $\phi$.
	\end{definition}

With this definition, we have
 \begin{equation*}
  \frac{1}{2n}\sum_{i,j} \phi(\|X_i-X_j\|_2^2)\,(X_i - X_j)(X_i - X_j)^\top = X\frac{\widehat L}{n}X^\top
\end{equation*}
which is exactly of the form \eqref{eq:k1}.

	\subsubsection{Ideal truncation matrices}
	To understand the expected behavior of the data truncation matrices, we consider the ideal truncation matrices.
	
	\begin{definition}[Ideal truncation matrices]\label{def:ideal-truncation-model}
		Let $p\ge 1$ and $k\ge 2$ (representing the number of clusters). Let
		\[
		\Theta^*=(\theta_1^\ast,\dots,\theta_k^\ast) \in \mathbb{R}^{p \times k}
		\]
		be vectors (representing true cluster centers). Let $n \ge 2$ and let
		\[
		z^\ast=(z_1^\ast,\dots,z_n^\ast)\in [k]^n
		\]
		be a vector (representing cluster assignment vector). Define the ideal selection matrix $A\in\mathbb{R}^{n\times n}$ by
		\[
		A_{ij} :=
		\begin{cases}
			1, &  z_i^\ast \neq z_j^\ast,\\
			0, & \text{otherwise},
		\end{cases}
		\]
		Thus $A$ equals one precisely on cross-cluster pairs and zero on same-cluster pairs. Let
		\[
		D := \mathcal{D}(A)
		\]
		be the ideal degree matrix. Let
		\[
		L := \mathcal{L}(A)
		\]
		be the ideal Laplacian matrix.
		Then $(A,D,L)$ is called the ideal truncation matrices for the true cluster configuration $(\Theta^*,z^*)$.
	\end{definition}

	Throughout the manuscript, $(\widehat A,\widehat D,\widehat L)$ denotes the data truncation matrices as in Definition \ref{def:datatrunc}, whereas $(A,D,L)$ denotes the ideal truncation matrices of Definition~\ref{def:ideal-truncation-model}.

	\subsection{The transformed data and noise reduction}
	Let 
	\[
	\widehat S := \left(\frac{\widehat L}{n}\right)^{1/2}\in\mathbb{R}^{n\times n}.
	\]
    be the square root of the scaled Laplacian. This is well defined by Lemma \ref{lem:psd} and Lemma~\ref{lem:sqrt}.
    We then work on the transformed data matrix
	\[
	Y := X\widehat S \in \mathbb{R}^{p\times n}.
	\]

    By definition we have $\widehat L=\mathcal L(\widehat A)$, so $YY^\top$ is the kernalised covariance matrix, i.e.,
    \begin{equation}
    \label{wb:eq:truncated-pairwise-covariance}
    \begin{split}
        YY^\top
        &=
        X\frac{\widehat L}{n}X^\top
        \\
        &=
        \frac1{2n}
        \sum_{i,j=1}^n
        \widehat A_{ij}
        (X_i-X_j)(X_i-X_j)^\top.
    \end{split}
    \end{equation}
    Observed that $\widehat L\one_{[n]}=0$ which implies $\widehat S\one_{[n]}=0$, and therefore
    \[
        Y=X\widehat S=XH_n\widehat S,
    \]
    i.e., the transformed data is translation invariant and implicitly centered.

    To see the intended noise reduction, let $A$ be the ideal selection matrix, let $L=\mathcal L(A)$, and consider the ideal transform
    \[
        S_0:=\left(\frac Ln\right)^{1/2},
        \quad
        Y_0:=XS_0.
    \]
    Then we have
    \[
        Y_0Y_0^\top
        =
        \frac1{2n}
        \sum_{z_i^*\ne z_j^*}
        (X_i-X_j)(X_i-X_j)^\top.
    \]
    The above identity shows that, compared with the raw centered covariance matrix, the ideal transform removes exactly the within-cluster contribution
    \[
        \frac1{2n}
        \sum_{z_i^*=z_j^*}
        (\varepsilon_i-\varepsilon_j)
        (\varepsilon_i-\varepsilon_j)^\top.
    \]
    This contribution contains no signal and only noise.

    Remarkably, this ideal transform removes these terms without distorting the centered signal.  Indeed, Remark~\ref{wb:prop:effective-dimension} shows that
    \[
        PS_0=P_c.
    \]
    In addition, if $n_{\max}$ is the size of a largest cluster and $m=n-n_{\max}$, then
    \begin{align}\label{eq:idealLnoise}
        \|ES_0\|_{\rm op}
        \lesssim \sigma\sqrt{p+m} \text{ with high probability } 1-\exp(-C(p+m)).
    \end{align}  The ideal transformed matrix therefore has the decomposition
    \[
        Y_0=P_c+ES_0,
    \]
    which preserves the full centered signal while reducing the magnitude of noise from the $O(\sigma\sqrt{p+n})$ to $O(\sigma\sqrt{p+m})$.

    In practice, our transformed data matrix $Y=X\widehat S$ has to use $\widehat A$ instead of the unknown ideal matrix $A$, so we need to choose a threshold function $\phi$ so that $\widehat A$ is close to $A$, which we will discuss in Section \ref{sec: thresholdchoice}. Also, the analysis needs to be done more carefully, because of the dependence of $\widehat A$  and the data matrix $X$. In particular, we cannot work on a good event where $\widehat A=A$ because the probability of this event is not strong enough for our purpose, i.e., to obtain $1-\exp(-C(p+m))$ high probability guarantee of dimension reduction and clustering, as in the ideal case. In fact, by standard calculation, when the separation is as weak as $\Delta \approx \sigma p^{1/4}$, 
the pairwise error $\Pb(\hat A_{ij}\ne A_{ij})$ is approximately only of constant order. So it does not lead to simultaneous matrix equality
$\hat A=A$ with our target high probability $1-\exp(-C(p+m))$.

	\subsection{Dimension reduction after truncation}
    \label{wb:subsec:dimension-reduction-after-truncation}

    We now describe the dimension reduction method using the transformed matrix $Y=X\widehat S$.  Let
    \[
        Y
        =
        \sum_{j=1}^{\min\{p,n\}}
        \widehat\sigma_j\,
        \widehat u_j\widehat v_j^\top,
        \quad
        \widehat\sigma_1\ge\widehat\sigma_2\ge\cdots\ge0,
    \]
    be a singular value decomposition.  For $1\le r\le\min\{p,n\}$, let
    \[
        \widehat U_r
        :=[\widehat u_1,\ldots,\widehat u_r],
        \quad
        \widehat\Sigma_r
        :=\diag(\widehat\sigma_1,\ldots,\widehat\sigma_r),
        \quad
        \widehat V_r
        :=[\widehat v_1,\ldots,\widehat v_r].
    \]
    where $\widehat U_r$ is a truncated estimator for the
subspace where the signals, i.e., true cluster centers, live in.
    Equivalently, $\widehat U_r$ can also be computed from the eigenvalue decomposition of the truncated covariance matrix, i.e.,
    \[
        \sum_{j=1}^{\min\{p,n\}}
        \widehat\sigma_j^2
        \widehat u_j\widehat u_j^\top=YY^\top
        =
        X\frac{\widehat L}{n}X^\top
        = X\frac{\widehat L}{n}X^\top,
    \]

    As before, 
    let $r_0=\rank(P_c)$ and let $U_0$ be an orthonormal basis for $\operatorname{col}(P_c)$. 
    The dimension-reduction error is measured by
    \[
        \left\|
            \widehat U_{r_0}\widehat U_{r_0}^\top
            -
            U_0U_0^\top
        \right\|_{\rm op}.
    \]

	\subsection{Clustering post dimension reduction}
Pursuant to choosing 
    \[
        r\in\{1,\ldots,\min\{p,k\}\}.
    \]
    (typical choices are $r=\min\{p,k\}$ or $r=\min\{p,k-1\}$), our clustering algorithm uses the projected data matrix
    \begin{equation}
    \label{wb:eq:transformed-embedding}
        \widehat P
        :=
        \widehat U_r^\top Y
        =
        \widehat\Sigma_r\widehat V_r^\top
        \in\mathbb R^{r\times n}.
    \end{equation}
    which reduces the dimension from $p$ to $r$.
    We estimate the labels by applying global $k$-means to the columns of $\widehat P$.  
	More precisely, given projected data points $\{\widehat P_{\cdot,i}\}_{i=1}^n\subset\mathbb{R}^r$, the global $k$-means problem finds labels $z\in[k]^n$ and centers $\nu_1,\dots,\nu_k\in\mathbb{R}^r$ that minimize
	\[
	\sum_{i=1}^n \bigl\| \widehat P_{\cdot,i} - \nu_{z_i} \bigr\|_2^2.
	\] 
    The formal algorithm is shown in Algorithm~\ref{clusteringalgo}.
	
	\begin{algorithm}{Clustering via a truncated covariance matrix}\label{clusteringalgo}
		\noindent\textbf{Input:}
		Data matrix $X=[X_1,\dots,X_n]\in\mathbb{R}^{p\times n}$; number of clusters $k$; truncation function $\phi$; embedding dimension $r\in\{1,\dots,\min\{p,k\}\}$.

		\smallskip
		\noindent\textbf{Output:}
		Estimated labels $\widehat z\in[k]^n$.

		\begin{enumerate}
			\item Form the kernel matrix $\widehat A\in\mathbb{R}^{n\times n}$ by $\widehat A_{ij}=\phi(\|X_i-X_j\|_2^2)$ for all $i,j\in[n]$.
			\item Form the row-sum matrix $\widehat D=\mathrm{diag}(\widehat A\one_{[n]})$ and set $\widehat L=\widehat D-\widehat A$.
			\item Compute the scaled positive-semidefinite square root $\widehat S=\left(\widehat L/n\right)^{1/2}$.
			\item Form the transformed data matrix $Y=X\widehat S\in\mathbb{R}^{p\times n}$.
			\item Compute a singular value decomposition $Y=\sum_{j=1}^{\min\{p,n\}}\widehat\sigma_j \widehat u_j \widehat v_j^{\mathsf{T}}$ with $\widehat\sigma_1\ge \widehat\sigma_2\ge \cdots$.
			\item Form $\widehat U_r=[\widehat u_1,\dots,\widehat u_r]$, $\widehat V_r=[\widehat v_1,\dots,\widehat v_r]$, and $\widehat\Sigma_r=\mathrm{diag}(\widehat\sigma_1,\dots,\widehat\sigma_r)$; set $\widehat P=\widehat U_r^\top Y\in\mathbb{R}^{r\times n}$.
			\item Let $\widehat z\in[k]^n$ be a global minimizer of the $k$-means objective on the columns $\{\widehat P_{\cdot,i}\}_{i=1}^n$.
		\end{enumerate}
	\end{algorithm}

	\subsection{Metrics for clustering error}
	Following \cite{LZZ2021}, we consider the  average clustering error below.
    \begin{definition}[Average clustering error] Let $z\in [k]^n$ and $z^\ast\in [k]^n$, the average clustering error is defined as
	\[
	\ell(\widehat z,z^\ast)
	:= \min_{\pi\in\mathfrak S_k} \frac{1}{n}\sum_{i=1}^n \mathbf{1}\{\pi(\widehat z_i)\neq z_i^\ast\},
	\]
	where $\mathfrak S_k$ is the set of all bijections (permutations) $\pi:[k]\to[k]$.
    \end{definition}
    In regimes where one large cluster dominates (so the complement of a largest cluster is small), an average loss can hide errors on those smaller clusters.  We therefore also consider the maximum clustering error.
    \begin{definition}[Maximum clustering error] Let $z\in [k]^n$ and $z^\ast\in [k]^n$, the maximum clustering error is defined as
	\[
	\ell_{\infty}(\widehat z,z^\ast)
	:=
	\min_{\pi\in\mathfrak S_k}
	\max_{a\in[k]}
	\frac{1}{n_a}\sum_{i\in B_a}\mathbf{1}\{\pi(\widehat z_i)\neq a\},
	\]
	where $\mathfrak S_k$ is the set of all bijections (permutations) $\pi:[k]\to[k]$.
    \end{definition}
	
    \subsection{On the choice of the kernel function}\label{sec: thresholdchoice}
In order to understand the suitability of the kernel function, let us consider the simplified setting where $k=2$ and $\norm{\theta_1^*-\theta_2^*}_2=\Delta$. There are two natural types of threshold functions, a smooth function, e.g., the one considered in \cite[Example~1]{AminiRazaee2021}, and a function with a jump, i.e., hard threshold. For a smooth threshold function, the most natural example is the one that leads to the Gaussian kernel matrix in \cite[Example~1]{AminiRazaee2021}.
More precisely, for $\tau>0$, define
\[f_\tau(\sqrt t)=e^{-t/(2\tau^2)}\] as in \cite[Example~1]{AminiRazaee2021}.
and choose the kernel function to be
\[
\phi_{\tau}(t)=1-f_\tau(\sqrt t),
\]
And write
$(\widehat A_{G,\tau})_{ij}=\phi(\norm{X_i-X_j}_2^2)$.
Let $s^2=1+2\sigma^2/\tau^2$. Applying the Gaussian expectation
formula in \cite[Example~1]{AminiRazaee2021},
for $i\ne j$, we have
\[
\E[(\widehat A_{G,\tau})_{ij}]
=\begin{cases}
a_\tau:=1-s^{-p}, & z_i^*=z_j^*,\\[2mm]
b_\tau:=1-s^{-p}
\exp\left(-\dfrac{\Delta^2}{2\tau^2s^2}\right),
& z_i^*\ne z_j^*.
\end{cases}
\]
Their difference vanishes in the weak-separation regime
$\Delta=o(\sigma\sqrt p)$ as $p\to\infty$, uniformly over all
deterministic bandwidths $\tau>0$, including those depending on $p$.
Indeed, using the short notation $q=s^{-2}\in(0,1)$, by direct calculation, we have
\[
\begin{aligned}
0\le b_\tau-a_\tau
&=q^{p/2}
  \left(1-e^{-\Delta^2(1-q)/(4\sigma^2)}\right)\\
&\le \frac{\Delta^2}{4\sigma^2}q^{p/2}(1-q)\\
&\le \frac{\Delta^2}{2\sigma^2(p+2)}
  \left(\frac{p}{p+2}\right)^{p/2}
=o(1).
\end{aligned}
\]
Here we used $1-e^{-x}\le x$ and maximized over $q\in(0,1)$.
This is an issue, because in the ideal case, we have $A_{ij}=0$ within clusters and $A_{ij}=1$ across clusters. So, to make $\E\widehat A_{G,\tau} \approx A$, we will need $a_\tau \approx 0$ and $b_\tau \approx 1$ so their differences cannot vanish. More precisely,
the entrywise bias is
\[
(\E\widehat A_{G,\tau}-A)_{ij}
=\begin{cases}
a_\tau, & z_i^*=z_j^*,\\
b_\tau-1, & z_i^*\ne z_j^*,
\end{cases}
\quad i\ne j.
\]
Because $0\le a_\tau\le b_\tau\le1$, we have
\[
\max\{a_\tau,1-b_\tau\}
\ge \frac{1-(b_\tau-a_\tau)}{2}
=\frac12-o(1).
\]
Therefore, no choice of $\tau$ can make 
$\E\widehat A_{G,\tau}$ close to $A$ simultaneously on within-cluster
and cross-cluster entries, once we enter the in the weak-separation regime
$\Delta=o(\sigma\sqrt p)$.

We now show that we can do much better by a hard threshold function with jump (transition) at
$u_0=2\sigma^2p+\Delta^2/2$, i.e.,
 \begin{align*}
            \phi(t)=\mathbbm{1}_{\{t\ge u_0\}},
        \quad t\ge0
        \end{align*}
which leads to
\[
\widehat A_{ij}
=\mathbbm{1}_{\{\norm{X_i-X_j}_2^2\ge u_0\}}.
\]
For $i\ne j$, the squared distance $\norm{X_i-X_j}_2^2$ of two data points has the central chi-square law
$2\sigma^2\chi_p^2$ if $X_i,X_j$ are from the same cluster. And $\norm{X_i-X_j}_2^2$ has the noncentral chi-square law
$
2\sigma^2\chi_p^2\!\left(\frac{\Delta^2}{2\sigma^2}\right)
$ if $X_i,X_j$ are from different clusters. So, after the scaling by $\frac{1}{2\sigma^2}$, the means for same cluster and cross cluster squared distances are
\[
p
\quad\text{and}\quad
p+\frac{\Delta^2}{2\sigma^2}
\]
respectively. The scaled threshold $p+\Delta^2/(4\sigma^2)$ lies in the middle
between the two means, and its distance from either mean is
$\Delta^2/(4\sigma^2)$. Meanwhile, the standard deviations for same cluster and cross cluster squared distances  are
\[
\sqrt{2p}
\quad\text{and}\quad
\sqrt{2\left(p+\frac{\Delta^2}{\sigma^2}\right)}.
\]
In the weak-separation regime, these standard deviations are both
of order $\sqrt p$. Therefore, hard truncation can distinguish the two
pair types once the shift $\Delta^2$ dominates the fluctuations
of order $\sigma^2\sqrt p$, which only requires $\Delta \gg \sigma p^{1/4}$.

As a result, in later parts of this paper, we do  dimension reduction and clustering using this hard threshold function, i.e.,
\begin{equation}\label{wb:eq:oracle-threshold}
        \phi(t)=\mathbbm{1}_{\{t\ge 2\sigma^2p+\Delta^2/2\}},
        \quad t\ge0.
\end{equation}

\section{Main Results}

	\subsection{Dimension Reduction}
In this section, we state our main result on dimension reduction via truncated covariance matrices.
    
\begin{theorem}[Dimension reduction via the truncated matrix estimator]\label{wb:thm:dimension-reduction}
Let $k\ge 2$, $n \ge 2$, and $p \ge 1$. Let
		\[
		\Theta^*=(\theta_1^\ast,\dots,\theta_k^\ast) \in \mathbb{R}^{p \times k}
		\]
		and
		\[
		z^\ast=(z_1^\ast,\dots,z_n^\ast)\in [k]^n.
		\]
        For $a\in[k]$, let $B_a=\{i\in[n]:z_i^*=a\}$ and $n_a=|B_a|>0$. Let $n_{\max}:=\max_{1\le a\le k}n_a$ and  $m:=n-n_{\max}$.
		Let $(X,P,E)$ be a Gaussian mixture observation model for the true cluster configuration $(\Theta^*,z^*)$ and noise level $\sigma$.
		Let 
        \begin{align*}
            \phi(t)=\mathbbm{1}_{\{t\ge 2\sigma^2p+\Delta^2/2\}},
        \quad t\ge0
        \end{align*} Let $(\hat A, \hat D, \hat L)$ be the data truncation matrices for the data $X$ using threshold function $\phi$. 
Let $\Delta>0$ and assume
\[
\norm{\theta_a^*-\theta_b^*}_2\ge\Delta\quad\text{for all }a\ne b.
\]
Let
\[
\hat S=\left(\frac{\hat L}{n}\right)^{1/2},
\quad
Y=X\hat S.
\]
Let
\[
H_n:=I_n-\frac1n\one_{[n]}\one_{[n]}^\top,
\quad
P_c:=PH_n,
\quad
r_0:=\rank(P_c).
\]
Assume $r_0\ge1$.  Let $U_0\in\mathbb R^{p\times r_0}$ be any orthonormal basis for $\operatorname{col}(P_c)$, and let $\hat U_{r_0}$ be the matrix of the top $r_0$ left singular vectors of $Y=X\hat S$.   There exist positive absolute constants
$c_{\ref{wb:thm:dimension-reduction}.1},\ldots,
c_{\ref{wb:thm:dimension-reduction}.4}$ such that if
\[
\frac{\Delta^2}{\sigma^2}
\ge
c_{\ref{wb:thm:dimension-reduction}.1}
\left[
h+\sqrt{ph}
\right], \text{ where }h:=1+\log_+\frac{n}{p+m}
\]
then, with probability at least
\[
1-\exp(-c_{\ref{wb:thm:dimension-reduction}.2}(p+m)),
\]
we have
\begin{equation}\label{eq:separation}
\begin{split}
&\norm{\hat U_{r_0}\hat U_{r_0}^\top-U_0U_0^\top}_{\op}
\\\le&
\frac{c_{\ref{wb:thm:dimension-reduction}.3}}{\sigma_{r_0}(P_c)}
\left[
\sigma\sqrt{p+m+\log n}
+
\Delta
\sqrt{
m
\exp\left(-c_{\ref{wb:thm:dimension-reduction}.4}\frac{\Delta^4/\sigma^4}{p+\Delta^2/\sigma^2}\right)
+
(p+m+\log n)\left(\frac{\sigma^2}{\Delta^2}+\frac{p\sigma^4}{\Delta^4}\right)
}
\right].
\end{split}
\end{equation}

\end{theorem}

\begin{remark}
\label{wb:rem:dimension-weak-regime}
The separation condition in Theorem~\ref{wb:thm:dimension-reduction} has the form
\[
\Delta^2/\sigma^2
\gtrsim
h+\sqrt{ph},\quad h:=1+\log_+\frac{n}{p+m}.
\]
If $h=o(p)$, this lower bound is $o(p)$, so the theorem permits choices with
$\Delta^2/\sigma^2=o(p)$, equivalently $\Delta=o(\sigma\sqrt p)$. Since
$h$ grows slowly, suppressing this logarithmic factor gives the
approximate separation regime
\[
\Delta\gtrsim\sigma p^{1/4}.
\]
This is already in the weak-separation regime \eqref{eq:weak-separation} described above: a single noise vector has
Euclidean norm of order $\sigma\sqrt p$, whereas
$\Delta\ll\sigma\sqrt p$. Nevertheless, even in this regime, the transformed spectral perturbation
is controlled by the reduced noise scale $\sigma\sqrt{p+m+\log n}$ together with the
explicit \mbox{truncation-error term}, with desired high probability $1-\exp\left(-c_{\ref{wb:thm:dimension-reduction}.2}(p+m)\right)$ as in the ideal case \eqref{eq:idealLnoise}.
\end{remark}

\begin{remark}
It may be observed that the separation regime $\Delta \simeq \sigma p^{1/4}$ is a genuine barrier in this approach. To see why, assume all clusters are separated by distance exactly $\Delta$ for simplicity as in section \ref{sec: thresholdchoice}. In this case, recall that, the scaled squared distance
$\norm{X_i-X_j}_2^2/(2\sigma^2)$ has law $\chi_p^2$ within
clusters and $\chi_p^2(\Delta^2/(2\sigma^2))$ across clusters, so the means for same cluster and cross cluster squared distances are
\[
p
\quad\text{and}\quad
p+\frac{\Delta^2}{2\sigma^2}
\]
respectively, and their standard deviations  are
$\sqrt{2p}$ and $\sqrt{2(p+\Delta^2/\sigma^2)}$.
The scaled threshold $p+\Delta^2/(4\sigma^2)$ has distance $\Delta^2/(4\sigma^2)$ from either mean.
Therefore, $\Delta^2/\sigma^2$ must be at least of order $\sqrt p$, i.e., equivalently
$\Delta$ must be at least of order $\sigma p^{1/4}$, to make the separation
between the two means be able to compete with the random fluctuations.

More precisely, when $p\to\infty$, if
$\Delta=o(\sigma p^{1/4})$, for each pair $i,j$, we have
\[
\Pb\!\left(\widehat A_{ij}\ne A_{ij}\right)
\longrightarrow
\frac12.
\]
So the kernel matrix $\widehat A$ almost contains no useful information about whether the pair is within
or across clusters (which is specified by the ideal selection matrix $A$).

\end{remark}

We compare our Theorem \ref{wb:thm:dimension-reduction} with the following guarantees for classical PCA based dimension-reduction.
\begin{theorem}[Classical PCA based dimension-reduction]\label{wb:thm:untransformed-dim}
Let $k\ge 2$ and $p \ge 1$. Let
		\[
		\Theta^*=(\theta_1^\ast,\dots,\theta_k^\ast) \in \mathbb{R}^{p \times k}
		\]
		and let $n \ge 2$. Let
		\[
		z^\ast=(z_1^\ast,\dots,z_n^\ast)\in [k]^n
		\]
		be a vector. Let $(X,P,E)$ be a Gaussian mixture observation model for the true cluster configuration $(\Theta^*,z^*)$ and noise level $\sigma$.
Define
\[
P_c:=P\left(I_n-\frac1n\one_{[n]}\one_{[n]}^\top\right),
\quad
X_c:=X\left(I_n-\frac1n\one_{[n]}\one_{[n]}^\top\right),
\quad
r_0:=\rank(P_c).
\]
Assume $r_0\ge1$. Let $U_c\in\bbR^{p\times r_0}$ be an orthonormal basis for the column space of $P_c$, and let $\hat U_{\rm PCA}\in\bbR^{p\times r_0}$ be the matrix of the top $r_0$ left singular vectors of $X_c$. There exist positive absolute constants
$c_{\ref{wb:thm:untransformed-dim}.1}$ and
$c_{\ref{wb:thm:untransformed-dim}.2}$ such that, with probability at least
\[
1-\exp(-c_{\ref{wb:thm:untransformed-dim}.1}n),
\]
\[
\norm{\hat U_{\rm PCA}\hat U_{\rm PCA}^\top-U_cU_c^\top}_{\op}
\le
c_{\ref{wb:thm:untransformed-dim}.2}
\frac{\sigma\sqrt{p+n}}{\sigma_{r_0}(P_c)}.
\]
\end{theorem}

\begin{remark}
\label{wb:rem:dim-comparison}In practical applications, it is reasonable to assume $\log n \le p$ so we can ignore that term. 
Under this assumption, the classical PCA based result, i.e., Theorem \ref{wb:thm:untransformed-dim},  bounds the perturbation by
\(\sigma\sqrt{p+n}\), whereas the transformed spectral perturbation (in Theorem \ref{wb:thm:dimension-reduction}) is
controlled by the main term \[ \sigma\sqrt{p+m}\]which replaces $n$ by $m$,
plus the 
truncation-error term \[e(m,p,\Delta,\sigma) := \Delta
\sqrt{
m
\exp\left(-c_{\ref{wb:thm:dimension-reduction}.4}\frac{\Delta^4/\sigma^4}{p+\Delta^2/\sigma^2}\right)
+
(p+m)\left(\frac{\sigma^2}{\Delta^2}+\frac{p\sigma^4}{\Delta^4}\right)
}\]
which also only depends on $p$ and $m$, not $n$. These suggest that our truncation based method provides a better guarantee when $n$ dominates. For example, if $\sqrt{p}(p+m) \ll n$, then we have both $\sigma\sqrt{p+m}=o(\sigma\sqrt{p+n})$ and $e(m,n,p,\Delta,\sigma) = o(\sigma\sqrt{p+n})$.
In this setting, Theorem \ref{wb:thm:dimension-reduction} gives an improved guarantee even under weak separation regime as defined in \eqref{eq:weak-separation}.

In particular, when the cluster sizes are imbalanced so that $
    m \ll \frac{n}{\sqrt p}
$
and also the sample size is larger than the dimension so that
$
    n \gg p^{3/2},
$
 then the condition $\sqrt{p}(p+m) \ll n$ holds.
In that setting (captured by the parameter range $n \gg \max\{p^{3/2},m\sqrt{p}\}$), for example, the transformed spectral perturbation obeys the bound
\begin{align*}
    &\norm{\hat U_{r_0}\hat U_{r_0}^\top-U_0U_0^\top}_{\op}
\\=&
O\left(\frac{1}{\sigma_{r_0}(P_c)}
\left[
\sigma\sqrt{p+m}
+
e(m,n,p,\Delta,\sigma)
\right]\right)
\\ \ll & \frac{1}{\sigma_{r_0}(P_c)}\sigma\sqrt{p+n}
\end{align*}
which is a direct improvement over the bound in Theorem \ref{wb:thm:untransformed-dim} for classical untransformed PCA.
It may be noted that the condition $\sqrt{p}(p+m) \ll n$ discussed above is merely an example of a sufficient regime for improvement over classical PCA; wider ranges of the parameter values $m,n,p$ would allow for such improvement as soon as the overall argument above can be executed.
\end{remark}

    \subsection{Clustering Accuracy}
In this section, we present our main results on the accuracy of clustering via truncated covariance matrices.

\begin{theorem}[Spectral clustering by the truncated estimator]
\label{wb:thm:weak-clustering} Let $k\ge 2$, $n \ge 2$, and $p \ge 1$. Let
		\[
		\Theta^*=(\theta_1^\ast,\dots,\theta_k^\ast) \in \mathbb{R}^{p \times k}
		\]
		and
		\[
		z^\ast=(z_1^\ast,\dots,z_n^\ast)\in [k]^n.
		\]
        For $a\in[k]$, let $B_a=\{i\in[n]:z_i^*=a\}$ and $n_a=|B_a|>0$. Let $n_{\max}:=\max_{1\le a\le k}n_a$, $n_{\min}:=\min_{1\le a\le k}n_a$, and  $m:=n-n_{\max}$.  Let $(X,P,E)$ be a Gaussian mixture observation model for the true cluster configuration $(\Theta^*,z^*)$ and noise level $\sigma$. 
		Let 
        \begin{align*}
            \phi(t)=\mathbbm{1}_{\{t\ge 2\sigma^2p+\Delta^2/2\}},
        \quad t\ge0
        \end{align*} Let $(\hat A, \hat D, \hat L)$ be the data truncation matrices for the data $X$ using threshold function $\phi$. 
Let $\Delta>0$ and assume
\[
\norm{\theta_a^*-\theta_b^*}_2\ge\Delta\quad\text{for all }a\ne b.
\]
Let
\[
\hat S=\left(\frac{\hat L}{n}\right)^{1/2},
\quad
Y=X\hat S.
\]
There exist positive absolute constants
$c_{\ref{wb:thm:weak-clustering}.1},\ldots,
c_{\ref{wb:thm:weak-clustering}.10}$ such that the following holds.
Set
\[
r_\ast:=\min\{k-1,p\},
\quad
h:=1+\log_+\frac{n}{p+m}.
\]
Assume
\begin{equation}\label{wb:eq:clustering-delta-solved}
\frac{\Delta^2}{\sigma^2}
\ge
c_{\ref{wb:thm:weak-clustering}.1}
\left\{
\left[
h+
\frac{r_\ast(p+m+\log n)}{n_{\min}}
\right]
+
\sqrt{
p\left[
h+
\frac{r_\ast(p+m+\log n)}{n_{\min}}
\right]
}
\right\}.
\end{equation}
Let $\hat U_{r_\ast}$ be the matrix of the top $r_\ast$ left singular vectors of $Y$, let
\[
\hat P:=\hat U_{r_\ast}^\top Y\in\bbR^{r_\ast\times n},
\]
and choose $(\hat z,\hat\nu_1,\ldots,\hat\nu_k)$ from
\[
\mathop{\argmin}_{z\in[k]^n,\,\nu_1,\ldots,\nu_k\in\mathbb R^{r_\ast}}
\sum_{i=1}^n\|\hat P_{\cdot,i}-\nu_{z_i}\|_2^2.
\]

Then, with probability at least
\[
1-\exp(-c_{\ref{wb:thm:weak-clustering}.2}(p+m)),
\]
the average and maximum clustering errors satisfy
\begin{align*}
&\ell(\hat z,z^\ast)
\\\le&
c_{\ref{wb:thm:weak-clustering}.3}r_\ast
\left[
\frac{\sigma^2(p+m+\log n)}{n\Delta^2}
+
\frac{p\sigma^4(p+m+\log n)}{n\Delta^4}
+
\frac mn
\exp\left(-c_{\ref{wb:thm:weak-clustering}.4}\frac{\Delta^4/\sigma^4}{p+\Delta^2/\sigma^2}\right)
\right],
\end{align*}
and
\begin{align*}
&\ell_\infty(\hat z,z^\ast)
\\\le&
c_{\ref{wb:thm:weak-clustering}.5}r_\ast
\left[
\frac{\sigma^2(p+m+\log n)}{n_{\min}\Delta^2}
+
\frac{p\sigma^4(p+m+\log n)}{n_{\min}\Delta^4}
+
\frac m{n_{\min}}
\exp\left(-c_{\ref{wb:thm:weak-clustering}.6}\frac{\Delta^4/\sigma^4}{p+\Delta^2/\sigma^2}\right)
\right].
\end{align*}

\end{theorem}

\begin{remark}
\label{wb:rem:weak-clustering-regime}
Let
\[
        H:=1+\log_+\frac{n}{p+m}+\frac{r_\ast(p+m+\log n)}{n_{\min}}.
\]
The sufficient separation condition \eqref{wb:eq:clustering-delta-solved} is
\[
        \Delta^2/\sigma^2\gtrsim H+\sqrt{pH}.
\]
When $H=o(p)$, the sufficient condition permits choices for which $\Delta/\sigma$ is of order $(pH)^{1/4}=o(\sqrt p)$, our truncation based clustering method is guaranteed to work in weak separation regime as defined in \eqref{eq:weak-separation}.
\end{remark}
We compare our Theorem \ref{wb:thm:weak-clustering} with the following guarantees for classical PCA based spectral clustering.

\begin{theorem}[Classical PCA based spectral clustering; see e.g., Proposition 3.1. in \cite{zhang-zhou2024}]\label{wb:thm:untransformed-clustering}
Let $k\ge 2$ and $p \ge 1$. Let
		\[
		\Theta^*=(\theta_1^\ast,\dots,\theta_k^\ast) \in \mathbb{R}^{p \times k}
		\]
		and let $n \ge 2$. Let
		\[
		z^\ast=(z_1^\ast,\dots,z_n^\ast)\in [k]^n
		\]
		be a vector. For $a\in[k]$, let $B_a=\{i\in[n]:z_i^*=a\}$ and $n_a=|B_a|>0$. Let $n_{\min}:=\min_{1\le a\le k}n_a$. Let $(X,P,E)$ be a Gaussian mixture observation model for the true cluster configuration $(\Theta^*,z^*)$ and noise level $\sigma$.
Let $\Delta>0$ and assume
\[
\norm{\theta_a^*-\theta_b^*}_2\ge\Delta\quad\text{for all }a\ne b.
\]
Define the centered raw data
\[
X_c=X\left(I_n-n^{-1}\one_{[n]}\one_{[n]}^\top\right).
\]
Let $r_\ast:=\min\{k-1,p\}$, let $\hat U_{\rm PCA}$ be the matrix of
the top $r_\ast$ left singular vectors of $X_c$, and set
\[
\hat P_{\rm PCA}:=\hat U_{\rm PCA}^\top X_c\in\bbR^{r_\ast\times n}.
\]
Choose $(\hat z_{\rm PCA},\hat\nu_1,\ldots,\hat\nu_k)$ from
\[
\mathop{\argmin}_{z\in[k]^n,\,\nu_1,\ldots,\nu_k\in\mathbb R^{r_\ast}}
\sum_{i=1}^n\|\hat P_{{\rm PCA},\cdot,i}-\nu_{z_i}\|_2^2.
\]

There exist positive absolute constants $c_{\ref{wb:thm:untransformed-clustering}.1},\ldots,
c_{\ref{wb:thm:untransformed-clustering}.4}$ 
such that the following holds.
If
\begin{equation*}
\frac{\Delta^2}{\sigma^2}
\ge
c_{\ref{wb:thm:untransformed-clustering}.1}
\frac{(k-1)(n+p)}{n_{\min}},
\end{equation*}
then, with probability at least
\[
1-\exp(-c_{\ref{wb:thm:untransformed-clustering}.2}n),
\]
the resulting labels $\hat z_{\rm PCA}$ satisfy
\[
\ell(\hat z_{\rm PCA},z^\ast)
\le
c_{\ref{wb:thm:untransformed-clustering}.3}(k-1)
\frac{\sigma^2(n+p)}{n\Delta^2},
\quad
\ell_\infty(\hat z_{\rm PCA},z^\ast)
\le
c_{\ref{wb:thm:untransformed-clustering}.4}(k-1)
\frac{\sigma^2(n+p)}{n_{\min}\Delta^2}.
\]
\end{theorem}

\begin{remark}
\label{wb:rem:clustering-comparison}
The classical PCA bounds in
Theorem~\ref{wb:thm:untransformed-clustering} contain the factor
$\sigma^2(n+p)/\Delta^2$. In
Theorem~\ref{wb:thm:weak-clustering}, the  bounds achieved by the truncation based method replaces $n+p$ by
$p+m+\log n$ with the cost of additional truncation error term.

Nevertheless, our main improvement is on the separation
requirements. Since the logarithmic factors grow slowly, for clarity, we suppress these factors and use $\approx$ when we we do so.
The right-hand
side of \eqref{wb:eq:clustering-delta-solved} then becomes
\[
\begin{split}
  &h+\frac{r_\ast(p+m+\log n)}{n_{\min}}
  +\sqrt{p\left[h+\frac{r_\ast(p+m+\log n)}{n_{\min}}\right]}\\
  \approx&
  \frac{r_\ast(p+m)}{n_{\min}}
  +\sqrt{\frac{pr_\ast(p+m)}{n_{\min}}}.
\end{split}
\]
To see the gain when one cluster dominates, fix $k$ and suppose that
\[
  m=o(n),\quad n_{\min}\asymp m,
  \quad p+\log n=O(m),
\]
i.e., clusters outside the largest one then have comparable sizes.
For the separation comparisons, take $\Delta=\Delta_*$.
The separation requirements become approximately (ignoring the logarithmic terms)
\[
\begin{aligned}
  \frac{\Delta}{\sigma}&\gtrsim p^{1/4}
  &&\text{for the truncation based method},\\
  \frac{\Delta}{\sigma}&\gtrsim\sqrt{\frac nm}
  &&\text{for the classical PCA method}.
\end{aligned}
\]
The former is substantially smaller when $\sqrt p\ll n/m$, i.e., when the first cluster significantly dominates and number of observations is not too small compared to the dimension. Note that this comparison of separation requirements also extends to stronger results for the classical PCA based spectral clustering \cite{LZZ2021,zhang-zhou2024}, where they achieves minimax rates, i.e., the expected average clustering error satisfies
\[\E\ell(\hat z,z^\ast) \le \exp(-(1-o(1))\frac{\Delta^2}{8\sigma^2})+\exp(-cn)\]
Those clustering error guarantees (e.g., Theorem~3.3 in \cite{zhang-zhou2024}) have similar separation requirement as in Theorem~\ref{wb:thm:untransformed-clustering}. For example, Theorem~3.3 in \cite{zhang-zhou2024} requires
$
  {\Delta/\sigma} \gtrsim 
       {k^3\sqrt{n/n_{\min}}(1+p/n)}.
$
When one cluster dominates, this separation requirement implies the same requirement
$
  \frac{\Delta}{\sigma}\gtrsim\sqrt{\frac nm}.
$

For a more concrete example, take $k=2$ and
$n_{\min}=m\asymp p\asymp n^{1/2}$. Our truncation based method only requires
$\Delta/\sigma\approx n^{1/8}$, compared with
$\Delta/\sigma\approx n^{1/4}$ needed for the classical PCA based spectral clustering.
\end{remark}

\section{Simulation experiments}\label{sec:simulations} We now show results from simulation experiments to corroborate the theoretical prediction that our truncation based method requires much weaker separation than the classical PCA based methods for successful clustering, under imbalanced regime. Throughout our simulation, we set $n=4096$, $p=128$, and $\sigma=1$. We generate independent
observations from model~\eqref{eq:gmm-def}, with fixed labels and
\[
 X_i=\theta_{z_i^*}^*+\varepsilon_i,\qquad
 \varepsilon_i\sim N(0,I_p),\qquad
 \theta_a^*=\frac{\Delta}{\sqrt{2}}q_a,\qquad
 q_a^{\mathsf T}q_b=\mathbf{1}\{a=b\}.
\]
For each $k$, the orthonormal directions are generated once, independently of
the noise, and held fixed. Therefore every pair of distinct cluster centers has distance exactly
$\Delta$. The uncentered signal matrix $P$ has rank $k$, and its centered version $P_c$ has
rank $k-1$.

For $k=2$, the cluster sizes are $(4096-m,m)$ with
$m=2048,512,128,32$. The first allocation is balanced, and the subsequent
minorities have sizes $4p$, $p$, and $p/4$. For $k=5$, the balanced allocation has cluster sizes $(820,819,819,819,819)$. And for the imbalanced allocations, we use
$m=512,128,64$, where $m$ denotes the total size outside the largest cluster.
For each $m$, the largest cluster has size $4096-m$, and the four smaller
clusters have the ratios in Table~\ref{tab:simulation-allocations}.

\begin{table}[htbp]
\centering
\caption{Minority cluster sizes for $k=5$. The remaining cluster has size
$4096-m$. Each column is a separate separation experiment at each value of $m$.}
\label{tab:simulation-allocations}
\small
\begin{tabular}{rcccc}
\toprule
$m$ & $1:1:1:1$ & $3:3:1:1$ & $4:2:1:1$ & $26:4:1:1$\\
\midrule
512 & $(128,128,128,128)$ & $(192,192,64,64)$ & $(256,128,64,64)$ & $(416,64,16,16)$\\
128 & $(32,32,32,32)$ & $(48,48,16,16)$ & $(64,32,16,16)$ & $(104,16,4,4)$\\
64  & $(16,16,16,16)$ & $(24,24,8,8)$ & $(32,16,8,8)$ & $(52,8,2,2)$\\
\bottomrule
\end{tabular}
\end{table}

We compare three procedures. \emph{Truncated} is
Algorithm~\ref{clusteringalgo} with $r=k-1$ and the hard threshold
\[
 \widehat A_{ij}=\mathbf{1}\!\left\{
 \|X_i-X_j\|_2^2\geq 2\sigma^2p+\frac{\Delta^2}{2}\right\}.
\]
It uses $Y=X(\widehat L/n)^{1/2}$ and clusters the columns of
$\widehat U_{k-1}^{\mathsf T}Y$. For simplicity, the truncated threshold uses 
the true minimum separation
as the specified separation lower bound. 
\emph{Centered PCA} uses the leading $k-1$ left singular directions of the centered data
$X-\bar X\mathbf{1}^{\mathsf T}$, as in
Theorem~\ref{wb:thm:untransformed-clustering}.
\emph{Zhang--Zhou PCA} uses the leading $k$ left singular directions of the uncentered data
$X$, following Algorithm~1 and the rank choice in Section~3.5 of
\cite{zhang-zhou2024}. Each procedure run $k$-means  on its corresponding projected data.
The only algorithmic substitution is to use Lloyd's iterative algorithm for the
global $k$-means step, implemented by scikit-learn with its greedy k-means++
initialization \cite{lloyd,ArthurVassilvitskii2007,sklearn2011}. 

The clustering experiments for each allocation uses the same $\Delta/\sigma$ grid
\[
 \Delta/\sigma=2,3,4,5,6,8,10,12,16,20,32,48,64
\]
and eight independent Gaussian noise replicates for every choice of $\Delta/\sigma$. Each figure
places the average loss above the maximum loss for the same allocation and we use the more straightforward label
``worst-cluster error'' to denote the maximum loss.
Curves show replicate means, with confidence intervals
$\bar\ell\mathbin{\pm}t_{7,0.975}s/\sqrt{8}$ (clipped to the meaningful range $[0,1]$).
Here $s$ is the sample standard deviation across replicates and
$t_{7,0.975}$ is the Student-$t$ quantile with seven degrees of freedom.
These are descriptive approximate 95\% pointwise intervals.

\begin{figure}[!htbp]
\centering
\small
\setlength{\abovecaptionskip}{6pt}
\includegraphics[width=6.15in]{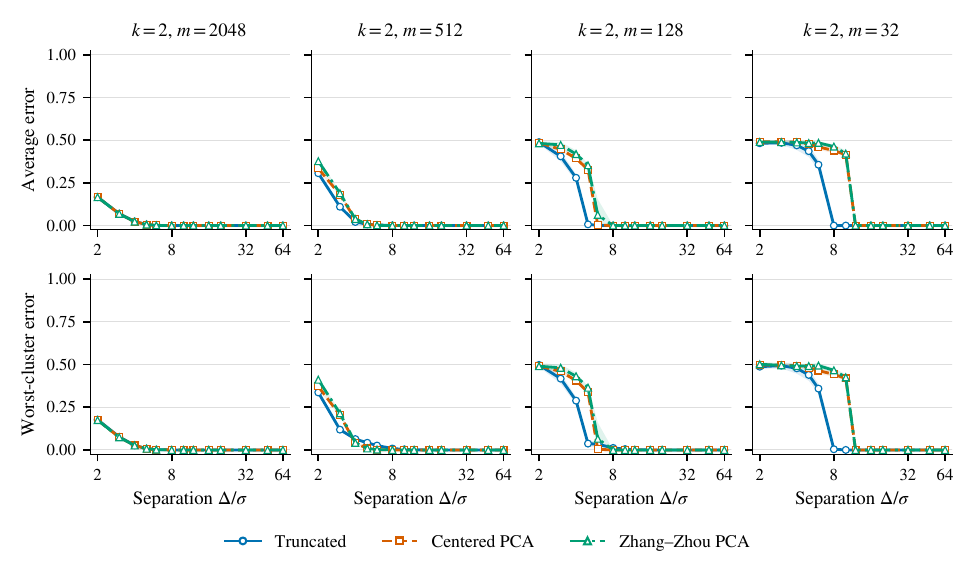}
\caption{Two-cluster separation and clustering experiments, with sizes $(4096-m,m)$ and
$m=2048,512,128,32$ from left to right. Here and in the subsequent figures, we use the following convention:
$n=4096$, $p=128$, and $\sigma=1$; the horizontal axis is $\Delta/\sigma$ on a
logarithmic scale; the top row shows average error and the bottom row shows
worst-cluster error; markers show means over eight independent noise
replicates, and shaded regions are the descriptive confidence intervals
specified in the text.}
\label{fig:simulation-k2}
\end{figure}

Figure \ref{fig:simulation-k2} shows the simulation results for two cluster case. The first two plots on the left shows that under balanced regime, our truncation based method has comparable performance with the classical PCA based methods. However, when the cluster sizes are imbalanced, our truncation based method starts to be effective from much weaker separation conditions than the classical PCA based methods. Also, the advantage of our truncation based method becomes more and more significant when the the cluster sizes become more and more imbalanced.

\begin{figure}[!htbp]
\centering
\small
\setlength{\abovecaptionskip}{6pt}
\includegraphics[width=6.15in]{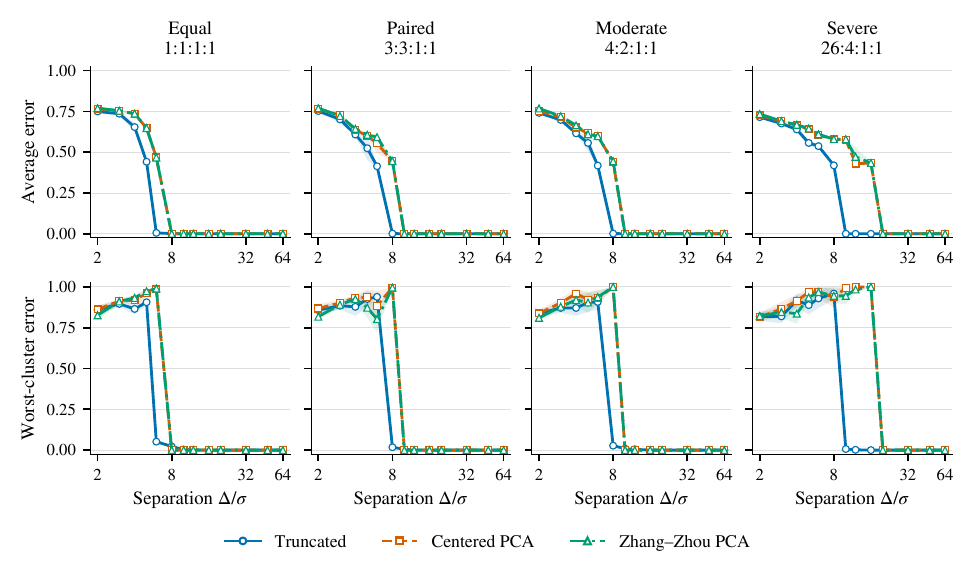}
\caption{Five-cluster separation and clustering experiments with total minority size
$m=512$ and largest-cluster size $3584$. Columns use minority ratios
$1:1:1:1$, $3:3:1:1$, $4:2:1:1$, and $26:4:1:1$; exact sizes are in
Table~\ref{tab:simulation-allocations}.}
\label{fig:simulation-k5-m512}
\end{figure}

\begin{figure}[!htbp]
\centering
\small
\setlength{\abovecaptionskip}{6pt}
\includegraphics[width=6.15in]{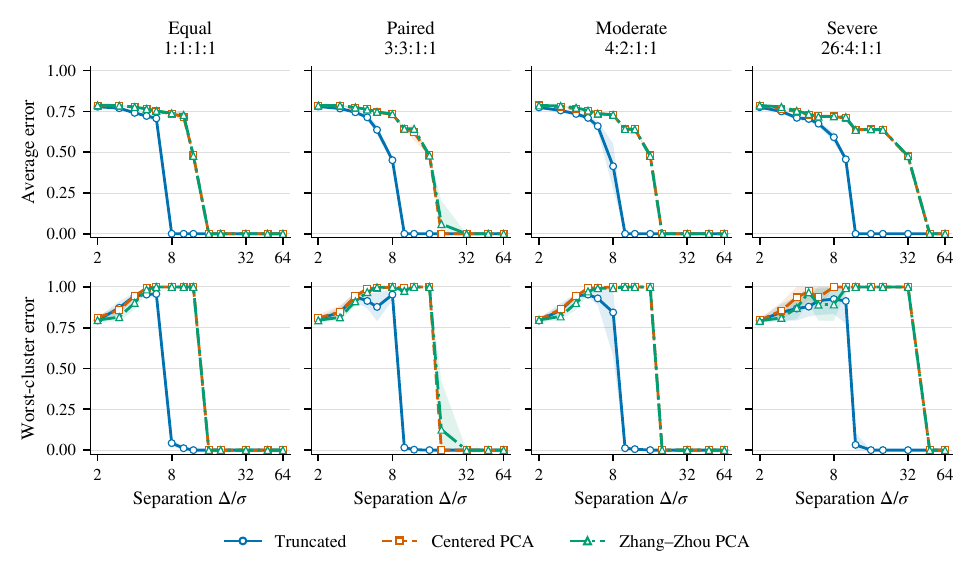}
\caption{Five-cluster separation and clustering experiments with total minority size
$m=128$ and largest-cluster size $3968$. Columns use the same four minority
ratios as Figure~\ref{fig:simulation-k5-m512}, giving smallest-cluster sizes
$32,16,16,4$.
}
\label{fig:simulation-k5-m128}
\end{figure}

\begin{figure}[!htbp]
\centering
\small
\setlength{\abovecaptionskip}{6pt}
\includegraphics[width=6.15in]{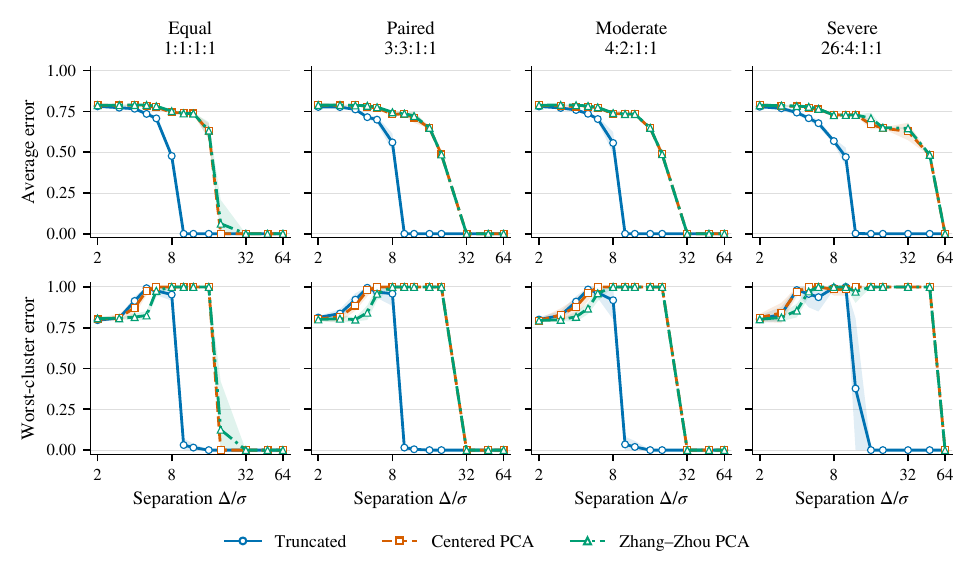}
\caption{Five-cluster separation and clustering experiments with total minority size
$m=64$ and largest-cluster size $4032$. Columns use the same four minority
ratios as Figure~\ref{fig:simulation-k5-m512}, giving smallest-cluster sizes
$16,8,8,2$. 
}
\label{fig:simulation-k5-m64}
\end{figure}

We extend the experiments to $k=5$ and the results are shown in Figures \ref{fig:simulation-k5-m512} to \ref{fig:simulation-k5-m64}. 
We decrease  $m$ (recall that $m$ is the sum of small clusters) and vary the allocation of the minority clusters. Theses extended experiments confirm that our truncation based method requires much weaker separation conditions to be effective than the classical PCA based methods under imbalanced multi-cluster settings. And this improved separation requirement holds when the minorities are also imbalanced.

\section{Proof overview}\label{wb:subsec:proof-overview}

The proof first bounds $\norm{X\hat S-P_c}_{op}$ which is the difference between the transformed data matrix $X\hat S$ and the centered signal matrix $P_c$. This bound will imply both our main results on dimension reduction and clustering by standard results on matrix perturbation theory and $k$-mean algorithm.
For this overview, we relabel the clusters so that $B_1$ is a largest
cluster, put $n_1=n-m$, and write
$S:=B_2\cup\cdots\cup B_k$, so $|S|=m$.  To motivate our discussion, we first consider the ideal truncation instead of the data dependent truncation, and analyze the corresponding error. And later we extend the analysis to data  dependent truncation, which is what we need eventually.

\subsection{Ideal truncation and noise reduction.} To understand the expected behavior of our truncated estimator, we first consider the ideal truncation $A$, which equals one on cross-cluster pairs and zero on
same-cluster pairs, with $L=\mathcal L(A)$ and $S_0=(L/n)^{1/2}$.  
Lemma~\ref{wb:lem:ideal-target} decomposes $L/n$ and therefore also $S_0=(L/n)^{1/2}$ as weighted sum of low rank projections. To explain the intuitive idea behind this decomposition, we observe that
\begin{align*}
    A=A_{[n]}-\sum\limits_{a=1}^{k}A_{B_a}
\end{align*}
where $A_{[n]}$ is the adjacency matrix of the complete graph with vertices in $[n]$ and $A_{B_a} \in \bbR^{n \times n}$ is the adjacency matrix of the complete graph with vertices in $B_a$. Then we have 
\begin{align*}
    L=\mathcal{L}(A)=\mathcal{L}(A_{[n]})-\sum\limits_{a=1}^{k}\mathcal{L}(A_{B_a}).
\end{align*}
Recall that for a subspace $T \subset \bbR^n$, $\Pi_T$ denote the orthogonal projection matrix on it. Also recall that for a subset $S\subseteq [n]$, the vector is defined by $\one_S\in\bbR^n$ by $(\one_S)_i=1$ if $i\in S$ and $(\one_S)_i=0$ otherwise. Now observe that for all $a$, we have
\begin{align*}
    A_{B_a} = \mathbbm{1}_{B_a} \mathbbm{1}_{B_a}^\top - \Pi_{\{x\in\bbR^n:\supp(x)\subseteq B_a\}}.
\end{align*}
Here $\Pi_{\{x\in\bbR^n:\supp(x)\subseteq B_a\}}$ is just the projection that sets all coordinates outside $B_a$ to $0$. In other words, we can write $\Pi_{\{x\in\bbR^n:\supp(x)\subseteq B_a\}}=I_{B_a}$ which is the identity restricted to $B_a$,  i.e., a diagonal matrix whose $i$th diagonal entry is $1$ if $i\in B_a$ and $0$ otherwise. And then we have
\begin{align*}
    A_{[n]} = \mathbbm{1}_{[n]} \mathbbm{1}_{[n]} ^\top - I_{n}.
\end{align*}
Therefore, by direct calculation, for all $a$, we have
\begin{align*}
    \mathcal{L}(A_{B_a}) = &n_a \Pi_{\{x\in\bbR^n:\supp(x)\subseteq B_a\}} - \mathbbm{1}_{B_a} \mathbbm{1}_{B_a} ^\top
    \\=&n_a \Pi_{\{x\in\bbR^n:\supp(x)\subseteq B_a,
\ \one_{B_a}^\top x=0\}}
\end{align*}
and 
\begin{align*}
    \mathcal{L} (A_{[n]}) = n I_n - \mathbbm{1}_{[n]} \mathbbm{1}_{[n]}^\top
\end{align*}
Therefore, we have
\begin{align}\label{eq:Ldecom1}
    L=&\mathcal{L}(A_{[n]})-\sum\limits_{a=1}^{k}\mathcal{L}(A_{B_a})
    \\=&n I_n - \mathbbm{1}_{[n]} \mathbbm{1}_{[n]}^\top -\sum\limits_{a=1}^{k}(n_a \Pi_{\{x\in\bbR^n:\supp(x)\subseteq B_a,
\ \one_{B_a}^\top x=0\}})\nonumber
\end{align}
In order to get a sum of projections (i.e., no ``$-$" sign), we will decompose $n I_n$ so that part of it will cancel out with $ \mathbbm{1}_{[n]} \mathbbm{1}_{[n]}^\top$ and \[\sum\limits_{a=1}^{k}(n_a \Pi_{\{x\in\bbR^n:\supp(x)\subseteq B_a,
\ \one_{B_a}^\top x=0\}})\] 
Defining
\[
q_a=\frac{\one_{B_a}}{\sqrt{n_a}},
\text{ and }
Q=[q_1\ \cdots\ q_k],
\] we can write the decomposition
\begin{align}\label{eq:Idecom1}
I_n=\sum\limits_{a=1}^{k}\Pi_{\{x\in\bbR^n:\supp(x)\subseteq B_a\}}
=&\sum\limits_{a=1}^{k}q_aq_a^\top
+\sum\limits_{a=1}^{k}\Pi_{\{x\in\bbR^n:\supp(x)\subseteq B_a,
\ \one_{B_a}^\top x=0\}}
\\=&QQ^T
+\sum\limits_{a=1}^{k}\Pi_{\{x\in\bbR^n:\supp(x)\subseteq B_a,
\ \one_{B_a}^\top x=0\}}\nonumber
\end{align}
Also, define
\[
s=(\sqrt{n_1},\ldots,\sqrt{n_k})^\top,
\]
and then we have
\[
Qs
=\sum_{a=1}^k
\frac{\one_{B_a}}{\sqrt{n_a}}\sqrt{n_a}
=\sum_{a=1}^k\one_{B_a}
=\one_{[n]}.
\]
As a result, we have
\begin{equation}\label{eq:1decom1}
\frac1n\one_{[n]}\one_{[n]}^\top
=Q\left(\frac1nss^\top\right)Q^\top.
\end{equation}
Combining \eqref{eq:Ldecom1}, \eqref{eq:Idecom1}, and \eqref{eq:1decom1}, we conclude
\begin{equation}\label{wb:eq:direct-ideal-L-final}
\frac Ln
=Q\left(I_k-\frac1nss^\top\right)Q^\top
+\sum_{a=1}^k
\left(1-\frac{n_a}{n}\right)
\Pi_{\{x\in\bbR^n:\supp(x)\subseteq B_a,
\ \one_{B_a}^\top x=0\}}.
\end{equation}
Note that
\begin{equation*}
Q\left(I_k-\frac1nss^\top\right)Q^\top
=
\Pi_{\left\{
x\in\bbR^n:
x\text{ is constant on every }B_a,
\ \one_{[n]}^\top x=0
\right\}}
\end{equation*}
is also a projection, so equation \eqref{wb:eq:direct-ideal-L-final}
decomposes $\frac Ln$ into weighted sum of orthogonal projections onto mutually orthogonal subspaces, which is essentially eigenvalue decomposition. To obtain a decomposition for $S_0=(L/n)^{1/2}$, we preserve the eigenspaces and take square root of the eigenvalues and obtain
\begin{equation}\label{wb:eq:direct-ideal-S_0-final}
S_0=(L/n)^{1/2}
=Q\left(I_k-\frac1nss^\top\right)Q^\top
+\sum_{a=1}^k
\sqrt{1-\frac{n_a}{n}}
\Pi_{\{x\in\bbR^n:\supp(x)\subseteq B_a,
\ \one_{B_a}^\top x=0\}}.
\end{equation}
With this decomposition, we show that 
\[
        PS_0=P_c=P\left(I_n-\frac1n\one_{[n]}\one_{[n]}^\top\right)
\]
which is the centered (i.e., global mean shifted to zero) version of the signal, so the ideal transform preserves the exact (centered) signal.
In addition, using the decomposition \eqref{wb:eq:direct-ideal-S_0-final} with Gaussian rotational invariance, we can decompose $ES_0$ as weighted sum of low dimensional Gaussian random matrices, and standard estimates for norm of Gaussian random matrices implies that, when $p+m\le n$, we have
\[
        \|ES_0\|_{\mathrm{op}}
        \le
        c_{\ref{wb:lem:ideal-noise-scale}.2}\sigma\sqrt{p+m}
\]
with high probability (see Lemma~\ref{wb:lem:ideal-noise-scale} for details). Comparing this with the magnitude of $\norm{E}_{\op}$ which is of order $\sqrt{p+n}$ (with high probability), we see that the ideal transform reduces the noise. In summary, the ideal transform preserves the centered signal and reduces the noise, which is exactly what we want.

\subsection{Error decomposition.} To extend the analysis from ideal truncation to data truncation, our
starting point is the decomposition
\begin{equation}\label{wb:eq:proof-overview-decomposition}
        X\hat S-P_c=P(\hat S-S_0)+E\hat S.
\end{equation}
where first term measures the perturbation of the signal and the second term measures the transformed noise.

\subsection{Bounding the signal perturbation.} We have seen that the ideal transform preserves the centered signal, but the data based transform $\hat S$ may deviate from $S_0$. So we need to estimate the extend to which the data based transform $\hat S$ perturbs the signal. 

To this end, the first step is linearization of the square root. The main difficulty in bounding $\norm{P(\hat S-S_0)}_{\op}$ is the
nonlinear dependence of $\hat S$ and $S_0$ on their Laplacians through
the matrix square root. 
To resolve this, we use Lemma~\ref{wb:lem:signal-reduction} which gives an important observation
\[
        \|P(\hat S-S_0)\|_{\mathrm{op}}
        \le
        \frac1n\|P(\hat L-L)\|_{\mathrm{op}}.
\]

The second step is to observe the cancellation of same-cluster terms.
For every symmetric zero-diagonal matrix \(H\), we have the identity
\[
P\mathcal L(H)
=
\sum_{1\le i<j\le n}
H_{ij}
(\theta_{z_i^*}^*-\theta_{z_j^*}^*)
(e_i-e_j)^\top.
\]
Here $e_i$ is the $i$th standard basis vector of $\mathbb R^n$.
Terms indexed by \(z_i^*=z_j^*\) vanish identically.  Therefore, the
same cluster part of \(\hat A-A\) (i.e., we set $H_{ij}=\mathbbm 1_{\{z_i^*= z_j^*\}}(\hat A_{ij}-A_{ij})$), which is the major source of error in the imbalanced case, makes no contribution to
\(P(\hat L-L)\). For $i,j\in[n]$, define the cross cluster  error of $\hat A_{ij}$ relative to $A_{ij}$ to be
\[
        M_{ij}
        :=\mathbbm 1_{\{z_i^*\ne z_j^*\}}(A_{ij}-\hat A_{ij})=\mathbbm 1_{\{z_i^*\ne z_j^*\}}(1-\hat A_{ij}).
\]
Then
\[
        P(\hat L-L)=-P\mathcal L(M),
\]
so this one matrix contains the entire signal perturbation error.

The third step is to bound this matrix directly.  For that purpose, we use
Lemma~\ref{wb:lem:disjoint-pair-partition} to partition the cross cluster
pairs into at most $n$ blocks $\mathcal I_a$, where distinct pairs in each
block have disjoint indices and $|\mathcal I_a|\le m$.  Define
\[
T_a:=\sum_{\{i,j\}\in\mathcal I_a}
M_{ij}(\theta_{z_i^*}^*-\theta_{z_j^*}^*)(e_i-e_j)^\top.
\]
The vectors $e_i-e_j$ for distinct pairs in a block are orthogonal.
Since entries of $M$ are 0 or 1, we have
\[
T_aT_a^\top
=2\sum_{\{i,j\}\in\mathcal I_a}
M_{ij}(\theta_{z_i^*}^*-\theta_{z_j^*}^*)(\theta_{z_i^*}^*-\theta_{z_j^*}^*)^\top.
\]
Thus each covariance matrix $T_aT_a^\top$ is a sum of independent positive-semidefinite
rank-one matrices.  We then
apply the matrix Bernstein inequality, i.e.,
Lemma~\ref{wb:thm:matrix-bernstein}, to control the norm for each $T_aT_a^\top$.
\subsection{Bounding the transformed noise.}
The main difficulty is that
$\hat S$ is constructed from the same Gaussian vectors that form $E$,
so $E\hat S$ cannot be analyzed by conditioning on $\hat S$. We resolve these difficulties in the following way.  Write 
\begin{align*}
&G_i:=\frac{\varepsilon_i}{\sigma}, \quad 1 \le i \le n, \\
&Z_{ij}:=\frac{G_i-G_j}{\sqrt2},
\quad 1\le i<j\le n.
\end{align*}
Since $\hat S^2=\hat L/n$, Lemma~\ref{wb:lem:pair-matrix-expansion} gives
\[
\norm{E\hat S}_{\op}^2
=\left\|E\frac{\hat L}{n}E^\top\right\|_{\op}
=\left\|\frac{2\sigma^2}{n}
\sum_{1\le i<j\le n}\hat A_{ij}Z_{ij}Z_{ij}^\top\right\|_{\op}.
\]
By Lemma~\ref{wb:lem:disjoint-pair-partition}, we partition all pairs
into the blocks $(\mathcal C_a)_{a=1}^{q_n}$ with $q_n \le n$,
such that distinct pairs in each block have disjoint indices. The
summands within each block are therefore independent.  

We then first bound $|v^\top (E\frac{\hat L}{n}E^\top) v|$ for each fixed $v\in S^{p-1}$.  Define
\[
Y_{ij}(v):=\hat A_{ij}\ip{Z_{ij},v}^2, \quad 1 \le i, j \le n,
\]
and note that
\[
    v^\top (E\frac{\hat L}{n}E^\top) v = \frac{2\sigma^2}{n}\sum_{a=1}^{q_n} \sum_{\{i,j\} \in \mathcal{C}_a} Y_{ij}(v).
\]
Applying Lemma~\ref{wb:thm:bernstein} for each
block, we can obtain a strong high probability bound for \[\left|\sum_{\{i,j\}\in\mathcal C_a}Y_{ij}(v) \right|.\] Then, a union bound over the blocks (at most $n$ many) and all points on the net of $\mathbb{S}^{p-1}$ described in Lemma~\ref{wb:lem:net} (containing at most $9^p$ points) gives simultaneous control of these sums.
Finally, by summing the blocks and applying Lemma~\ref{wb:lem:net} we obtain the suitable high probability bound on $\left\|E\frac{\hat L}{n}E^\top\right\|_{\op}$.

\subsection{Dimension reduction and clustering error bounds.} By combining the bounds for the signal perturbation and the transformed noise, we can give a bound for the full perturbation
\[
\Xi_0:=\norm{X\hat S-PS_0}_{\op}
=\norm{X\hat S-P_c}_{\op}
\le
\norm{P(\hat S-S_0)}_{\op}+\norm{E\hat S}_{\op}.
\]
for the transformed data (see Proposition~\ref{wb:thm:main}), which is the key technical result of the paper. For dimension reduction, we use Lemma~\ref{wb:lem:subspace-det} from standard spectral perturbation theory (e.g., see
\cite{ORVW2018}) to compare the leading left
singular space of $X\hat S$ with $\range(P_c)$.  For
clustering, we use Lemma~\ref{wb:thm:det} to bound the errors of
$k$-means. 
	
\section{Acknowledgements and AI Statement}
\subsection{AI Statement}
In this work, we have used standard generative AI tools to support our work.  The authors supplied the research directions and detailed proof ideas, and standard generative AI tools were used for exploratory simulations, literature search, calculations,  the development and simplification of parts of the proofs and typesetting. The authors, in particular, proposed the truncated estimator based on kernelised covariance matrices and related dimension reduction and clustering methods, as well as the analytical perspective based on kernel Laplacians. The authors have verified all steps, and take responsibility for the final content of this work.
\subsection{Acknowledgements}
Subhro Ghosh was supported in part by the NUS Dean’s Chair Associate Professorship E-146-00-0037-01 and the Singapore MOE grants A-8002014-00-00 and A-8003802-00-00. Soumendu Sundar Mukherjee was supported by the Prime Minister Early Career Research Grant ANRF/ECRG/2024/006704/PMS from the Anusandhan National Research Foundation, Govt.~of India.
\printbibliography

\newpage

\appendix
	
	\section{Expectation of Data Truncation Matrices and Ideal Truncation}\label{sec:propertytruncation}

The result below gives a closed-form representation of the square root of
the ideal Laplacian as a weighted sum of orthogonal projections.  This
representation both reduces the effective noise dimension and preserves
the centered signal exactly. It is the key structural fact behind the
truncation method.
	
\begin{lemma}[Ideal Laplacian decomposition]\label{wb:lem:ideal-target}
 Let $k,n,p, \Theta^*, z^*, B_a, n_a, (X,P,E), \sigma$ be as in Theorem \ref{wb:thm:dimension-reduction}. Let $(A,D,L)$ be the ideal truncation matrices for the true cluster
configuration $(\Theta^*,z^*)$.
Set
\[
q_a:=n_a^{-1/2}\one_{B_a},\quad
Q:=[q_1\ \cdots\ q_k],\quad
s:=(\sqrt{n_1},\dots,\sqrt{n_k})^\top,
\]
and
\[
\Pi_a:=\diag(\one_{B_a})-\frac1{n_a}\one_{B_a}\one_{B_a}^\top.
\]
(i) Then we have the following decomposition
\begin{equation}\label{wb:eq:ideal-L-decomp}
\frac Ln
=
Q\left(I_k-\frac1nss^\top\right)Q^\top
+
\sum_{a=1}^k\left(1-\frac{n_a}{n}\right)\Pi_a
\end{equation}
and
\begin{equation}\label{wb:eq:P-Ln-Hn}
        P\frac Ln=PH_n=P_c
\end{equation}
where
\[
H_n:=I_n-\frac1n\one_{[n]}\one_{[n]}^\top,
\]
(ii) Let $S_0=\left(\frac Ln\right)^{1/2}$, we have
\begin{equation}\label{wb:eq:S0-explicit}
S_0
=
Q\left(I_k-\frac1nss^\top\right)Q^\top
+
\sum_{a=1}^k\sqrt{1-\frac{n_a}{n}}\,\Pi_a,
\quad S_0\one_{[n]}=0,
\end{equation}
and
\begin{equation}\label{wb:eq:PS0-Pc}
PS_0=P_c.
\end{equation}
i.e., if
\[
\bar\theta:=\frac1nP\one_{[n]}=\sum_{b=1}^k\frac{n_b}{n}\theta_b^*,
\]
then the $i$th column of $PS_0$ is $\theta_{z_i^*}^*-\bar\theta$.
\end{lemma}

Using above decomposition of $\left(\frac{L}{n} \right)^{1/2}$ and  Gaussian rotational invariance, we can show that the ideal-truncation reduces the noise  scale from $\sigma\sqrt{p+n}$ to
$\sigma\sqrt{p+m}$.

\begin{lemma}[Ideal truncation reduced noise]\label{wb:lem:ideal-noise-scale}
Let $k,n,p,B_a, n_a,n_{\max}, m, \Theta^*, z^*, (X,P,E), \sigma$ be as in Theorem \ref{wb:thm:dimension-reduction}.   Let
$(A,D,L)$ be the ideal truncation matrices for the true cluster
configuration $(\Theta^*,z^*)$, and define
\[
S_0:=\left(\frac Ln\right)^{1/2}.
\]
There exist positive absolute constants
$c_{\ref{wb:lem:ideal-noise-scale}.1},\ldots,
c_{\ref{wb:lem:ideal-noise-scale}.4}$ such that the following holds.
If $p+m\le n$, then, with probability at least
$1-\exp(-c_{\ref{wb:lem:ideal-noise-scale}.1}(p+m))$,
\[
        \norm{ES_0}_{\op}\le c_{\ref{wb:lem:ideal-noise-scale}.2}\sigma\sqrt{p+m}.
\]
By contrast, we have
\[
        \E[\norm E_{\op}]\ge c_{\ref{wb:lem:ideal-noise-scale}.4}\sigma\sqrt{p+n}.
\]
\end{lemma}

\begin{remark}\label{wb:prop:effective-dimension}
    Lemma \ref{wb:lem:ideal-noise-scale} implies that, if $p+m\le n$, then, with probability at least
$       1-\exp(-c_{\ref{wb:prop:effective-dimension}.1}(p+m))$, we have
\[
        XS_0=P_c+ES_0,
        \quad
        \norm{XS_0-P_c}_{\op}
        =\norm{ES_0}_{\op}
        \le c_{\ref{wb:prop:effective-dimension}.2}\sigma\sqrt{p+m}
\]
for some absolute constants
$c_{\ref{wb:prop:effective-dimension}.1}>0$ and
$c_{\ref{wb:prop:effective-dimension}.2}>0$. In other words, the ideal transform replaces the $\sigma \sqrt{p+n}$ scale of the untransformed gaussian error $E$ by the improved effective scale $\sigma\sqrt{p+m}$.  
\end{remark}

\section{Error Induced by Fluctuation of the Data Truncation Matrices}\label{sec:errorfluctuation}

For the practical truncation based on data, $\hat L$ is random and dependent with the random error $E$. In this case, we use the following exact error decomposition:
\[
        X\hat S-P_c=P(\hat S-S_0)+E\hat S.
\]
We then bound the two terms, the signal perturbation ($\norm{P(\hat S-S_0)}_{\op}$) and the noise part ($\norm{E\hat S}_{\op}$).

\subsection{Signal perturbation}

This part controls the effect of truncation on the population
signal.  The argument has two components.  First, we transform the
matrix square roots difference into the linear matrix difference
\(P(\hat L-L)\).  Second, all same-cluster entries cancel from this
linear difference after multiplication by \(P\).  The remaining part is transformed into a sum of random matrices
(with dependence), and we divide the terms into blocks of disjoint
index pairs. And we control these by matrix
Bernstein.

The main difficulty in bounding $\norm{P(\hat S-S_0)}_{\op}$ is the
nonlinear dependence of $\hat S$ and $S_0$ on their Laplacians through
the matrix square root.  The next lemma which follows from \cite{ShmueliDrineasAvron2024} converts the nonlinear error $\hat S-S_0$ into a linear
Laplacian error.
\begin{lemma}[Square-root linearization]\label{wb:lem:signal-reduction}
Let $k,n,p, \Theta^*, z^*, P$ be as in Theorem \ref{wb:thm:dimension-reduction}. 
Let $(A,D,L)$ be the ideal truncation matrices for the true cluster
configuration $(\Theta^*,z^*)$.
Let $\hat A\in\bbR^{n\times n}$ be any symmetric matrix with
nonnegative entries, and let $\hat L:=\mathcal L(\hat A)$, where
$\mathcal L$ is the Laplacian operator of
Definition~\ref{def:degree-laplacian-operators}.  Set
\[
\hat S=(\hat L/n)^{1/2},
\quad
S_0=(L/n)^{1/2}.
\]
Then
\begin{equation}\label{wb:eq:signal-reduction-identity}
P(\hat S-S_0)
=
\frac1nP(\hat L-L)(\hat S+I_n)^{-1}.
\end{equation}
Consequently,
\begin{equation}\label{wb:eq:signal-reduction}
\norm{P(\hat S-S_0)}_{\op}
\le
\frac1n\norm{P(\hat L-L)}_{\op}.
\end{equation}
\end{lemma}

We now bound
$\frac1n\norm{P(\hat L-L)}_{\op}$ directly by writing it as a sum of
random matrices.  To deal with the dependence of the terms, we divide
them into blocks of disjoint index pairs. 

\begin{lemma}[Bound for linearized signal perturbation]
\label{wb:lem:distance-cross-signal}
Let \[k,n,p,B_a, n_a,n_{\max}, m, \Theta^*, z^*, (X,P,E), \sigma, \Delta, \phi, (\hat{A}, \hat{D}, \hat{L})\] be as in Theorem \ref{wb:thm:dimension-reduction}. Let $(A,D,L)$ be the
ideal truncation matrices for the true cluster configuration
$(\Theta^*,z^*)$.
There exist positive absolute constants
$c_{\ref{wb:lem:distance-cross-signal}.1},\ldots,
c_{\ref{wb:lem:distance-cross-signal}.4}$ such that the following
holds.  Assume
\[
\kappa_\Delta:=\frac{\Delta^4/\sigma^4}{p+\Delta^2/\sigma^2}
\ge c_{\ref{wb:lem:distance-cross-signal}.1}.
\]
Then, with probability at least
\[
1-\exp(-c_{\ref{wb:lem:distance-cross-signal}.2}(p+m)),
\]
\begin{equation}\label{wb:eq:distance-cross-signal}
\frac1n\norm{P(\hat L-L)}_{\op}
\le
c_{\ref{wb:lem:distance-cross-signal}.3}\Delta
\sqrt{
m\exp\left(-c_{\ref{wb:lem:distance-cross-signal}.4}
\frac{\Delta^4/\sigma^4}{p+\Delta^2/\sigma^2}\right)
+(p+m+\log n)\left(\frac{\sigma^2}{\Delta^2}
+\frac{p\sigma^4}{\Delta^4}\right)
}.
\end{equation}
More generally, for every $t>0$, with probability at least $1-e^{-t}$,
\begin{equation}\label{wb:eq:matching-signal-confidence}
\frac1n\norm{P(\hat L-L)}_{\op}
\le c_{\ref{wb:lem:distance-cross-signal}.3}\Delta
\sqrt{
m e^{-c_{\ref{wb:lem:distance-cross-signal}.4}\kappa_\Delta}
+\frac{t+\log(np)}{\kappa_\Delta}}.
\end{equation} 
\end{lemma}

The preceding lemma controls the linear matrix difference.  The next theorem
transfers this estimate to the original square-root signal term through
Lemma~\ref{wb:lem:signal-reduction}. 

\begin{proposition}[Signal perturbation bound]
\label{wb:thm:signal}
Let 
\[
    k,n,p,B_a, n_a,n_{\max}, m, \Theta^*, z^*, (X,P,E), \sigma, \Delta, \phi, (\hat{A}, \hat{D}, \hat{L}), \hat{S}
\]
be as in Theorem \ref{wb:thm:dimension-reduction}.  Let $(A,D,L)$ be the
ideal truncation matrices for the true cluster configuration
$(\Theta^*,z^*)$, and let
\[
S_0=(L/n)^{1/2}.
\]
There exist positive absolute constants
$c_{\ref{wb:thm:signal}.1},\ldots,c_{\ref{wb:thm:signal}.4}$ such
that the following holds.  Assume
\begin{equation}\label{wb:eq:signal-basic-condition}
\frac{\Delta^4/\sigma^4}{p+\Delta^2/\sigma^2}
\ge c_{\ref{wb:thm:signal}.1}.
\end{equation}
Then, with probability at least
\[
1-\exp(-c_{\ref{wb:thm:signal}.2}(p+m)),
\]
\begin{equation}\label{wb:eq:signal-simple-bound}
\norm{P(\hat S-S_0)}_{\op}
\le
c_{\ref{wb:thm:signal}.3}\Delta
\sqrt{
m\exp\left(-c_{\ref{wb:thm:signal}.4}
\frac{\Delta^4/\sigma^4}{p+\Delta^2/\sigma^2}\right)
+(p+m+\log n)\left(\frac{\sigma^2}{\Delta^2}
+\frac{p\sigma^4}{\Delta^4}\right)
}.
\end{equation}
More generally, for every $t>0$, the bound
\eqref{wb:eq:signal-simple-bound} holds with probability at least
$1-e^{-t}$ when the factor $p+m+\log n$ on its right-hand side is replaced
by $t+\log(np)$.

\end{proposition}

\begin{proof}
By Lemma~\ref{wb:lem:signal-reduction},
\[
\norm{P(\hat S-S_0)}_{\op}
\le
\frac1n\norm{P(\hat L-L)}_{\op}.
\]
Lemma~\ref{wb:lem:distance-cross-signal} supplies both bounds.
Relabeling its positive absolute constants gives the four constants in
the present statement.
\end{proof}

The cancellation used above is specific to the population signal:
\(P(e_i-e_j)=0\) whenever \(z_i^*=z_j^*\).  It does not hold for the
noise matrix because
\(E(e_i-e_j)=\varepsilon_i-\varepsilon_j\) is generally nonzero.  The
noise term therefore requires a separate argument that controls both
 cross-cluster entries and same-cluster false-positive entries
while preserving their dependence on the observations.

\subsection{Contribution from the noise}

The aim of this subsection is to control $E\hat S$ without ever treating
the data-defined transform $\hat S$ as independent of the Gaussian noise
matrix $E$.  We use the pairwise Laplacian identity and partition all
pairs into blocks with disjoint indices.  For each fixed direction,
the summands within a block are independent, so we can apply the scalar
Bernstein inequality in Lemma~\ref{wb:thm:bernstein}.  A union bound
over the blocks and a sphere net then gives the following bound.
Independence between blocks is not required.

\begin{proposition}[Transformed noise bound]
\label{wb:thm:clean-cap}
Let \[k,n,p,B_a, n_a,n_{\max}, m, \Theta^*, z^*, (X,P,E), \sigma,\hat S\] be as in Theorem \ref{wb:thm:dimension-reduction}.
There exist positive absolute constants
$c_{\ref{wb:thm:clean-cap}.1},\ldots,c_{\ref{wb:thm:clean-cap}.3}$
such that the following holds. Assume that
\begin{equation}\label{eq:kappa_delta_lower_bound}
\frac{\Delta^4/\sigma^4}{p+\Delta^2/\sigma^2}
\ge
c_{\ref{wb:thm:clean-cap}.1} (1+\log_+(n/(p+m))).
\end{equation}
Then
\[
\Pb\left(
\norm{E\hat S}_{\op}
\le c_{\ref{wb:thm:clean-cap}.2}\sigma\sqrt{p+m+\log n}
\right)
\ge
1-e^{-c_{\ref{wb:thm:clean-cap}.3}(p+m)}.
\]
\end{proposition}

\subsection{Final perturbation bound}

Recall the decomposition
\[
X\hat S-P_c=X\hat S-PS_0
=
P(\hat S-S_0)+E\hat S.
\]
Therefore, combining previous analysis, we can obtain a bound for $\|X\hat S-P_c\|_\op$.

\begin{proposition}[Perturbation bound of transformed data]\label{wb:thm:main}
Let \[k,n,p,B_a, n_a,n_{\max}, m, \Theta^*, z^*, (X,P,E), \sigma,\hat S\] be as in Theorem \ref{wb:thm:dimension-reduction}. Let \[\Xi_0:=\norm{X\hat S-PS_0}_{\op}.\]

There exist positive absolute constants
$c_{\ref{wb:thm:main}.1},\ldots,c_{\ref{wb:thm:main}.6}$ such that the
following holds.  Let
\[
h:=1+\log_+\frac{n}{p+m}.
\]
If
\[
\frac{\Delta^2}{\sigma^2}
\ge
c_{\ref{wb:thm:main}.1}
\left[
h+\sqrt{ph}
\right],
\]
then, with probability at least
\[
1-\exp(-c_{\ref{wb:thm:main}.2}(p+m)),
\]
we have
\[
\Xi_0
\le
c_{\ref{wb:thm:main}.3}\left[
\sigma\sqrt{p+m+\log n}
+
\Delta
\sqrt{
m e^{-c_{\ref{wb:thm:main}.4}\frac{\Delta^4/\sigma^4}{p+\Delta^2/\sigma^2}}
+
(p+m+\log n)\left(\frac{\sigma^2}{\Delta^2}
+\frac{p\sigma^4}{\Delta^4}\right)
}
\right].
\]

\end{proposition}

\begin{proof}
The identity
\[
X\hat S-PS_0=P(\hat S-S_0)+E\hat S
\]
follows from $X=P+E$.  Put
\[
u:=\frac{\Delta^2}{\sigma^2},
\quad h:=1+\log_+\frac{n}{p+m},
\quad \kappa_\Delta:=\frac{u^2}{p+u}.
\]
Since $h\ge1$, applying \eqref{wb:eq:kappa-explicit-implication}
with $T=c_1h$, for a sufficiently large positive absolute constant
$c_1$, shows that the separation hypothesis implies both the lower
bound on $\kappa_\Delta$ in Proposition~\ref{wb:thm:signal} and
$\kappa_\Delta\ge128h$.  Apply
\eqref{wb:eq:matching-signal-confidence} with $t=2(p+m)$ and use
Lemma~\ref{wb:lem:signal-reduction}.  Since $\log p\le p$,
\[
2(p+m)+\log(np)\le3(p+m+\log n).
\]
Thus, with probability at least $1-e^{-2(p+m)}$,
\[
\norm{P(\hat S-S_0)}_{\op}
\le
c_2\Delta
\sqrt{
m e^{-c_3u^2/(p+u)}
+(p+m+\log n)\frac{p+u}{u^2}}.
\]
Also, $\kappa_\Delta\ge128h$ gives
$n e^{-\kappa_\Delta/128}\le(p+m)/2$.  Applying
\eqref{wb:eq:matching-noise-covariance-bound} with $t=2(p+m)$ and
using $\norm{E\hat S}_{\op}^2=\norm{E\hat L E^\top/n}_{\op}$,
we obtain, with probability at least $1-e^{-2(p+m)}$,
\[
\norm{E\hat S}_{\op}\le c_4\sigma\sqrt{p+m+\log n}.
\]
Since $p+m\ge1$, the union of the two failure events has probability
at most
\[
2e^{-2(p+m)}\le e^{-(p+m)}.
\]
The triangle inequality now gives the first assertion, with positive
absolute constants $c_5,c_6$, after using
\[
\frac{p+u}{u^2}=\frac{\sigma^2}{\Delta^2}
+\frac{p\sigma^4}{\Delta^4}.
\]

\end{proof}

\section{Proof of Main Results on Dimension reduction}

In this section, we prove our main result on dimension reduction, i.e., Theorem~\ref{wb:thm:dimension-reduction}. We use the following standard result from spectral perturbation theory.

\begin{lemma}[Singular-subspace perturbation]\label{wb:lem:subspace-det}
Let $p,n\ge1$, let $r$ be an integer satisfying $1\le r\le\min\{p,n\}$, and let $M\in\bbR^{p\times n}$ have rank $r$. Let $U\in\bbR^{p\times r}$ be an orthonormal basis for its column space. Let $Y\in\bbR^{p\times n}$, and let $\hat U\in\bbR^{p\times r}$ be the matrix of the top $r$ left singular vectors of $Y$.  In this lemma, extend the singular-value sequence of every $p\times n$ matrix by setting $\sigma_j=0$ for $j>\min\{p,n\}$.  Then $\sigma_r(M)>0$, and
\[
\norm{\hat U\hat U^\top-UU^\top}_{\op}
\le
\frac{2\norm{Y-M}_{\op}}{\sigma_r(M)}.
\]
\end{lemma}

\begin{proof}
This follows directly from \cite[Theorem~19]{ORVW2018}.
\end{proof}

With the standard tool above, we can now prove Theorem~\ref{wb:thm:dimension-reduction}.

\begin{proof}[Proof of Theorem~\ref{wb:thm:dimension-reduction}]
By Lemma~\ref{wb:lem:ideal-target}, we have $PS_0=P_c$. So we have
\[
\norm{Y-P_c}_{\op}=\norm{X\hat S-PS_0}_{\op}.
\]
The separation hypothesis implies the corresponding requirements of
Proposition~\ref{wb:thm:main}, after choosing the appropriate constants.  Therefore, with probability at least $1-e^{-c_1(p+m)}$, we have
\begin{align*}&\norm{Y-P_c}_{\op}\\=&\norm{X\hat S-PS_0}_{\op}\\
\le &c_2\left[
\sigma\sqrt{p+m+\log n}+
\Delta\sqrt{
m\exp\left(-c_3\frac{\Delta^4/\sigma^4}
{p+\Delta^2/\sigma^2}\right)
+(p+m+\log n)\left(\frac{\sigma^2}{\Delta^2}
+\frac{p\sigma^4}{\Delta^4}\right)}
\right]
\end{align*}
for positive absolute constants $c_2,c_3$.  Applying
Lemma~\ref{wb:lem:subspace-det} with $M=P_c$ and $r=r_0$, we obtain
\[
\norm{\hat U_{r_0}\hat U_{r_0}^\top-U_0U_0^\top}_{\op}
\le
\frac{2\norm{Y-P_c}_{\op}}{\sigma_{r_0}(P_c)}
\]
which proves the desired result after plugging in the bound for $\norm{Y-P_c}_{\op}$.
\end{proof}

\section{Proof of Main Results on Clustering Accuracy}

The ideal target $M=PS_0=P_c$ is cluster-wise constant, preserves the
original pairwise center separations, and has rank at most
$r_\ast=\min\{k-1,p\}$.  Thus the deterministic spectral $k$-means
theorem applies once the perturbation $\Xi_0$ is small relative to the
minimum separation and the smallest cluster size.  The following proof
shows that the explicit separation condition in
Theorem~\ref{wb:thm:weak-clustering} implies exactly that deterministic
margin, then substitutes the perturbation bound from
Proposition~\ref{wb:thm:main}.

\begin{proof}[Proof of Theorem~\ref{wb:thm:weak-clustering}]
For convenience, we introduce the notation
\[
u:=\frac{\Delta^2}{\sigma^2}, \quad h:=1+\log_+\frac{n}{p+m},
\quad
T:=h+\frac{r_\ast(p+m+\log n)}{n_{\min}},
\quad
\kappa:=\frac{u^2}{p+u}.
\]
By Proposition~\ref{wb:thm:main} there exist positive absolute constants
$c_1,\ldots,c_6$ such that if
\[
u\ge c_1\left[h+\sqrt{ph}\right],
\]
then, with probability at least $1-e^{-c_2(p+m)}$, we have
\begin{equation}\label{wb:eq:weak-proof-xi}
\Xi_0\le c_3\left[
\sigma\sqrt{p+m+\log n}+
\Delta\sqrt{m e^{-c_4\kappa}+\frac{p+m+\log n}{\kappa}}
\right].
\end{equation}

Let
\[
c_7:=\max\{1,(54c_3)^2\},
\quad
c_8:=\max\{1,(c_4e)^{-1}\},
\quad
c_{9}:=c_7c_8,
\]
and
\[
c_{10}:=\max\{c_7,c_{9},\sqrt{c_{9}},c_1\}.
\]
We now require the separation
\begin{equation}\label{eq: separaproof}
u=\frac{\Delta^2}{\sigma^2}
\ge
c_{10}
\left\{
\left[
h+
\frac{r_\ast(p+m+\log n)}{n_{\min}}
\right]
+
\sqrt{
p\left[
h+
\frac{r_\ast(p+m+\log n)}{n_{\min}}
\right]
}
\right\}.
\end{equation}
Since $T\ge h$, this choice implies the separation hypothesis of
Proposition~\ref{wb:thm:main}.  The separation requirement \eqref{eq: separaproof} implies
\begin{equation}\label{wb:eq:km-kappa-large-1}
u\ge c_7T.
\end{equation} And using
\eqref{wb:eq:kappa-explicit-implication} with $c_{9}T$, we have
\begin{equation}\label{wb:eq:km-kappa-large-2}
\kappa\ge c_{9}T.
\end{equation}
Thus \eqref{wb:eq:weak-proof-xi} holds outside an event of probability
at most $e^{-c_2(p+m)}$.

Define the ideal truncation matrices and transforms
\[
A_{ij}:=\mathbbm{1}_{\{z_i^*\ne z_j^*\}},
\quad
L:=\diag(A\one_{[n]})-A,
\quad
S_0:=\left(\frac Ln\right)^{1/2},
\quad
M:=PS_0.
\]
Recall from Lemma~\ref{wb:lem:ideal-target} that
\[
M=P-\frac1n(P\one_{[n]})\one_{[n]}^\top.
\]
Therefore the columns of $M_{\cdot,j}$ are the same on $j \in B_a$ for each $a$, $\rank(M)\le r$, and the distinct columns are the centered cluster centers, i.e.,
\[
\mu_a^0:=\theta_a^*-\frac1nP\one_{[n]},\quad a\in[k],
\]
which imply
\begin{equation}\label{wb:eq:ideal-sep-proof}
\delta_0:=\min_{a\ne b}\|\mu_a^0-\mu_b^0\|_2
=\min_{a\ne b}\|\theta_a^*-\theta_b^*\|_2\ge\Delta.
\end{equation}

We next verify that the separation requirement \eqref{wb:eq:det-cond} is satisfied, i.e., we want to show
\[
18\Xi_0\sqrt{\frac {r_\ast} {n_{\min}}}\le\delta_0.
\]
Using the fact $T \ge \frac{r_*(p+m +\log n)}{n_{\min}}$ and
the inequality
\eqref{wb:eq:km-kappa-large-1}, we obtain
\begin{equation}\label{wb:eq:margin-pop-noise}
18c_3\sigma\sqrt{p+m+\log n}\sqrt{\frac {r_\ast} {n_{\min}}}\le\frac\Delta3.
\end{equation}
The inequality \eqref{wb:eq:km-kappa-large-2}, the fact
$c_{9}\ge c_7$, and $T \ge \frac{r_*(p+m +\log n)}{n_{\min}}$ give
\begin{equation}\label{wb:eq:margin-kappa-cost}
\frac{p+m+\log n}{\kappa}\le\frac{n_{\min}}{c_7r_\ast}.
\end{equation}
Finally, $xe^{-x}\le e^{-1}$ for $x\ge0$, so
$e^{-c_4\kappa}\le(c_4e\kappa)^{-1}\le c_8/\kappa$.
Using $m\le p+m$, $c_{9}=c_7c_8$, and \eqref{wb:eq:margin-kappa-cost}, we obtain
\begin{equation}\label{wb:eq:margin-signal-exp}
m e^{-c_4\kappa}
\le\frac{c_8(p+m)}{\kappa}
\le\frac{n_{\min}}{c_7r_\ast}.
\end{equation}
Using $\sqrt{x+y}\le\sqrt x+\sqrt y$, and combining
\eqref{wb:eq:weak-proof-xi}--\eqref{wb:eq:margin-signal-exp}, we have
\[
18\Xi_0\sqrt{\frac {r_\ast} {n_{\min}}}\le\Delta\le\delta_0.
\]

Therefore, we can apply Lemma~\ref{wb:thm:det} to $Y$ and $M$ and obtain
\[
\ell(\hat z,z^*)\le\frac{288r\Xi_0^2}{n\delta_0^2},
\quad
\ell_\infty(\hat z,z^*)\le
\frac{288r_\ast\Xi_0^2}{n_{\min}\delta_0^2}.
\]
Using $(a+b)^2\le2a^2+2b^2$ in
\eqref{wb:eq:weak-proof-xi} and
\[
\frac1\kappa=\frac{\sigma^2}{\Delta^2}
+\frac{p\sigma^4}{\Delta^4},
\]
we prove the two loss bounds.

\end{proof}

	\section{Preliminaries}

    For many of the proofs we use the quantity
\begin{equation}\label{wb:eq:kappa}
        \kappa_\Delta:=\frac{\Delta^4/\sigma^4}{p+\Delta^2/\sigma^2}
\end{equation}
We have the following elementary result that converts lower bounds on \(\Delta^2/\sigma^2\) into lower bounds on \(\kappa_\Delta\).
\begin{lemma} Let $\Delta>0,\sigma>0,p>0$ and 
    \begin{equation*}
        \kappa_\Delta:=\frac{\Delta^4/\sigma^4}{p+\Delta^2/\sigma^2}
\end{equation*}
For every \(T>0\) with
\begin{equation*}
\frac{\Delta^2}{\sigma^2}
\ge
T+\sqrt{pT}
\end{equation*}
we have
\begin{equation}\label{wb:eq:kappa-explicit-implication}
\kappa_\Delta\ge T
\end{equation}
\end{lemma}
\begin{proof}
For convenience, we use the notation $u:=\Delta^2/\sigma^2$.  The conclusion is equivalent to
\[
u^2\ge T(p+u).
\]
At $u=T+\sqrt{pT}$, we calculate
\[
u^2-Tu-pT=T\sqrt{pT}\ge0.
\]
Since $u\mapsto u^2-Tu-pT$ is increasing for
$u\ge T+\sqrt{pT}$, we know that $u
\ge
T+\sqrt{pT}$ implies $u^2\ge T(p+u)$.
\end{proof}
	
	\begin{lemma}[Principal PSD square root]\label{lem:sqrt}
		Let $n\ge1$ be an integer and let $A\in\bbR^{n\times n}$ be symmetric and positive semidefinite. Then there exists a unique symmetric positive semidefinite matrix $A^{1/2}$ such that $(A^{1/2})^2=A$.
	\end{lemma}
	
	\begin{proof}
			Existence follows from the spectral theorem for real symmetric matrices (see, e.g., \cite[Corollary~2.5.11(a)]{HornJohnsonMA}). There exists an orthogonal matrix $U\in\bbR^{n\times n}$ and a diagonal matrix $\Lambda=\diag(\lambda_1,\dots,\lambda_n)$ such that
		\[
		A = U\Lambda U^\top,
		\quad \lambda_i\ge 0\ \text{for all }i
		\]
		(the nonnegativity follows from $A\succeq 0$). Define $\Lambda^{1/2}:=\diag(\sqrt{\lambda_1},\dots,\sqrt{\lambda_n})$ and set
		\[
		B := U\Lambda^{1/2}U^\top.
		\]
		Then $B$ is symmetric, and for any $x\in\bbR^n$,
		\[
		x^\top B x = (U^\top x)^\top \Lambda^{1/2} (U^\top x)\ge 0,
		\]
		so $B\succeq 0$. Finally,
		\[
		B^2 = U\Lambda^{1/2}U^\top U\Lambda^{1/2}U^\top = U(\Lambda^{1/2})^2 U^\top = U\Lambda U^\top = A.
		\]
		Thus $B$ is a symmetric PSD square root of $A$.
		
		For uniqueness, suppose $C\succeq 0$ is symmetric and $C^2=A$. Let $v$ be an eigenvector of $A$ with eigenvalue $\lambda\ge 0$, so $Av=\lambda v$. Then
		\[
		A(Cv)=C^2(Cv)=C(Av)=C(\lambda v)=\lambda (Cv).
		\]
		Thus $Cv$ lies in the $\lambda$-eigenspace of $A$, which shows that every eigenspace of $A$ is invariant under $C$. Restrict $C$ to a $\lambda$-eigenspace. On that subspace, $A=\lambda I$, so $C^2=\lambda I$. Therefore every eigenvalue of the restriction of $C$ solves $t^2=\lambda$, hence equals $\pm\sqrt{\lambda}$. Since $C\succeq 0$, all eigenvalues of $C$ are nonnegative, so they must in fact equal $\sqrt{\lambda}$. Therefore $C$ acts as $\sqrt{\lambda}I$ on the $\lambda$-eigenspace.
		
		Since this holds for every eigenspace and these eigenspaces span $\bbR^n$, the matrix $C$ is uniquely determined and coincides with the $B$ constructed above. Hence the symmetric PSD square root is unique.
	\end{proof}

\begin{lemma}[Best rank-$r$ approximation in operator norm]\label{wb:lem:best-r}
Let $p,n\ge1$, let $Y\in\bbR^{p\times n}$, and let $r$ be an integer satisfying
\[
1\le r\le\min\{p,n\}.
\]
Let
\[
Y_r:=\sum_{j=1}^r\sigma_j(Y)u_jv_j^\top
\]
be the rank-$r$ truncated singular value decomposition of $Y$. If $M\in\bbR^{p\times n}$ satisfies $\rank(M)\le r$, then
\[
\norm{Y-Y_r}_{\op}\le \norm{Y-M}_{\op}.
\]
\end{lemma}

\begin{proof}
If $\rank(Y)\le r$, then $Y_r=Y$ and the conclusion is immediate.
Otherwise $r<\rank(Y)$, and the operator-norm
Eckart--Young--Mirsky theorem
\cite[Section~4.1.4]{Vershynin2018} gives
\[
\norm{Y-Y_r}_{\op}
=\min_{\rank(B)\le r}\norm{Y-B}_{\op}.
\]
Since $\rank(M)\le r$, taking $B=M$ proves the claim.
\end{proof}

\begin{lemma}[Rank-one expansion of Laplacian]
\label{wb:lem:pair-matrix-expansion}
Let \(p,n\ge1\) be integers, let \(H\in\bbR^{n\times n}\) be symmetric with zero
diagonal, and let $\mathcal L$ be the Laplacian operator of
Definition~\ref{def:degree-laplacian-operators}.
Then
\[
\mathcal L(H)
=
\sum_{1\le i<j\le n}
H_{ij}(e_i-e_j)(e_i-e_j)^\top.
\]
Consequently, for every
\(V=[v_1,\ldots,v_n]\in\bbR^{p\times n}\),
\[
V\mathcal L(H)V^\top
=
\sum_{1\le i<j\le n}
H_{ij}(v_i-v_j)(v_i-v_j)^\top.
\]
\end{lemma}

\begin{proof}
The diagonal and off-diagonal entries of the first matrix on the
right are, respectively,
\(\sum_{j\ne i}H_{ij}\) and \(-H_{ij}\).  These are precisely the
entries of \(\diag(H\one_{[n]})-H=\mathcal L(H)\).  Multiplication on
the left by \(V\) and on the right by \(V^\top\) proves the second
identity.
\end{proof}

\section{Partition into classes of disjoint pairs}

For a finite set \(V\), write
\[
\mathcal P_2(V)
:=
\bigl\{\{x,y\}:x,y\in V,\ x\ne y\bigr\}.
\]
We need the following lemma, which is a reformulation of standard edge-partition results (see
\cite[Proposition~5.3.1 and Theorem~5.3.2]{Diestel2017}). For completeness, we include the proof here.

\begin{lemma}[Partition into classes of disjoint pairs]
\label{wb:lem:disjoint-pair-partition}
Let \(V\) be a finite set with \(|V|=N\ge1\), set
\[
\mathcal P_2(V)
:=
\bigl\{\{x,y\}:x,y\in V,\ x\ne y\bigr\},
\quad
q_N
:=
\begin{cases}
N-1, & \text{if \(N\) is even},\\
N,   & \text{if \(N\) is odd}.
\end{cases}
\]
There exist subsets
\(\mathcal C_1,\ldots,\mathcal C_{q_N}\) of \(\mathcal P_2(V)\) such
that
\[
\mathcal P_2(V)
=
\bigsqcup_{a=1}^{q_N}\mathcal C_a,
\quad
|\mathcal C_a|
=
\left\lfloor\frac N2\right\rfloor,
\]
and
\[
C\cap C'=\varnothing
\quad
\text{for all distinct }C,C'\in\mathcal C_a.
\]

Suppose, in addition, that there is an integer $k\ge1$ for which
\[
V=B_1\sqcup\cdots\sqcup B_k,
\quad
S:=V\setminus B_1,
\quad
m:=|S|,
\]
and define
\[
\mathcal I
:=
\left\{
\{i,j\}\in\mathcal P_2(V):
i\in B_r,\ j\in B_s
\text{ for some }r\ne s
\right\}.
\]
Then
\[
\mathcal I_a:=\mathcal C_a\cap\mathcal I,
\quad 1\le a\le q_N,
\]
partition \(\mathcal I\), still satisfying the disjoint-index property, and
satisfy
\[
|\mathcal I_a|\le m
\quad
\text{for every }a.
\]
\end{lemma}

\begin{proof}
Suppose first that \(N=2d\) is even.  Identify \(V\) with
\[
\{\infty\}\cup\mathbb Z/(2d-1)\mathbb Z.
\]
For \(t\in\mathbb Z/(2d-1)\mathbb Z\), define
\[
\mathcal C_t
:=
\bigl\{\{\infty,t\}\bigr\}
\cup
\bigl\{
\{t+x,t-x\}:x=1,\ldots,d-1
\bigr\},
\]
where all arithmetic is modulo \(2d-1\).  Since
\[
\{\pm1,\ldots,\pm(d-1)\}
=
(\mathbb Z/(2d-1)\mathbb Z)\setminus\{0\},
\]
we conclude that each element of \(V\) appears in exactly one pair in
\(\mathcal C_t\).  Therefore the members of \(\mathcal C_t\) are pairwise
disjoint and
\[
|\mathcal C_t|=d=\frac N2.
\]

We verify that these classes partition \(\mathcal P_2(V)\).  A pair
\(\{\infty,u\}\) belongs only to \(\mathcal C_u\).  If $u\ne v$ belong to $\mathbb Z/(2d-1)\mathbb Z$, then
\[
\gcd(2,2d-1)=1,
\]
so the residue class of $2$ is a unit in
$\mathbb Z/(2d-1)\mathbb Z$
(see, e.g., \cite{DummitFoote2004}). In fact, its inverse is the residue
class of $d$, since
\[
2d\equiv1\pmod{2d-1}.
\]
As a result, multiplication by $2$ is bijective, and therefore
\[
t:=\frac{u+v}{2},
\quad
x:=\frac{u-v}{2}
\]
are uniquely determined.  Also,
\[
x:=\frac{u-v}{2}
\]
is nonzero, and exactly one of \(x\) and \(-x\) has a representative in
\(\{1,\ldots,d-1\}\).  So,
\[
\{u,v\}=\{t+x,t-x\}
\]
belongs to exactly one class.  This proves the desired result when \(N\) is
even.

If \(N\) is odd, add an extra element
\(v_{\rm extra}\notin V\) to $V$ and apply the even construction to
\(V\cup\{v_{\rm extra}\}\), which has \(N+1\) elements.  It produces \(N\)
classes, each of which partitions \(V\cup\{v_{\rm extra}\}\) into pairs and
therefore contains exactly one pair containing \(v_{\rm extra}\).  Removing
that pair from every class leaves a partition of
\(\mathcal P_2(V)\) into \(N\) classes, each containing
\[
\frac{N+1}{2}-1
=
\frac{N-1}{2}
=
\left\lfloor\frac N2\right\rfloor
\]
pairwise disjoint pairs.  This argument also covers \(N=1\).

For the additional claim, observe that every \(C\in\mathcal I\) satisfies
\(C\cap S\ne\varnothing\), because otherwise \(C\subseteq B_1\), contrary to
the definition of \(\mathcal I\).  For fixed \(a\), the nonempty sets
\[
\{C\cap S:C\in\mathcal I_a\}
\]
are pairwise disjoint.  Therefore, we conclude
\[
|\mathcal I_a|
\le
\sum_{C\in\mathcal I_a}|C\cap S|
=
\left|
\bigcup_{C\in\mathcal I_a}(C\cap S)
\right|
\le |S|=m.
\]
\end{proof}

\section{Spectral k-means}

We use the following standard result on $k$-means algorithm to bound the clustering error by a perturbation bound.

\begin{lemma}[Spectral $k$-means theorem]\label{wb:thm:det}
Let $p,n\ge1$ and $k\ge2$ be integers, and let
$M\in\bbR^{p\times n}$ satisfy $\rank(M)\le r$ for some integer $r$ with
\[
1\le r\le\min\{p,n\}.
\]
Assume that there exist labels $z^\ast\in[k]^n$ and centers $\mu_1,\dots,\mu_k\in\bbR^p$ such that
\[
M_{\cdot,i}=\mu_{z_i^\ast},
\quad i\in[n].
\]
Let
\[
B_a:=\{i\in[n]:z_i^\ast=a\},
\quad
n_a:=|B_a|,
\quad
n_{\min}:=\min_{1\le a\le k}n_a,
\quad
\delta:=\min_{a\neq b}\norm{\mu_a-\mu_b}_2.
\]
Assume that every true cluster is nonempty, i.e., $n_{\min}\ge1$ and that the population centers are pairwise distinct, i.e., $\delta>0$.
Let $Y\in\bbR^{p\times n}$ be arbitrary. Let $\hat U_r$ be a matrix
whose columns are $r$ left singular vectors of $Y$ associated with its
$r$ largest singular values, and let
\[
\hat P:=\hat U_r^\top Y\in\bbR^{r\times n}.
\]
Let $\hat z\in[k]^n$ be any label vector for which there exist
$\hat\theta_1,\ldots,\hat\theta_k\in\bbR^r$ satisfying
\[
(\hat z,\hat\theta_1,\ldots,\hat\theta_k)
\in
\argmin_{\substack{z\in[k]^n\\
                    \theta_1,\ldots,\theta_k\in\bbR^r}}
\sum_{i=1}^n\norm{\hat P_{\cdot,i}-\theta_{z_i}}_2^2.
\]
For $z\in[k]^n$, define
\[
\ell(z,z^\ast)
:=
\min_{\pi\in\mathfrak S_k}
\frac1n\sum_{i=1}^n\mathbf 1\{\pi(z_i)\ne z_i^\ast\},
\]
and
\[
\ell_\infty(z,z^\ast)
:=
\min_{\pi\in\mathfrak S_k}
\max_{a\in[k]}
\frac1{n_a}\sum_{i\in B_a}\mathbf 1\{\pi(z_i)\ne a\},
\]
where $\mathfrak S_k$ is the set of permutations of $[k]$. Set
\[
\xi:=\norm{Y-M}_{\op}.
\]
If
\begin{equation}\label{wb:eq:det-cond}
\delta\ge 18\,\xi\sqrt{\frac r{n_{\min}}},
\end{equation}
then
\begin{equation}\label{wb:eq:det-loss}
\ell(\hat z,z^\ast)
\le
\frac{288\,r\,\xi^2}{n\delta^2},
\quad
\ell_\infty(\hat z,z^\ast)
\le
\frac{288\,r\,\xi^2}{n_{\min}\delta^2}.
\end{equation}
\end{lemma}

\begin{proof}
Let $\Pi:=\hat U_r\hat U_r^\top$ and set $Y_r:=\Pi Y$.  By the
choice of $\hat U_r$, this is a best rank-$r$ singular-value truncation
of $Y$.  By Lemma~\ref{wb:lem:best-r}, we have
\[
\norm{Y-Y_r}_{\op}\le\xi,
\]and by triangle inequality, we have\[
\norm{Y_r-M}_{\op}\le2\xi.
\]
Since $\rank(Y_r-M)\le2r$, we have
\[
\norm{Y_r-M}_{\mathrm F}\le2\sqrt{2r}\,\xi.
\]
Since $\hat U_r^\top\hat U_r=I_r$ and $\hat P=\hat U_r^\top Y$, we have
\[
Y_r=\hat U_r\hat P.
\]
For every $i\in[n]$ and every $\theta\in\bbR^r$, the isometry
$\theta\mapsto\hat U_r\theta$ gives
\[
\norm{\hat P_{\cdot,i}-\theta}_2^2
=
\norm{\hat U_r(\hat P_{\cdot,i}-\theta)}_2^2
=
\norm{(Y_r)_{\cdot,i}-\hat U_r\theta}_2^2.
\]
This map is a bijection from $\bbR^r$ onto $\range(\hat U_r)$.
Therefore global $k$-means on the columns of $\hat P$ is equivalent to
global $k$-means on the columns of $Y_r$ with centers constrained to lie
in $\range(\hat U_r)$, i.e.,

\begin{align}\label{eq:kmean1}
{\operatornamewithlimits{minimize}_{z\in[k]^n, \theta_1,\dots,\theta_k\in\mathbb{R}^r}}\sum_{i=1}^n \bigl\| \widehat P_{\cdot,i} - \theta_{z_i} \bigr\|_2^2
	\end{align}
is equivalent to
\begin{align}\label{eq:kmean2}
{\operatornamewithlimits{minimize}_{z\in[k]^n, \zeta_1,\dots,\zeta_k\in \range(\hat U_r)}}\sum_{i=1}^n \bigl\| ( {Y_r})_{\cdot,i} - \zeta_{z_i} \bigr\|_2^2.
	\end{align}
Let $\hat C\in\bbR^{p\times n}$ be the fitted centers from \eqref{eq:kmean2}, and write
\[
\hat C_{\cdot,i}:=\hat\zeta_{\hat z_i}=\hat U_r\hat\theta_{\hat z_i},
\quad i\in[n].
\]
Let $\Pi=\hat U_r\hat U_r^\top$ and define the projected population center matrix
$\widetilde M:=\Pi M$. Observe that $\widetilde M$ is feasible in \eqref{eq:kmean2}.
Since \(\Pi Y_r=Y_r\), we obtain
\[
\widetilde M-Y_r=\Pi(M-Y_r).
\]
By optimality of $\hat C$ in \eqref{eq:kmean2}, we have
\[
\norm{\hat C-Y_r}_{\mathrm F}^2=\sum_{i=1}^n \bigl\| ( {Y_r})_{\cdot,i} - \hat C_{\cdot,i} \bigr\|_2^2
\le \sum_{i=1}^n \bigl\| ( {Y_r})_{\cdot,i} - \widetilde M_{\cdot,i} \bigr\|_2^2=
\norm{\widetilde M-Y_r}_{\mathrm F},
\]
and combining this with the fact that orthogonal projection is contractive in Frobenius norm, we have
\[
\norm{\hat C-Y_r}_{\mathrm F} \le
\norm{\widetilde M-Y_r}_{\mathrm F}
\le
\norm{M-Y_r}_{\mathrm F},
\]
Also by triangle inequality, we have
\[
\norm{\hat C-M}_{\mathrm F}
\le
2\norm{Y_r-M}_{\mathrm F}
\le
4\sqrt{2r}\,\xi,
\]
and therefore we have
\begin{align}\label{eq:kmeanfrob}
\norm{\hat C-M}_{\mathrm F}^2\le32r\xi^2.
\end{align}
Define the set of bad data points
\[
\mathcal B:=\left\{i\in[n]:\norm{\hat C_{\cdot,i}-M_{\cdot,i}}_2>\frac\delta3\right\}.
\]
If $\xi=0$, then we have $\hat C=M$, so $\mathcal B=\emptyset$.  If $\xi>0$, by definition of $\mathcal B$, inequality \eqref{eq:kmeanfrob}, and the separation condition \eqref{wb:eq:det-cond}, we have
\[
|\mathcal B|
<\frac{9}{\delta^2}\norm{\hat C-M}_{\mathrm F}^2
\le\frac{288r\xi^2}{\delta^2}
\le\frac89n_{\min}
<n_{\min}.
\]
In conclusion, in all cases, we have
\begin{equation}\label{wb:eq:det-bad-count}
|\mathcal B|\le\frac{288r\xi^2}{\delta^2},
\text{ and }
|\mathcal B|<n_{\min}.
\end{equation}
Consequently, every true cluster contains at least one point outside $\mathcal B$. If $i\in B_a\setminus\mathcal B$ and $j\in B_b\setminus\mathcal B$ with $a\neq b$, then $\hat z_i\neq\hat z_j$, because otherwise we have $\hat C_{\cdot,i}=\hat C_{\cdot,j}$ and therefore by triangle inequality we have
\[
\delta
\le
\norm{\mu_a-\hat C_{\cdot,i}}_2+
\norm{\hat C_{\cdot,j}-\mu_b}_2
\le\frac{2\delta}{3}
\]
which is a contradiction. This just means outside $\mathcal B$, points from different true clusters must be assigned to different estimated labels. For each $a \in [k]$, we choose one representative \(i_a\in
B_a\setminus\mathcal B\), and define $\phi(a):=\hat z_{i_a}$. The
preceding argument shows that the \(k\) labels
\(\phi(1), \ldots, \phi(k)\) are distinct and therefore exhaust
\([k]\). Thus, $\phi$ is a permutation of the set $[k]$.

Now consider arbitrary $a \in [k]$. We now show that all the data points in $B_a\setminus\mathcal B$ are assigned to the same estimated label $\phi(a)$ (note that $\phi(a)$ is not necessarily same as $a$). In fact, if a point \(i\in B_a\) had
\(\hat z_i\ne \phi(a)\), then \(\hat z_i= \phi(b)\) for some
\(b\ne a\), but this means $i$ and $i_b$ are from different true clusters but assigned to the same estimated label, which is not possible.  Therefore, every point in $B_a\setminus\mathcal B$ has estimated label
$\phi(a)$. If we relabel $\phi(a)$ by $a$ for all $a \in [k]$, then every point in $[n] \setminus\mathcal{B}$ is correctly
classified, i.e.,
\[
    \phi^{-1}(\hat{z}_{i}) = z^*_i
\]
for all $i \in [n] \setminus \mathcal{B}$.  Therefore, we conclude
\begin{align*}
&\ell(\hat z,z^\ast) \le 
\frac1n\sum_{i=1}^n\mathbf 1\{\phi^{-1}(\hat{z}_i)\ne z_i^\ast\}
\le
\frac{|\mathcal B|}{n}
\le
\frac{288r\xi^2}{n\delta^2},\\
&\ell_\infty(\hat z,z^\ast)
\le \max_{a\in[k]}
\frac1{n_a}\sum_{i\in B_a}\mathbf 1\{\phi^{-1}(\hat{z}_i)\ne a\} \le
\frac{|\mathcal B|}{n_{\min}}
\le
\frac{288r\xi^2}{n_{\min}\delta^2}.
\end{align*}
\end{proof}

\section{Measure concentration}

We use the following Bernstein inequality in \cite[Theorem~2.10]{BoucheronLugosiMassart2013}.

\begin{lemma}[Bernstein inequality]
\label{wb:thm:bernstein}
For a real-valued random variable $Z$, write
\[
\psi_Z(\lambda):=\log \E e^{\lambda Z},
\quad \lambda\ge 0,
\]
and
\[
\psi_Z^*(t)
:=
\sup_{\lambda\ge 0}
\bigl\{\lambda t-\psi_Z(\lambda)\bigr\},
\quad t\ge 0.
\]
Let $n\ge1$ be an integer, and let $X_1,\dots,X_n$ be independent
real-valued random variables.
Assume that there exist positive numbers $v$ and $b$ such that
\[
\sum_{i=1}^n \E\bigl[X_i^2\bigr]\le v
\]
and
\[
\sum_{i=1}^n \E\bigl[(X_i)_+^q\bigr]
\le
\frac{q!}{2}vb^{q-2}
\quad
\text{for all integers }q\ge 3,
\]
where
\[
x_+:=\max(x,0).
\]
If
\[
S:=\sum_{i=1}^n\bigl(X_i-\E X_i\bigr),
\]
then, for every $\lambda\in(0,1/b)$,
\[
\psi_S(\lambda)
\le
\frac{v\lambda^2}{2(1-b\lambda)}.
\]
In addition, for every $t>0$,
\[
\psi_S^*(t)
\ge
\frac{v}{b^2}
\left[
1+\frac{bt}{v}-\sqrt{1+\frac{2bt}{v}}
\right].
\]
In particular, for every $t>0$,
\[
\Pb\left\{
S\ge \sqrt{2vt}+bt
\right\}
\le e^{-t}.
\]
\end{lemma}

We use the following standard matrix Bernstein inequality for controlling the norm of sum of independent random matrices (see, e.g., \cite{Tropp2012UserFriendlyTailBounds}).  
\begin{lemma}[Matrix Bernstein]
\label{wb:thm:matrix-bernstein}
Let $d,N\ge1$, and let $Z_1,\ldots,Z_N$ be independent random
real symmetric $d\times d$ matrices.  Suppose that $R>0$ and
deterministic real symmetric $d\times d$ matrices $A_1,\ldots,A_N$
satisfy
\[
\E Z_j=0,
\quad
\E[Z_j^s]\preceq\frac{s!}{2}R^{s-2}A_j^2
\quad\text{for every integer }s\ge2\text{ and }j\in[N].
\]
Set
\[
v:=\left\|\sum_{j=1}^N A_j^2\right\|_{\op}.
\]
Then, for every $u\ge0$, we have
\begin{equation}\label{wb:eq:matrix-bernstein-tail}
\Pb\left\{\lambda_{\max}\left(\sum_{j=1}^N Z_j\right)\ge u\right\}
\le d\exp\left(-\frac{u^2}{2(v+Ru)}\right).
\end{equation}
\end{lemma}

For $v>0$ and $t>0$, let $u=\sqrt{2vt}+2Rt$.  Direct calculation
gives $u^2\ge2t(v+Ru)$, so
\eqref{wb:eq:matrix-bernstein-tail} implies
\begin{equation}\label{wb:eq:matrix-bernstein-confidence}
\Pb\left\{\lambda_{\max}\left(\sum_{j=1}^N Z_j\right)
\ge\sqrt{2vt}+2Rt\right\}\le de^{-t}.
\end{equation}

We also need the following concentration inequality for Chi-square random variables.

\begin{lemma}[Chi-square concentration]\label{wb:lem:noncentral-tools}
Let $p\ge1$ be an integer, let $\lambda\ge0$, and let
$Z\sim\chi_p^2(\lambda)$.

\smallskip
\noindent\textup{(i)} If $N\sim\mathrm{Poisson}(\lambda/2)$ and, conditionally on $N$, $Y_N\sim\chi_{p+2N}^2$, then $Y_N$ has the same distribution as $Z$.

\smallskip
\noindent\textup{(ii)} There is a positive absolute constant
$c_{\ref{wb:lem:noncentral-tools}.1}$ such that, for every $u\ge0$,
\[
\Pb\{Z\le p+\lambda-u\}
\le
\exp\left(-c_{\ref{wb:lem:noncentral-tools}.1}\frac{u^2}{p+2\lambda}\right).
\]
\end{lemma}

\begin{proof}
For \textup{(i)}, the moment generating function of $\chi_p^2(\lambda)$ is
\[
\E e^{tZ}
=(1-2t)^{-p/2}\exp\left(\frac{\lambda t}{1-2t}\right),
\quad t<\frac12 .
\]
On the other hand, using the moment generating function of a central chi-square variable and then the moment generating function of a Poisson variable, by independence, we have
\[
\begin{aligned}
\E e^{tY_N}
&=\E\left[(1-2t)^{-(p+2N)/2}\right]  \\
&=(1-2t)^{-p/2}\E\left[(1-2t)^{-N}\right] \\
&=(1-2t)^{-p/2}
\exp\left(\frac\lambda2\left((1-2t)^{-1}-1\right)\right) \\
&=(1-2t)^{-p/2}\exp\left(\frac{\lambda t}{1-2t}\right).
\end{aligned}
\]
The two moment generating functions agree on a neighborhood of the origin, proving the mixture representation.

For \textup{(ii)}, write $V:=p+2\lambda$. For $s\ge0$, we calculate
\[
\log \E\exp(-s(Z-p-\lambda))
=
 p\left(s-\frac12\log(1+2s)\right)+\frac{2\lambda s^2}{1+2s}.
\]
The elementary inequality $s-\frac12\log(1+2s)\le s^2$ for $s\ge0$ gives
\[
\log \E\exp(-s(Z-p-\lambda))
\le (p+2\lambda)s^2=Vs^2.
\]
By Chernoff's bound, for every $s\ge0$, we have
\[
\Pb\{Z-p-\lambda\le -u\}
\le
\exp(Vs^2-su).
\]
Taking $s=u/(2V)$ proves the desired lower-tail estimate.
\end{proof}

\begin{lemma}[Distance monotonicity]\label{wb:lem:distance-calculus}
Let $p\ge1$ be an integer and $\sigma>0$.  For every $\Delta>0$, the
map
\[
d\longmapsto\frac{d^4/\sigma^4}{p+d^2/\sigma^2}
\]
is nondecreasing on $[\Delta,\infty)$ and, for every $d\ge\Delta$, we have
\begin{equation}\label{wb:eq:kappa-distance-ratio}
\frac{\Delta^4/\sigma^4}{\Delta^2(p+\Delta^2/\sigma^2)}d^2
\le
\frac{d^4/\sigma^4}{p+d^2/\sigma^2}.
\end{equation}
Consequently, if
\[
\frac{\Delta^4/\sigma^4}{p+\Delta^2/\sigma^2}\ge1,
\]
then, for every $a>0$ and every $d\ge\Delta$, we have
\begin{equation}\label{wb:eq:distance-calculus-final}
 d^2\exp\left(-a\frac{d^4/\sigma^4}{p+d^2/\sigma^2}\right)
 \le \frac{2}{ae}\Delta^2
 \exp\left(-\frac a2
 \frac{\Delta^4/\sigma^4}{p+\Delta^2/\sigma^2}\right)
\text{ and }
 \frac{d^2}{\frac{d^4/\sigma^4}{p+d^2/\sigma^2}}
 \le
 \frac{\Delta^2}{\frac{\Delta^4/\sigma^4}{p+\Delta^2/\sigma^2}}.
\end{equation}
\end{lemma}

\begin{proof}
For the proof only, write
\[
\kappa(d):=\frac{d^4/\sigma^4}{p+d^2/\sigma^2},
\quad
\kappa_\Delta:=\kappa(\Delta).
\]
Let $x=d^2/\sigma^2$ and $x_0=\Delta^2/\sigma^2$.  The function $x\mapsto x^2/(p+x)$ is nondecreasing for $x>0$, which proves monotonicity.  In addition, we have
\[
\frac{\kappa(d)}{d^2}
=\frac{1}{\sigma^2}\frac{x}{p+x}
\ge
\frac{1}{\sigma^2}\frac{x_0}{p+x_0}
=\frac{\kappa_\Delta}{\Delta^2},
\]
which proves \eqref{wb:eq:kappa-distance-ratio}.  The second inequality in \eqref{wb:eq:distance-calculus-final} follows by rearranging,
\[
\frac{d^2}{\kappa(d)}
=\sigma^2+\frac{p\sigma^4}{d^2},
\]
which is nonincreasing in $d$.  Finally, \eqref{wb:eq:kappa-distance-ratio} gives $d^2/\Delta^2\le\kappa(d)/\kappa_\Delta$.  Since $\kappa(d)\ge\kappa_\Delta\ge1$, we have
\[
\frac{d^2}{\Delta^2}e^{-a\kappa(d)}
\le
\frac{\kappa(d)}{\kappa_\Delta}e^{-a\kappa(d)}
\le
\frac{2}{ae}e^{-(a/2)\kappa(d)}
\le
\frac{2}{ae}e^{-(a/2)\kappa_\Delta},
\]
where we used $\sup_{y\ge0}y e^{-ay/2}=2/(ae)$.  This proves the first
inequality.
\end{proof}

\section{Ingredients from high dimensional probability}

The following three statements are the Gaussian specializations and
net consequences needed below.  Vershynin states the first two source
theorems more generally for independent sub-Gaussian coordinates or
entries; standard Gaussian variables have an absolute sub-Gaussian
norm, so the constants below are absolute~\cite{Vershynin2018}.

\begin{lemma}[Norm of Gaussian vector]
\label{wb:thm:gaussian-vector-norm}
Let \(p\ge1\) be an integer and \(g\sim\mathcal N(0,I_p)\). There exists a positive
absolute constant \(c_{\ref{wb:thm:gaussian-vector-norm}.1}\) such that,
for every \(t\ge0\),
\[
\Pb\left(\left|\norm{g}_2-\sqrt p\right|\ge t\right)
\le 2e^{-c_{\ref{wb:thm:gaussian-vector-norm}.1}t^2}.
\]
Consequently, if \(\sigma>0\) and
\(\varepsilon\sim\mathcal N(0,\sigma^2I_p)\), then
\[
\Pb\left(
\left|\norm{\varepsilon}_2-\sigma\sqrt p\right|
\ge \sigma t
\right)
\le 2e^{-c_{\ref{wb:thm:gaussian-vector-norm}.1}t^2}.
\]
\end{lemma}

\begin{proof}
The coordinates of $g$ are independent, have unit second moment, and
have sub-Gaussian norms bounded by an absolute constant.
Equation~(3.3) in the proof of \cite[Theorem~3.1.1]{Vershynin2018},
specialized with dimension $p$, therefore gives the first display
for a positive absolute constant $c_1$.  The
second display follows because
$\varepsilon/\sigma\sim\mathcal N(0,I_p)$.  Relabeling $c_1$ gives the
assertion.
\end{proof}

We also need the standard bound of the norm of gaussian random matrices (see, e.g., \cite{Gordon1985}, \cite{DavidsonSzarek2001}, \cite{RudelsonVershynin2010}, and \cite
% [Theorem~4.4.5]
{Vershynin2018}).

\begin{lemma}[Norm of Gaussian matrix]
\label{wb:thm:gaussian-matrix}
Let \(p,n\ge1\) be integers, and let \(G\in\bbR^{p\times n}\) have independent
\(N(0,1)\) entries. There exist positive absolute constants
\(c_{\ref{wb:thm:gaussian-matrix}.1},
c_{\ref{wb:thm:gaussian-matrix}.2},
c_{\ref{wb:thm:gaussian-matrix}.3}\) such
that, for every \(t\ge0\), we have
\[
\Pb\left(
\norm{G}_{\op}>
c_{\ref{wb:thm:gaussian-matrix}.1}
\left(\sqrt n+\sqrt p+t\right)
\right)
\le 2e^{-c_{\ref{wb:thm:gaussian-matrix}.2}t^2}.
\]
In particular, for every \(d\ge1\), we have
\[
\Pb\left(
\norm{G}_{\op}>
c_{\ref{wb:thm:gaussian-matrix}.3}
\left(\sqrt n+\sqrt p+\sqrt d\right)
\right)
\le e^{-d}.
\]
\end{lemma}

One general method to bound the norm of a random matrix is to use the epsilon net argument, which is based on the following standard lemma.

\begin{lemma}[Size of net on sphere]
\label{wb:lem:net}
For every integer \(p\ge1\), there exists a \(1/4\)-net
\(\mathcal N\subset S^{p-1}\) satisfying
\[
|\mathcal N|\le9^p.
\]
In addition, for every symmetric matrix \(A\in\bbR^{p\times p}\),
\[
\norm{A}_{\op}
\le
2\sup_{x\in\mathcal N}|x^\top Ax|.
\]
\end{lemma}

\begin{proof}
Corollary~4.2.13 of \cite{Vershynin2018}, with
$\varepsilon=1/4$, gives a net of cardinality at most
$(1+2/\varepsilon)^p=9^p$.  Exercise~4.4.3(b) gives, for a symmetric
matrix and an $\varepsilon$-net with $\varepsilon<1/2$,
\[
\norm A_{\op}
\le
\frac{1}{1-2\varepsilon}
\sup_{x\in\mathcal N}|x^\top A x|.
\]
Taking $\varepsilon=1/4$ yields the factor $2$.
\end{proof}

\section{Proofs for classical PCA results}

\begin{proof}[Proof of Theorem \ref{wb:thm:untransformed-dim}]
We decompose the centered raw data  as
\[
X_c=P_c+E_c,
\quad
E_c:=E\left(I_n-\frac1n\one_{[n]}\one_{[n]}^\top\right).
\]
Since the centering matrix is an orthogonal projector, we have
\[
\norm{E_c}_{\op}\le \norm E_{\op}.
\]
By Lemma \ref{wb:thm:gaussian-matrix}, we have
\[
\norm E_{\op}\le c_1\sigma\sqrt{p+n}
\]
with probability at least $1-e^{-c_2n}$. Applying Lemma~\ref{wb:lem:subspace-det} with $M=P_c$ and $Y=X_c$ we obtain bound on the perturbation of eigenspaces. 
%\bk 
\end{proof}

\begin{proof}[Proof of Theorem \ref{wb:thm:untransformed-clustering}]
Write
\[
P_c:=P\left(I_n-\frac1n\one_{[n]}\one_{[n]}^\top\right),
\quad
E_c:=E\left(I_n-\frac1n\one_{[n]}\one_{[n]}^\top\right),
\]
so that $X_c=P_c+E_c$. Since all $k$ clusters are nonempty, $k\le n$, $k\ge2$, and $p\ge1$, we have $1\le r_\ast\le\min\{p,n\}$. We will apply Lemma~\ref{wb:thm:det} with $M=P_c$, $Y=X_c$, and $r=r_\ast$. Since $P_c\in\bbR^{p\times n}$ is cluster-wise constant, $\rank(P_c)\le\min\{p,k-1\}=r_\ast$. Its population center for cluster $a$ is $\theta_a^*-n^{-1}P\one_{[n]}$, so its cluster separation is
\[
\delta_c=\min_{a\ne b}\norm{\theta_a^*-\theta_b^*}_2\ge\Delta.
\]
By Lemma \ref{wb:thm:gaussian-matrix}, we have
\[
\norm{E_c}_{\op}\le c_1\sigma\sqrt{p+n}
\]
with probability at least $1-e^{-c_2n}$. By assuming the separation condition \[
\frac{\Delta^2}{\sigma^2}
\ge
18^2c_1^2
\frac{(k-1)(n+p)}{n_{\min}},
\]
we guarantee
\[
\delta_c\ge\Delta
\ge18c_1\sigma
\sqrt{\frac{(k-1)(n+p)}{n_{\min}}}
\ge18\norm{E_c}_{\op}\sqrt{\frac{r_\ast}{n_{\min}}},
\]
Therefore, by Theorem \ref{wb:thm:det}, we have
\[
\ell(\hat z_{\rm PCA},z^\ast)
\le
288r_\ast\frac{\norm{E_c}_{\op}^2}{n\Delta^2}
\le
c_3(k-1)\frac{\sigma^2(n+p)}{n\Delta^2},
\]
\[
\ell_\infty(\hat z_{\rm PCA},z^\ast)
\le
288r_\ast\frac{\norm{E_c}_{\op}^2}{n_{\min}\Delta^2}
\le
c_4(k-1)\frac{\sigma^2(n+p)}{n_{\min}\Delta^2}.
\]
\end{proof}
	
	\section{More details about the expectation of data truncation matrices and ideal truncation}\label{app:expectation-decomp}

    We first show that $\widehat L$ is symmetric positive semidefinite so that Algorithm~\ref{clusteringalgo} is well defined.

\begin{lemma}[Positive semidefiniteness]\label{lem:psd}
Let $p,n\ge1$, let
$X=[X_1,\ldots,X_n]\in\mathbb R^{p\times n}$ be a data matrix,
and let $\phi:[0,\infty)\to[0,1]$ be Borel measurable.  Let
$(\hat A,\hat D,\hat L)$ be the data truncation matrices for the data
$X$ using threshold function $\phi$.
Then, for any $(X_1,\ldots,X_n)$ (i.e., every realization of the random observation), the matrix $\hat L$
is symmetric positive semidefinite and satisfies
$\hat L\one_{[n]}=0$.
	\end{lemma}
	
	\begin{proof}
		We first claim that $\widehat L$ is symmetric. In fact, for any $i,j\in[n]$, $\|X_i-X_j\|_2^2=\|X_j-X_i\|_2^2$, and therefore $\widehat A_{ij}=\widehat A_{ji}$ and $\widehat A$ is symmetric. The degree matrix $\widehat D$ is diagonal, so $\widehat L=\widehat D-\widehat A$ is symmetric as well.
		
		Next, let $x\in\bbR^n$ be arbitrary, and we will show that $x^\top \widehat L x \ge 0$. Using $\widehat L=\widehat D-\widehat A$, we have
		\[
		x^\top \widehat L x = x^\top \widehat D x - x^\top \widehat A x.
		\]
		Since $\widehat D_{ii}=\sum_{j=1}^n \widehat A_{ij}$, we have
		\[
		x^\top \widehat D x = \sum_{i=1}^n \widehat D_{ii}x_i^2 = \sum_{i=1}^n\sum_{j=1}^n \widehat A_{ij}x_i^2,
		\quad
		x^\top \widehat A x = \sum_{i=1}^n\sum_{j=1}^n \widehat A_{ij}x_ix_j.
		\]
		Therefore, we conclude
		\[
		x^\top \widehat L x = \sum_{i,j}\widehat A_{ij}x_i^2 - \sum_{i,j}\widehat A_{ij}x_ix_j.
		\]
        and we will complete the squares.
		Since $\widehat A$ is symmetric, $\sum_{i,j}\widehat A_{ij}x_i^2=\sum_{i,j}\widehat A_{ij}x_j^2$. Averaging these equal expressions, we obtain
		\begin{align*}
			x^\top \widehat L x
			&=\frac12\sum_{i,j}\widehat A_{ij}(x_i^2+x_j^2) - \sum_{i,j}\widehat A_{ij}x_ix_j\\
			&=\frac12\sum_{i,j}\widehat A_{ij}(x_i^2+x_j^2-2x_ix_j)
			=\frac12\sum_{i,j}\widehat A_{ij}(x_i-x_j)^2.
		\end{align*}
		
		Because $\phi$ takes values in $[0,1]$, each entry $\widehat A_{ij}\ge 0$. Also $(x_i-x_j)^2\ge 0$. So the $\frac12\sum_{i,j}\widehat A_{ij}(x_i-x_j)^2$ is nonnegative for every $x\in\bbR^n$, and therefore $\widehat L\succeq 0$.
		
		In addition, by definition $\widehat D=\diag(\widehat A\one_{[n]})$, so $\widehat D\one_{[n]}=\widehat A\one_{[n]}$. So we have
		\[
		\widehat L\one_{[n]}=(\widehat D-\widehat A)\one_{[n]}=\widehat D\one_{[n]}-\widehat A\one_{[n]}=0.
		\]
	\end{proof}

To study the properties of the data truncation matrices, we first analyze its expectation.
	\begin{lemma}[Expectation of data truncation matrices]\label{prop:barAbarL}
Let $k\ge2$, $n\ge2$, and $p\ge1$.  Let
\[
\Theta^*=(\theta_1^*,\ldots,\theta_k^*)\in\mathbb R^{p\times k}
\]
and
\[
z^*=(z_1^*,\ldots,z_n^*)\in[k]^n.
\]
Let $(X,P,E)$ be a Gaussian mixture observation model for the true
cluster configuration $(\Theta^*,z^*)$ and noise level $\sigma$.
For $a\in[k]$, set
$B_a:=\{i\in[n]:z_i^*=a\}$ and $n_a:=|B_a|$.  Let
$\phi:[0,\infty)\to[0,1]$ be Borel measurable, and let
$(\hat A,\hat D,\hat L)$ be the data truncation matrices for the data
$X$ using threshold function $\phi$.
Let $\varepsilon\sim\mathcal N(0,\sigma^2I_p)$.
		Then for all $i,j\in[n]$, we have:
		
		(i) ${\E [\hat A]}_{ii}=\phi(0)$.
		
		(ii) If $i\neq j$ and $z^*_i=z^*_j$, then
\[
{\E [\hat A]}_{ij}
=\E\left[\phi\left(2\|\varepsilon\|_2^2\right)\right].
\]
		
		(iii) If $z^*_i=a$, $z^*_j=b$ and $a\neq b$, then
\[
{\E [\hat A]}_{ij}
=\E\left[\phi\left(\|\theta_a^*-\theta_b^*
+\sqrt2\,\varepsilon\|_2^2\right)\right].
\]
		
		(iv) If $z^*_i=l$, we have
\[
\E \hat D_{ii}=\phi(0)
+(n_l-1)\E\left[\phi\left(2\|\varepsilon\|_2^2\right)\right]
+\sum_{t\ne l}n_t\E\left[
\phi\left(\|\theta_l^*-\theta_t^*+\sqrt2\,\varepsilon\|_2^2\right)
\right].
\]
		
		(v) Consequently, for $z^*_i=a$ and $z^*_j=b$:
		\[
		{\E [\hat L]}_{ij}=
		\begin{cases}
			\displaystyle
\sum_{t\ne a}n_t\E\left[
\phi\left(\|\theta_a^*-\theta_t^*+\sqrt2\,\varepsilon\|_2^2\right)
\right]
+(n_a-1)\E\left[\phi\left(2\|\varepsilon\|_2^2\right)\right],
& i=j,\\[2mm]
			\displaystyle-\E\left[\phi\left(2\|\varepsilon\|_2^2\right)\right],
& i\neq j\ \text{and}\ a=b,\\[2mm]
			\displaystyle-\E\left[
\phi\left(\|\theta_a^*-\theta_b^*+\sqrt2\,\varepsilon\|_2^2\right)
\right],
& a\neq b.
		\end{cases}
		\]

	\end{lemma}

    \begin{proof}
For convenience, we introduce the notation
		\[
		\alpha_{\rm same}
		:=
		\E\left[\phi\left(2\|\e\|_2^2\right)\right],
		\quad
		\gamma_{a,b}
		:=
		\E\left[
		\phi\left(\|\theta_a^*-\theta_b^*+\sqrt2\,\e\|_2^2\right)
		\right]
		\quad(a\ne b).
		\]
		
		Result (i) follows from direct calculation
		\[
		\widehat A_{ii}=\phi(\|X_i-X_i\|_2^2)=\phi(0).
		\]
		
		For (ii), fix $i\neq j$ with $z_i^*=z_j^*=l$. Then
		\[
		X_i-X_j = (\theta_l^*+\varepsilon_i)-(\theta_l^*+\varepsilon_j)=\varepsilon_i-\varepsilon_j.
		\]
		Since $\varepsilon_i,\varepsilon_j$ are independent $\mathcal{N}(0,\sigma^2 I_p)$ random vectors, their difference is Gaussian with mean $0$ and covariance
		\[
		\cov(\varepsilon_i-\varepsilon_j)=\cov(\varepsilon_i)+\cov(\varepsilon_j)=\sigma^2 I_p+\sigma^2 I_p=2\sigma^2 I_p.
		\]
		Therefore there exists a $p$-dimensional Gaussian random vector $\e\sim \mathcal{N}(0,\sigma^2 I_p)$ such that
		\[
		\varepsilon_i-\varepsilon_j \stackrel{d}{=} \sqrt{2}\,\e.
		\]
		So in conclusion we have
		\[
		\widehat A_{ij}=\phi(\|X_i-X_j\|_2^2)=\phi(\|\varepsilon_i-\varepsilon_j\|_2^2)
		\stackrel{d}{=}\phi(\|\sqrt{2}\e\|_2^2)=\phi(2\|\e\|_2^2).
		\]
		Taking expectations yields
		\[
		\E[\widehat A_{ij}] = \E\big[\phi(2\|\e\|_2^2)\big]=\alpha_{\rm same}.
		\]
		
		To prove (iii), fix $i\in B_a$ and $j\in B_b$ with $a\neq b$. Then
		\[
		X_i-X_j = (\theta_a^*+\varepsilon_i)-(\theta_b^*+\varepsilon_j)=(\theta_a^*-\theta_b^*)+(\varepsilon_i-\varepsilon_j).
		\]
		As above, we can write $\varepsilon_i-\varepsilon_j \stackrel{d}{=} \sqrt{2}\,\e$ for $\e\sim\mathcal{N}(0,\sigma^2 I_p)$, so that we can write
		\[
		\widehat A_{ij}=\phi(\|X_i-X_j\|_2^2)\stackrel{d}{=}\phi(\|\theta_a^*-\theta_b^*+\sqrt{2}\e\|_2^2),
		\]
		and the taking expectation we have
		\[
		\E[\widehat A_{ij}] = \E\left[\phi\left(\|\theta_a^*-\theta_b^*+\sqrt{2}\e\|_2^2\right)\right]=\gamma_{a,b}.
		\]
		
		For (iv), by definition of the degree matrix, we have
		\[
		\widehat D_{ii} = \sum_{j=1}^n \widehat A_{ij}.
		\]
		By linearity of expectation, we obtain
		\[
		\E[\widehat D_{ii}] = \sum_{j=1}^n \E[\widehat A_{ij}].
		\]
		Fix $i\in B_l$. Splitting the sum over $j$ into the cases $j=i$, $j\in B_l\setminus\{i\}$, and $j\in B_t$ for each $t\neq l$, and then using (i)--(iii), we obtain
		\begin{align*}
			\E[\widehat D_{ii}]
			&= \E[\widehat A_{ii}] + \sum_{\substack{j\in B_l\\ j\neq i}}\E[\widehat A_{ij}] + \sum_{t\neq l}\sum_{j\in B_t}\E[\widehat A_{ij}]\\
			&= \phi(0) + (n_l-1)\alpha_{\rm same} + \sum_{t\neq l} n_t \gamma_{l,t}.
		\end{align*}
		
		To prove (v), note first that $\widehat D$ is diagonal, so for $i\neq j$ we have $(\E\widehat D)_{ij}=0$, and therefore we have
		\[
		({\E [\hat L]})_{ij}=(\E[\widehat D])_{ij}-(\E[\widehat A])_{ij}= -(\E[\widehat A])_{ij}.
		\]
		The off-diagonal cases now follow directly from (ii)--(iii), i.e.,
		\[
		{\E [\hat L]}_{ij}=
		\begin{cases}
			-\alpha_{\rm same},& i\neq j,\ z_i^*=z_j^*,\\
			-\gamma_{a,b},& z_i^*=a,\ z_j^*=b,\ a\neq b.
		\end{cases}
		\]
		For the diagonal, using $\widehat L_{ii}=\widehat D_{ii}-\widehat A_{ii}$ together with part (iv), we get
		\[
		\E[\widehat L_{ii}] = \E[\widehat D_{ii}] - \E[\widehat A_{ii}]
		= \big(\phi(0)+(n_a-1)\alpha_{\rm same}+\sum_{t\neq a}n_t\gamma_{a,t}\big)-\phi(0)
		=\sum_{t\neq a}n_t\gamma_{a,t}+(n_a-1)\alpha_{\rm same}.
		\]
	\end{proof}

    \begin{lemma}[Monotonicity]\label{wb:lem:cross-monotonicity}
Let $p\ge1$, $\sigma>0$, $\Delta>0$, and let
$\theta_a^*,\theta_b^*\in\bbR^p$ satisfy
$\norm{\theta_a^*-\theta_b^*}_2\ge\Delta$, and let
$\varepsilon\sim\mathcal N(0,\sigma^2I_p)$.  Let
$Z_\Delta\sim\chi_p^2(\Delta^2/(2\sigma^2))$.  Then
\[
\Pb\left(
\norm{\theta_a^*-\theta_b^*+\sqrt2\,\varepsilon}_2^2
\ge2\sigma^2p+\frac{\Delta^2}{2}
\right)
\ge
\Pb\left(
2\sigma^2Z_\Delta
\ge2\sigma^2p+\frac{\Delta^2}{2}
\right).
\]
In addition, there exists a positive absolute constant
$c_{\ref{wb:lem:cross-monotonicity}.1}$ such that
\[
\Pb\left(
\norm{\theta_a^*-\theta_b^*+\sqrt2\,\varepsilon}_2^2
<2\sigma^2p+\frac{\Delta^2}{2}
\right)
\le
\exp\left(-c_{\ref{wb:lem:cross-monotonicity}.1}
\frac{\norm{\theta_a^*-\theta_b^*}_2^4/\sigma^4}{p+\norm{\theta_a^*-\theta_b^*}_2^2/\sigma^2}
\right).
\]
\end{lemma}

    \begin{proof}
For each $d\ge0$, let $Z_d$ denote a random variable with distribution
$\chi_p^2(d^2/(2\sigma^2))$; no joint construction of this family is
used.  For the proof only, set
\[
\gamma(d):=
\Pb\left(
2\sigma^2Z_d
\ge2\sigma^2p+\frac{\Delta^2}{2}
\right),
\quad d\ge0.
\]
Rotational invariance of the Gaussian noise identifies the probability
in the first display of the statement with
$\gamma(\norm{\theta_a^*-\theta_b^*}_2)$.  We next prove monotonicity
in $d$. By Lemma~\ref{wb:lem:noncentral-tools}\textup{(i)}, we obtain
\[
\chi_p^2(\lambda)\stackrel d=\chi_{p+2N}^2,
\quad N\sim {\rm Poisson}(\lambda/2).
\]
If $\lambda_2\ge\lambda_1$, couple $N_2=N_1+N'$ with $N_1\sim{\rm Poisson}(\lambda_1/2)$ and $N'\sim{\rm Poisson}((\lambda_2-\lambda_1)/2)$ independent. Conditional on $(N_1,N')$, realize
\[
\chi_{p+2N_2}^2=\chi_{p+2N_1}^2+\chi_{2N'}^2
\]
with independent central chi-square summands. Hence $\chi_p^2(\lambda_1)$ is stochastically dominated by $\chi_p^2(\lambda_2)$. Since $d\mapsto d^2/(2\sigma^2)$ is nondecreasing, $\gamma(d)$ is nondecreasing in $d$, which proves the first display.

For the tail bound, set $d:=\norm{\theta_a^*-\theta_b^*}_2$ and let
$Z\sim\chi_p^2(\lambda)$ with $\lambda=d^2/(2\sigma^2)$.  Since
$d\ge\Delta$ and the keep event in the definition of
$\gamma(d)$ uses a weak inequality,
\[
1-\gamma(d)
=
\Pb\left\{Z<p+\frac{\Delta^2}{4\sigma^2}\right\}
\le
\Pb\left\{Z\le p+\frac{d^2}{4\sigma^2}\right\}
=
\Pb\left\{Z\le p+\lambda-\frac{d^2}{4\sigma^2}\right\}.
\]
Lemma~\ref{wb:lem:noncentral-tools}\textup{(ii)}, applied with
$u=d^2/(4\sigma^2)$, gives, for a positive absolute constant $c_1$,
\[
1-\gamma(d)
\le
\exp\left(-c_1\frac{d^4/\sigma^4}{p+d^2/\sigma^2}\right).
\]
Rotational invariance again identifies $1-\gamma(d)$ with the
strict-inequality probability in the statement.  Taking its positive
absolute constant to be $c_1$ proves the desired estimate.
\end{proof}

We now prove the following subspace decomposition of $\bbR^n$, which is used in deriving the decomposition of ideal $L$.

\begin{lemma}[Orthogonal decomposition of the space]\label{lem:decomp}
Let $n,k\ge1$, and let $B_1,\ldots,B_k$ be a partition of $[n]$ into
nonempty sets.  Put $n_a:=|B_a|$ and define
\[
q_a:=\frac{\one_{B_a}}{\sqrt{n_a}},\quad
Q:=[q_1\ \cdots\ q_k],\quad
\mathcal C:=\{y\in\bbR^n:y\text{ is constant on every }B_a\}.
\]
Also set
\[
s:=(\sqrt{n_1},\ldots,\sqrt{n_k})^\top,
\quad
\mathcal C_0:=\mathcal C\cap(\operatorname{span}(\one_{[n]}))^\perp,
\quad
H_n:=I_n-\frac1n\one_{[n]}\one_{[n]}^\top.
\]
For $a\in[k]$, define
\[
\mathcal Z_a:=\{x\in\bbR^n:\supp(x)\subseteq B_a,
\ \one_{B_a}^\top x=0\},\quad
\Pi_a:=\diag(\one_{B_a})-\frac{\one_{B_a}\one_{B_a}^\top}{n_a}.
\]
Then $Q^\top Q=I_k$, $QQ^\top$ is the orthogonal projector onto
$\mathcal C=\range(Q)$, and $\Pi_a$ is the orthogonal projector onto
$\mathcal Z_a$.  In addition, we have
\[
Q^\top\one_{[n]}=s,
\quad
Qs=\one_{[n]},
\quad
\norm{s}_2^2=n,
\]
$I_k-n^{-1}ss^\top$ is the orthogonal projector onto $(\operatorname{span}(s))^\perp$, and
\begin{equation*}
Q\left(I_k-\frac1nss^\top\right)Q^\top
=QQ^\top-\frac1n\one_{[n]}\one_{[n]}^\top
\end{equation*}
is the orthogonal projector onto
$\mathcal C_0=Q((\operatorname{span}(s))^\perp)$.  The subspaces
$\mathcal C,\mathcal Z_1,\ldots,\mathcal Z_k$ are mutually orthogonal,
and
\[
QQ^\top+\sum_{a=1}^k\Pi_a=I_n,
\quad
\Pi_a\Pi_b=0\ (a\ne b),
\quad
Q^\top\Pi_a=0\ (a\in[k]).
\]
In particular, we have
\begin{equation*}
H_n
=Q\left(I_k-\frac1nss^\top\right)Q^\top
+\sum_{a=1}^k\Pi_a,
\end{equation*}
and $\bbR^n$ is the orthogonal direct sum of
$\Span\{\one_{[n]}\},\mathcal C_0,\mathcal Z_1,\ldots,\mathcal Z_k$.
\end{lemma}

\begin{proof}
Disjointness of the sets gives
$q_a^\top q_b=\mathbbm{1}_{\{a=b\}}$, so $Q^\top Q=I_k$ and $QQ^\top$
is the orthogonal projector onto $\range(Q)=\mathcal C$.  Direct
multiplication shows that each $\Pi_a$ is symmetric and idempotent.  Its
range consists precisely of the vectors supported on $B_a$ whose coordinates
sum to zero, so its range is $\mathcal Z_a$.  When $n_a=1$, we have
$\Pi_a=0$ and $\mathcal Z_a=\{0\}$, so this assertion includes singleton
blocks.  The stated orthogonality follows
from disjoint supports and from $\one_{B_a}^\top x=0$ for
$x\in\mathcal Z_a$.  In addition, we have
\[
QQ^\top=\sum_{a=1}^k\frac{\one_{B_a}\one_{B_a}^\top}{n_a},
\quad
\sum_{a=1}^k\diag(\one_{B_a})=I_n,
\]
which proves the decomposition of $I_n$. And the remaining product identities
follow from the same orthogonality.

Because the $B_a$ form a partition, we have
\[
Q^\top\one_{[n]}
=\bigl(\sqrt{n_1},\ldots,\sqrt{n_k}\bigr)^\top=s,
\quad
Qs=\sum_{a=1}^k\one_{B_a}=\one_{[n]},
\]
and $\norm{s}_2^2=\sum_{a=1}^k n_a=n$.  It follows by direct
multiplication that $I_k-n^{-1}ss^\top$ is symmetric and idempotent,
with kernel $\Span\{s\}$ and range $(\operatorname{span}(s))^\perp$.

Next, we show $Q((\operatorname{span}(s))^\perp)=\mathcal C_0$. Consider arbitrary $y\in\mathcal C_0$, then we have $\one_{[n]}^\top y=0$ and we can write $y=Qv$ for some $v \in \bbR^k$, and therefore we have
\[
0=\one_{[n]}^\top y
=(Q^\top\one_{[n]})^\top v=s^\top v.
\]
So $v \in (\operatorname{span}(s))^\perp$.
Then we conclude that $\mathcal C_0 \subset Q((\operatorname{span}(s))^\perp)$. And the exact  same argument starting from $y=Q v$ with $s^\top v=0$ shows $Q((\operatorname{span}(s))^\perp) \subset \mathcal C_0 $.

Since $Q$ is an isometry from
$\bbR^k$ onto $\mathcal C$, the matrix
$Q(I_k-n^{-1}ss^\top)Q^\top$ is therefore the orthogonal projector
onto $\mathcal C_0$.  By the identities $Qs=\one_{[n]}$ and
$Q^\top Q=I_k$, we also have
\[
Q\left(I_k-\frac1nss^\top\right)Q^\top
=QQ^\top-\frac1n\one_{[n]}\one_{[n]}^\top,
\]
which implies decomposition of $H_n$ by direct calculation.

Finally, combining the orthogonal decomposition
$\mathcal C=\Span\{\one_{[n]}\}\oplus\mathcal C_0$ with the orthogonal decomposition we conclude that $\bbR^n$ is the orthogonal direct sum of
$\Span\{\one_{[n]}\},\mathcal C_0,\mathcal Z_1,\ldots,\mathcal Z_k$.
\end{proof}

\begin{lemma}[Factorization of the signal matrix]\label{lem:P}
Let $p,n,k\ge1$, let $\theta_1^*,\ldots,\theta_k^*\in\bbR^p$, and let
$z^*=(z_1^*,\ldots,z_n^*)\in[k]^n$ contain every label in $[k]$.
Define
\[
B_a:=\{i\in[n]:z_i^*=a\},\quad n_a:=|B_a|,
\quad
\Theta^*:=[\theta_1^*\ \cdots\ \theta_k^*],
\quad
P:=[\theta_{z_1^*}^*\ \cdots\ \theta_{z_n^*}^*],
\]
and
\[
Q:=\left[\frac{\one_{B_1}}{\sqrt{n_1}}\ \cdots\
\frac{\one_{B_k}}{\sqrt{n_k}}\right],
\quad
s:=(\sqrt{n_1},\ldots,\sqrt{n_k})^\top,
\quad
\Pi_a:=\diag(\one_{B_a})-\frac{\one_{B_a}\one_{B_a}^\top}{n_a}.
\]
Also set
\[
H_n:=I_n-\frac1n\one_{[n]}\one_{[n]}^\top,
\quad
\bar\theta:=\frac1nP\one_{[n]}.
\]
Then
\[
P=\Theta^*\diag(\sqrt{n_1},\ldots,\sqrt{n_k})Q^\top,
\quad
PQ=\Theta^*\diag(\sqrt{n_1},\ldots,\sqrt{n_k}),
\]
and
\[
PQQ^\top=P,\quad P\Pi_a=0\quad(a\in[k]).
\]
Furthermore,
\begin{equation*}
PQ\left(I_k-\frac1nss^\top\right)Q^\top
=PH_n
=P-\bar\theta\one_{[n]}^\top.
\end{equation*}
In particular, the $i$th column of each matrix in
the preceding display is
$\theta_{z_i^*}^*-\bar\theta$.
\end{lemma}

\begin{proof}
If $i\in B_a$, then $Q^\top e_i=e_a/\sqrt{n_a}$, so the $i$th column
of $\Theta^*\diag(\sqrt{n_1},\ldots,\sqrt{n_k})Q^\top$ is
$\theta_a^*=P_{\cdot,i}$.  This proves the factorization.  Right
multiplication by $Q$ and the identity $Q^\top Q=I_k$ give the formula
for $PQ$; multiplying that formula by $Q^\top$ gives $PQQ^\top=P$.
Finally,
$\Pi_a x$ is supported on $B_a$ and its coordinates sum to zero; because
the columns of $P$ are all equal to $\theta_a^*$ on $B_a$, this gives
$P\Pi_a x=0$ for every $x\in\bbR^n$.

Since the $B_a$ partition $[n]$,
$Qs=\sum_{a=1}^k\one_{B_a}=\one_{[n]}$, and therefore
\begin{align*}
PQ\left(I_k-\frac1nss^\top\right)Q^\top
&=PQQ^\top-\frac1n(PQs)(Qs)^\top\\
&=P-\frac1n(P\one_{[n]})\one_{[n]}^\top\\
&=PH_n
=P-\bar\theta\one_{[n]}^\top.
\end{align*}
This proves the matrix identity, and the formula follows from
$P_{\cdot,i}=\theta_{z_i^*}^*$.
\end{proof}

With above preparation, we can now prove Lemma \ref{wb:lem:ideal-target}.
\begin{proof}[Proof of Lemma \ref{wb:lem:ideal-target}]
We follow the complete-graph subtraction described in
Subsection~\ref{wb:subsec:proof-overview}.  For a nonempty set
$T\subseteq[n]$, define
\[
K_T:=\one_T\one_T^\top-\diag(\one_T).
\]
Thus $K_T$ is the adjacency matrix of the complete graph on $T$,
embedded in $\bbR^{n\times n}$.  In particular, its diagonal is zero.
From Definition~\ref{def:ideal-truncation-model}, an off-diagonal entry
of $A$ equals one exactly when its two indices do not lie in the same
cluster.  Therefore, we have the following matrix identity
\begin{equation*}
A=K_{[n]}-\sum_{a=1}^k K_{B_a}.
\end{equation*}

The maps $H\mapsto\mathcal D(H)$ and
$H\mapsto\mathcal L(H)=\mathcal D(H)-H$ are linear.  Using
$K_T\one_{[n]}=(|T|-1)\one_T$, we have
\begin{align*}
\mathcal L(K_T)
&=(|T|-1)\diag(\one_T)
  -\bigl(\one_T\one_T^\top-\diag(\one_T)\bigr)\\
&=|T|\diag(\one_T)-\one_T\one_T^\top.
\end{align*}
For $T=[n]$, this is
$\mathcal L(K_{[n]})=nI_n-\one_{[n]}\one_{[n]}^\top$; for
$T=B_a$, it is $\mathcal L(K_{B_a})=n_a\Pi_a$.  Therefore, we have
the matrix identity
\begin{equation*}
L=nI_n-\one_{[n]}\one_{[n]}^\top-
\sum_{a=1}^k n_a\Pi_a.
\end{equation*}

We next transform this identity into spectral decomposition.  By
Lemma~\ref{lem:decomp},
\[
I_n=QQ^\top+\sum_{a=1}^k\Pi_a,
\quad
Qs=\one_{[n]},
\quad
\norm{s}_2^2=\sum_{a=1}^k n_a=n.
\]
It follows that
\[
QQ^\top-\frac1n\one_{[n]}\one_{[n]}^\top
=Q\left(I_k-\frac1nss^\top\right)Q^\top.
\]
Substituting these identities into the preceding formula for $L$ and
dividing by $n$ proves \eqref{wb:eq:ideal-L-decomp}.

For convenience, write
\[
\Pi_0:=Q(I_k-\frac1nss^\top)Q^\top,
\quad
\lambda_a:=1-\frac{n_a}{n}\quad(a\in[k]).
\]
Since $\norm{s}_2^2=n$, $(I_k-\frac1nss^\top)$ is the orthogonal projector onto
$(\operatorname{span}(s))^\perp$.  Lemma~\ref{lem:decomp} then shows that $\Pi_0$ and
$\Pi_1,\ldots,\Pi_k$ are pairwise orthogonal projectors,
with
\[
\range(\Pi_0)
=\{x\in\bbR^n:x\text{ is constant on each }B_a,
\ \one_{[n]}^\top x=0\}.
\]
Because every cluster is nonempty and $k\ge2$, we have
$0<\lambda_a<1$ for every $a$.  So we have the spectral decomposition
\[
\frac Ln=\Pi_0+\sum_{a=1}^k\lambda_a\Pi_a,
\]
where the orthogonal subspace
$\Span\{\one_{[n]}\}$ is omitted, because $L/n$ is zero on it. Using $P\Pi_a=0$, we have
\[
P\frac Ln=P\Pi_0=PH_n=P_c,
\]
which proves \eqref{wb:eq:P-Ln-Hn}.

The matrix
\[
R_0:=\Pi_0+\sum_{a=1}^k\sqrt{\lambda_a}\,\Pi_a
\]
is symmetric positive semidefinite, and pairwise orthogonality gives
$R_0^2=L/n$.  By uniqueness of the symmetric positive-semidefinite
square root, $R_0=S_0$.  This proves the formula in
\eqref{wb:eq:S0-explicit}.  Each of $\Pi_0,\Pi_1,\ldots,\Pi_k$
annihilates $\one_{[n]}$, so the same formula gives
$S_0\one_{[n]}=0$.

It remains to identify the action on the population signal.  The columns
of $P$ are constant within each $B_a$, so Lemma~\ref{lem:P} implies
$P\Pi_a=0$ for every $a$ and
\[
P\Pi_0
=PQ\left(I_k-\frac1nss^\top\right)Q^\top
=PH_n.
\]
Therefore, we have
\[
PS_0=P\Pi_0+\sum_{a=1}^k\sqrt{\lambda_a}\,P\Pi_a
=PH_n=P_c,
\]
which proves \eqref{wb:eq:PS0-Pc}.  Since
$P\one_{[n]}/n=\bar\theta$, the $i$th column of $PH_n$ is
$\theta_{z_i^*}^*-\bar\theta$.  
\end{proof}

\begin{proof}[Proof of Lemma \ref{wb:lem:ideal-noise-scale}]
Without loss of generality, we relabel the clusters so that
\[
n_1=n_{\max}=n-m.
\]
If $n_1=1$, then all clusters must have size $1$ and therefore there must be $n$ clusters. In this case, the condition $p+m\le n$ will force $p=1$, which reduces the model to the scalar case, where the analysis becomes straightforward. So we works with the case $n_1 \ge 2$.
For $a\in[k]$, let
\[
q_a:=n_a^{-1/2}\one_{B_a},
\quad Q:=[q_1\ \cdots\ q_k],
\quad s:=(\sqrt{n_1},\ldots,\sqrt{n_k})^\top,
\]
and
\[
\Pi_a:=\diag(\one_{B_a})-
\frac1{n_a}\one_{B_a}\one_{B_a}^\top
\]
as in Lemma \ref{wb:lem:ideal-target}.
By Lemma~\ref{wb:lem:ideal-target}, we obtain
\[
S_0
=
Q\left(I_k-\frac1nss^\top\right)Q^\top
+
\sum_{a=1}^k
\sqrt{1-\frac{n_a}{n}}\,\Pi_a.
\]
Since \(1-n_1/n=m/n\), write
\[
S_0=T+\sqrt{\frac mn}\,\Pi_1,
\]
where
\[
T:=
Q\left(I_k-\frac1nss^\top\right)Q^\top
+
\sum_{a=2}^k
\sqrt{1-\frac{n_a}{n}}\,\Pi_a.
\]
By Lemma~
\ref{lem:decomp} and 
Lemma~\ref{wb:lem:ideal-target} we have
\[
\norm{T}_{\op}\le1
\]
and
\[
\rank(T)
=(k-1)+\sum_{a=2}^k(n_a-1)
=m.
\]

Choose \(R\in\bbR^{n\times m}\) whose columns form an orthonormal
eigenbasis for the restriction of \(T\) to \(\range(T)\). Then
\[
T=RDR^\top
\]
for some diagonal matrix \(D\) satisfying \(\norm D_{\op}\le1\).
Similarly, since \(n_1\ge2\), let
\(R_1\in\bbR^{n\times(n_1-1)}\) have orthonormal columns spanning
\(\range(\Pi_1)\).  
So we conclude
\[
\Pi_1=R_1R_1^\top.
\]
Therefore, we have
\[
ES_0
=
ERDR^\top
+
\sqrt{\frac mn}\,ER_1R_1^\top.
\]
By the triangle inequality, we have
\[
\norm{ES_0}_{\op}
\le
\norm{ER}_{\op}
+
\sqrt{\frac mn}\,\norm{ER_1}_{\op}.
\]

By rotational invariance of the Gaussian matrix \(E\), we know that
\[
\frac1\sigma ER\in\bbR^{p\times m}
\]
is a standard Gaussian matrix.  In addition, 
% if \(n_1\ge2\), then
\(\sigma^{-1}ER_1\in\bbR^{p\times(n_1-1)}\) is also a standard
Gaussian matrix. 
Apply Lemma~\ref{wb:thm:gaussian-matrix} with
tail parameters $p+m$ and $n$, respectively.  Since $p\le n$,
$n_1-1\le n$, and $p+m\ge2$, a union bound gives positive absolute
constants $c_1,c_2,c_3$ such that, with probability at least
$1-e^{-c_1(p+m)}$, simultaneously we have
\[
\norm{ER}_{\op}\le c_2\sigma\sqrt{p+m},
\]
and
\[
\norm{ER_1}_{\op}\le c_3\sigma\sqrt n.
\]
Therefore, on the
same high probability event, we have
\[
\norm{ES_0}_{\op}
\le
c_2\sigma\sqrt{p+m}
+
c_3\sigma\sqrt{\frac mn}\sqrt n
\le
c_4\sigma\sqrt{p+m}.
\]

The lower bound for $\E[\norm E_{\op}]$ follows from standard results on asymptotic behavior of norm of gaussian random matrices (see, e.g.,  \cite{ BaiYin1993,YinBaiKrishnaiah1988,Geman1980}). 
\end{proof}

\section{Detailed proofs in section \ref{sec:errorfluctuation}}

\begin{proof}[Proof of Lemma \ref{wb:lem:signal-reduction}]
Set
\[
P_c:=P-\frac1n(P\one_{[n]})\one_{[n]}^\top.
\]
Since \(\hat L\one_{[n]}=L\one_{[n]}=0\), their principal square roots
satisfy
\[
\hat S\one_{[n]}=0,
\quad
S_0\one_{[n]}=0.
\]
which imply
\[
P(\hat S-S_0)=P_c(\hat S-S_0).
\]
By Lemma~\ref{wb:lem:ideal-target}, we have
\[
PS_0=P_c,
\quad
P\frac Ln=P_c.
\]
Because \(S_0^2=L/n\), these identities imply
\[
P_cS_0=(PS_0)S_0=P\frac Ln=P_c
\]
and hence \(P_cS_0^2=P_c\).  Therefore, we obtain
\[
\begin{aligned}
P_c(\hat S-S_0)(\hat S+I_n)
&=P_c\hat S^2+P_c\hat S-P_cS_0\hat S-P_cS_0\\
&=P_c\hat S^2-P_c\\
&=\frac1nP_c(\hat L-L).
\end{aligned}
\]
where we use $P_c\frac L n=P_c$ from Lemma~\ref{wb:lem:ideal-target}

The symmetric matrix \(\hat L-L\) has zero column sums, so
\[
P_c(\hat L-L)=P(\hat L-L).
\]
Finally, \(\hat S\succeq0\) implies
\(\hat S+I_n\succeq I_n\), and therefore
\[
\norm{(\hat S+I_n)^{-1}}_{\op}\le1.
\]
Right multiplication by \((\hat S+I_n)^{-1}\) proves
\eqref{wb:eq:signal-reduction-identity}; taking operator norms gives
\eqref{wb:eq:signal-reduction}.
\end{proof}

\begin{proof}[Proof of Lemma \ref{wb:lem:distance-cross-signal}]
For convenience, we introduce the short notation
\[u_0:=2\sigma^2p+\Delta^2/2.\]
It suffices to prove the results when
$\kappa_\Delta\ge c_1$, where $c_1\ge1$ is a positive absolute
constant.

For distinct \(i,j\in[n]\), define the miss indicator
\[
U_{ij}
:=
\mathbbm{1}_{\{\norm{X_i-X_j}_2^2<u_0\}}
=1-\hat A_{ij}.
\]
Let \(M\in\{0,1\}^{n\times n}\) be the symmetric zero-diagonal matrix
defined by
\[
M_{ij}
:=
U_{ij}\mathbbm{1}_{\{z_i^*\ne z_j^*\}},
\quad \text{for } i\ne j,
\]
and let \(F\in\{0,1\}^{n\times n}\) be the symmetric zero-diagonal
matrix defined by
\[
F_{ij}
:=
\hat A_{ij}\mathbbm{1}_{\{z_i^*=z_j^*\}},
\quad  \text{for }  i\ne j.
\]
On a cross-cluster pair, i.e., $z_i^* \ne z_j^*$, we have \(A_{ij}=1\) and
\(\hat A_{ij}=1-U_{ij}=1-M_{ij}\). And on a same-cluster pair, i.e., $z_i^* =z_j^*$, we have \(A_{ij}=0\) and $\hat A_{ij}=F_{ij}$.
Therefore, we have
\[
\hat A-A=F-M.
\]
For every symmetric zero-diagonal matrix \(H\), by Lemma \ref{wb:lem:pair-matrix-expansion}, we have
\[
P\mathcal L(H)
=
\sum_{1\le i<j\le n}
H_{ij}
\bigl(\theta_{z_i^*}^*-\theta_{z_j^*}^*\bigr)
(e_i-e_j)^\top.
\]
Every summand supported on a same-cluster pair is zero.  Since \(F\)
is supported on such pairs, we have an important observation that
\[
P\mathcal L(F)=0.
\]
By linearity of \(\mathcal L\), we have
\begin{equation}\label{wb:eq:unified-cross-identity}
P(\hat L-L)=P\mathcal L(\hat A-A)=P\mathcal L(F-M)=-P\mathcal L(M).
\end{equation}

On a cross-cluster pair, i.e., $z_i^* \ne z_j^*$, using Lemma~\ref{wb:lem:cross-monotonicity}, we have
\begin{equation}\label{wb:eq:miss-prob-distance}
\Pb(U_{ij}=1)
\le
\exp\left(-c_2\kappa\left(
\norm{\theta_{z_i^*}^*-\theta_{z_j^*}^*}_2
\right)\right),
\quad \text{where }
\kappa(d):=
\frac{d^4/\sigma^4}{p+d^2/\sigma^2}.
\end{equation}

Relabel the clusters, if necessary, so that $n_1=n_{\max}$, and let
\[
S:=[n]\setminus B_1,
\quad
|S|=m.
\]
Also let
\[
\mathcal I
:=
\bigl\{\{i,j\}:1\le i<j\le n,\ z_i^*\ne z_j^*\bigr\}.
\]
By Lemma~\ref{wb:lem:disjoint-pair-partition}, applied with
\(V=[n]\), we can construct a partition
\[
\mathcal I
=
\bigsqcup_{a=1}^{q_n}\mathcal I_a,
\quad
q_n\le n,
\quad
|\mathcal I_a|\le m,
\]
such that distinct pairs in the same class have disjoint indices.
Accordingly, for each \(a\), the family
\[
\{U_{ij}:\{i,j\}\in\mathcal I_a\}
\]
is independent: every indicator is a measurable function of
\((X_i,X_j)\), and the sample vectors are independent.

For every pair $\{i,j\}\in\mathcal I$, we write
\[
d_{ij}:=\norm{\theta_{z_i^*}^*-\theta_{z_j^*}^*}_2.
\]
For each $a\in[q_n]$, define
\[
T_a:=\sum_{\{i,j\}\in\mathcal I_a}
U_{ij}(\theta_{z_i^*}^*-\theta_{z_j^*}^*)(e_i-e_j)^\top,
\quad
G_a:=\sum_{\{i,j\}\in\mathcal I_a}
U_{ij}(\theta_{z_i^*}^*-\theta_{z_j^*}^*)(\theta_{z_i^*}^*-\theta_{z_j^*}^*)^\top.
\]
By \eqref{wb:eq:unified-cross-identity}, we have
\[
P(\hat L-L)=-\sum_{a=1}^{q_n}T_a.
\]
Distinct pairs in $\mathcal I_a$ have disjoint indices, so the corresponding vectors
$e_i-e_j$ are orthogonal. Also each of these vectors has squared norm $2$.  Since
$U_{ij}^2=U_{ij}$, we obtain the exact identity
\begin{equation}\label{wb:eq:matching-gram-identity}
T_aT_a^\top=2G_a,
\quad
\norm{T_a}_{\op}^2=2\norm{G_a}_{\op}.
\end{equation}
Therefore, it suffices to control
\[G_a=\sum_{\{i,j\}\in\mathcal I_a}
U_{ij}(\theta_{z_i^*}^*-\theta_{z_j^*}^*)(\theta_{z_i^*}^*-\theta_{z_j^*}^*)^\top\]

For convenience, let
\[
R_\Delta:=\frac{2\Delta^2}{c_2\kappa_\Delta}.
\]
By Lemma~\ref{wb:lem:distance-calculus}, we have
\[
\kappa(d_{ij})\ge
\frac{\kappa_\Delta}{\Delta^2}d_{ij}^2.
\]
Combining this with \eqref{wb:eq:miss-prob-distance}, we obtain
\begin{equation}\label{wb:eq:gram-probability-split}
(\E U_{ij})
\le e^{-c_2\kappa(d_{ij})}
\le
e^{-c_2\kappa(d_{ij})/2}e^{-d_{ij}^2/R_\Delta}.
\end{equation}
For a fixed nonempty block $\mathcal I_a$, define the independent,
mean-zero symmetric matrices
\[
Y_{ij}:=(U_{ij}-(\E U_{ij}))(\theta_{z_i^*}^*-\theta_{z_j^*}^*)(\theta_{z_i^*}^*-\theta_{z_j^*}^*)^\top,
\quad \{i,j\}\in\mathcal I_a,
\]
and the deterministic positive-semidefinite matrices
\[
V_{ij}:=
\left(R_\Delta e^{-c_2\kappa(d_{ij})/2}
(\theta_{z_i^*}^*-\theta_{z_j^*}^*)(\theta_{z_i^*}^*-\theta_{z_j^*}^*)^\top\right)^{1/2}.
\]
We check the moment condition of Lemma~\ref{wb:thm:matrix-bernstein}.
For every integer $s\ge2$, we have
\[
\E|U_{ij}-(\E U_{ij})|^s
\le \E|U_{ij}-(\E U_{ij})|^2
=(\E U_{ij})(1-(\E U_{ij}))\le(\E U_{ij}),
\]
and
\[
((\theta_{z_i^*}^*-\theta_{z_j^*}^*)(\theta_{z_i^*}^*-\theta_{z_j^*}^*)^\top)^s
=d_{ij}^{2s-2}(\theta_{z_i^*}^*-\theta_{z_j^*}^*)(\theta_{z_i^*}^*-\theta_{z_j^*}^*)^\top.
\]
For $x\ge0$, the exponential series gives
$x^{s-1}e^{-x}\le(s-1)!$.  Therefore,
\eqref{wb:eq:gram-probability-split} implies
\begin{align*}
\E[Y_{ij}^s]
&\preceq (\E U_{ij})d_{ij}^{2s-2}(\theta_{z_i^*}^*-\theta_{z_j^*}^*)(\theta_{z_i^*}^*-\theta_{z_j^*}^*)^\top\\
&\preceq (s-1)!R_\Delta^{s-1}
e^{-c_2\kappa(d_{ij})/2}(\theta_{z_i^*}^*-\theta_{z_j^*}^*)(\theta_{z_i^*}^*-\theta_{z_j^*}^*)^\top\\
&\preceq \frac{s!}{2}R_\Delta^{s-2}V_{ij}^2.
\end{align*}
Here the first inequality also holds for odd $s$, because the scalar
moment is bounded above by the corresponding absolute moment and
$(\theta_{z_i^*}^*-\theta_{z_j^*}^*)(\theta_{z_i^*}^*-\theta_{z_j^*}^*)^\top\succeq0$.

Write
\[
\Gamma_a:=\sum_{\{i,j\}\in\mathcal I_a}
e^{-c_2\kappa(d_{ij})/2}(\theta_{z_i^*}^*-\theta_{z_j^*}^*)(\theta_{z_i^*}^*-\theta_{z_j^*}^*)^\top \text{ and } \gamma_a:=\norm{\Gamma_a}_{\op}.
\]
The variance parameter in Lemma~\ref{wb:thm:matrix-bernstein} is
\[
\left\|\sum_{\{i,j\}\in\mathcal I_a}V_{ij}^2\right\|_{\op}
=R_\Delta\gamma_a.
\]
Also, by \eqref{wb:eq:miss-prob-distance}, we have
\[
0\preceq\E G_a
=\sum_{\{i,j\}\in\mathcal I_a}(\E U_{ij})(\theta_{z_i^*}^*-\theta_{z_j^*}^*)(\theta_{z_i^*}^*-\theta_{z_j^*}^*)^\top
\preceq\Gamma_a.
\]
Let $t>0$.  Apply \eqref{wb:eq:matrix-bernstein-confidence} in dimension $p$.
With probability at least $1-pe^{-t}$, we have
\[
\lambda_{\max}(G_a-\E G_a)
\le\sqrt{2R_\Delta\gamma_a t}+2R_\Delta t.
\]
Since $G_a\succeq0$, on this event we obtain
\begin{align*}
\norm{G_a}_{\op}
&\le\gamma_a+\sqrt{2R_\Delta\gamma_a t}+2R_\Delta t\\
&\le2\gamma_a+3R_\Delta t.
\end{align*}
For an empty block, the last bound holds deterministically with
$G_a=0$ and $\gamma_a=0$.

Since $\kappa_\Delta\ge1$, by Lemma~\ref{wb:lem:distance-calculus}, we have
\[
d_{ij}^2e^{-c_2\kappa(d_{ij})/2}
\le\frac{4}{c_2e}\Delta^2e^{-c_2\kappa_\Delta/4}.
\]
Therefore, we have
\begin{align*}
\gamma_a
&\le\sum_{\{i,j\}\in\mathcal I_a}
d_{ij}^2e^{-c_2\kappa(d_{ij})/2}\\
&\le c_3m\Delta^2e^{-c_4\kappa_\Delta},
\end{align*}
where $c_3,c_4>0$ are absolute constants.  Combining these estimates
with \eqref{wb:eq:matching-gram-identity}, we conclude that, with
probability at least $1-pe^{-t}$, we have
\begin{equation}\label{wb:eq:matching-gram-bound}
\norm{T_a}_{\op}^2
\le c_5\Delta^2
\left[m e^{-c_4\kappa_\Delta}
+\frac{t}{\kappa_\Delta}\right].
\end{equation}
The scale $R_\Delta$ entered through the verified moment condition;
no upper bound on the distances $d_{ij}$ is required.

Fix $t>0$ and apply \eqref{wb:eq:matching-gram-bound} with
$t+\log(np)$ in place of $t$.  By union bound over the at most $n$
blocks, we conclude that the total failure probability is at most
\[
q_n p e^{-t-\log(np)}\le e^{-t}.
\]
In other words, with probability at least $1-e^{-t}$, the bound holds
simultaneously for every block.  On this event, by triangle inequality and $q_n\le n$, we have
\[
\frac1n\norm{P(\hat L-L)}_{\op}
\le\frac1n\sum_{a=1}^{q_n}\norm{T_a}_{\op}
\le c_6\Delta
\sqrt{m e^{-c_4\kappa_\Delta}
+\frac{t+\log(np)}{\kappa_\Delta}}.
\]
This proves \eqref{wb:eq:matching-signal-confidence}.  Taking $t=p+m$ and using
$\log p\le p$, we have
\[
p+m+\log(np)\le2(p+m+\log n).
\]
Therefore, with probability at least $1-e^{-(p+m)}$, we have
\[
\frac1n\norm{P(\hat L-L)}_{\op}
\le c_7\Delta
\sqrt{m e^{-c_4\kappa_\Delta}
+\frac{p+m+\log n}{\kappa_\Delta}}.
\]
The desired result \eqref{wb:eq:distance-cross-signal} follows from the identity
\[
\frac1{\kappa_\Delta}
=\frac{\sigma^2}{\Delta^2}+\frac{p\sigma^4}{\Delta^4}.
\]
\end{proof}

\begin{proof}[Proof of Proposition \ref{wb:thm:clean-cap}]
Write
\[
G_i:=\frac{\varepsilon_i}{\sigma},
\quad
Z_{ij}:=\frac{G_i-G_j}{\sqrt2},
\quad 1\le i<j\le n,
\]
so that $G_1,\ldots,G_n$ are independent $\mathcal N(0,I_p)$ vectors
and each $Z_{ij}$ has distribution $\mathcal N(0,I_p)$.  Set
\[
\kappa_\Delta:=\frac{\Delta^4/\sigma^4}{p+\Delta^2/\sigma^2},
\quad
\eta:=e^{-\kappa_\Delta/128}.
\]
By Lemma~\ref{wb:lem:pair-matrix-expansion}, we have
\begin{equation}\label{wb:eq:matching-noise-pair-expansion}
H:=E\frac{\hat L}{n}E^\top
=\frac{2\sigma^2}{n}
\sum_{1\le i<j\le n}\hat A_{ij}Z_{ij}Z_{ij}^\top.
\end{equation}

Clearly $\norm{E \hat{S}}_\op^2 = \norm{H}_\op$. Relabel the clusters according to \eqref{wb:eq:onebig-convention}, so
that $B_1$ is a largest cluster and $|[n]\setminus B_1|=m$.
Lemma~\ref{wb:lem:disjoint-pair-partition} partitions all unordered
pairs into $q_n\le n$ blocks with disjoint indices within each block.
Denote the blocks as $\mathcal{C}_1, \ldots, \mathcal{C}_{q_n}$. For each $a \le q_n$, we have
\begin{equation}\label{eq:block_bound}
    |\mathcal{C}_a| \le \frac{n}{2}, \quad |\{\{i,j\} \in \mathcal{C}_a: z_{i}^* \ne z_{j}^*\}| \le m.
\end{equation}

Note that if $z_i^*=z_j^*$, then $X_i-X_j=\sigma(G_i-G_j)$, and therefore we have
\[
\Pb(\hat A_{ij}=1)
=\Pb\left(\chi_p^2\ge p+\frac{\Delta^2}{4\sigma^2}\right).
\]
The moment generating function of the $\chi^2_p$ distribution is given by
$M_{\chi^2_p} (\lambda) =(1-2\lambda)^{-p/2}$, for
$0<\lambda<1/2$, and by Markov's inequality we have
\[
\Pb(\hat A_{ij}=1)
\le
\exp\left(-\frac12\left(
\frac{\Delta^2}{4\sigma^2}
-p\log\left(1+\frac{\Delta^2}{4\sigma^2p}\right)
\right)\right).
\]
Using the elementary inequality
\[
s-\log(1+s)
=\int_0^s\frac{x}{1+x}\,dx
\ge\frac{s^2}{2(1+s)},
\quad s\ge0,
\]
we obtain
\[
\Pb(\hat A_{ij}=1)
\le
\exp\left(-\frac{\Delta^4/\sigma^4}
{64(p+\Delta^2/(4\sigma^2))}\right)
\le e^{-\kappa_\Delta/64}
=\eta^2.
\]

Fix $v\in S^{p-1}$ and define
\[
Y_{ij}(v):=\hat A_{ij}\ip{Z_{ij},v}^2,
\quad
w_{ij}:=
\begin{cases}
1,&z_i^*\ne z_j^*,\\
\eta,&z_i^*=z_j^*.
\end{cases}
\]
Clearly, for any $i, j \in [n]$, $\bbP(\hat{A}_{ij} = 1) \le w_{ij}^2$. For $Z\sim\mathcal N(0,1)$ and every integer $q\ge1$, note that
\[
\E Z^{4q}
=\prod_{\ell=1}^q(4\ell-3)(4\ell-1)
\le16^q(q!)^2.
\]
Since $\hat A_{ij}$ is an indicator, for $q\ge1$, by Cauchy--Schwarz we have
\begin{equation}\label{wb:eq:matching-noise-moment}
\begin{aligned}
\E[Y_{ij}(v)^q]
&\le
\Pb(\hat A_{ij}=1)^{1/2}
\left(\E\ip{Z_{ij},v}^{4q}\right)^{1/2}\\
&\le q!4^q w_{ij}.
\end{aligned}
\end{equation}

For each $a\in[n]$, let
\[
S_a(v):=\sum_{\{i,j\}\in\mathcal C_a}Y_{ij}(v),
\quad
W_a:=\sum_{\{i,j\}\in\mathcal C_a}w_{ij}.
\]
From \eqref{eq:block_bound}, we get the trivial bound
\[
W_a\le m+\frac n2\eta.
\]
Each $Y_{ij}(v)$ is a measurable function of $(G_i,G_j)$ alone. Therefore, the disjoint-index property implies independence of the
summands in $S_a(v)$. Thus, for $a \le q_n$, by
\eqref{wb:eq:matching-noise-moment}, we have
\[
\E S_a(v)\le4W_a,
\quad
\sum_{\{i,j\}\in\mathcal C_a}\E[Y_{ij}(v)^2]
\le32W_a=(32W_a),
\]
and, for every integer $q\ge3$, we have
\[
\sum_{\{i,j\}\in\mathcal C_a}\E[(Y_{ij}(v))_+^q]
\le q!4^qW_a
=\frac{q!}{2}(32W_a)b^{q-2}.
\]
Now, for each $a \le q_n$, we apply
Lemma~\ref{wb:thm:bernstein} with $b=4$ and variance parameter
$32W_a$, and conclude that, for every
$u>0$, with probability at least $1-e^{-u}$, we have
\begin{equation}\label{wb:eq:matching-noise-block-bound}
\begin{aligned}
S_a(v)
&\le \E S_a(v)+\sqrt{2(32W_a)u}+4u\\
&\le4W_a+8\sqrt{W_a u}+4u\\
&\le8(W_a+u)\le8(m+\frac n2\eta+u)
\end{aligned}
\end{equation}
where we used $2\sqrt{W_a u}\le W_a+u$. 

Let $\mathcal N\subset S^{p-1}$ be the $1/4$-net from
Lemma~\ref{wb:lem:net}, so that $|\mathcal N|\le9^p$.  Fix $t>0$
and choose
\[
u:=t+p\log9+\log n.
\]
Applying \eqref{wb:eq:matching-noise-block-bound} for every
$a \le q_n$ and $v\in\mathcal N$, a union bound gives that 
\[
    S_a(v) \le 8(m+\frac n2\eta+ t+p\log9+\log n) \quad \text{for all}\,\, a \le q_n,
\]
with probability at least
\[
1 - q_n |\mathcal N|e^{-u}\ge 1 - n9^p e^{-u}=1 - e^{-t}.
\]
On this high probability event,
\eqref{wb:eq:matching-noise-pair-expansion} gives, for all
$v\in\mathcal N$,
\[
v^\top Hv
=\frac{2\sigma^2}{n}\sum_{a=1}^n S_a(v)
\le16\sigma^2(m+\frac n2\eta+p\log9+t+\log n).
\]
The matrix $H$ is positive semidefinite by
\eqref{wb:eq:matching-noise-pair-expansion}.  Therefore, by
Lemma~\ref{wb:lem:net} we have
\[
\norm H_{\op}\le2\sup_{v\in\mathcal N}v^\top Hv.
\]
It follows that, for every $t>0$,
\begin{equation}\label{wb:eq:matching-noise-covariance-bound}
\Pb\left(
\norm H_{\op}
>32\sigma^2\left(m+\frac n2e^{-\kappa_\Delta/128}
+p\log9+t+\log n\right)
\right)
\le e^{-t}.
\end{equation}

Choose $c_{\ref{wb:thm:clean-cap}.1}\ge128$.  Since
$ \kappa_{\Delta} > c_{\ref{wb:thm:clean-cap}.1} (1+\log_+(n/(p+m)))$, we have
\[
n e^{-\kappa_\Delta/128}
\le e^{-1}\min\{n,p+m\}\le\frac{p+m}{2}.
\]
Taking $t=p+m$ in
\eqref{wb:eq:matching-noise-covariance-bound} and using
$m+p\log9\le(p+m)\log9$, we conclude that
\[
\norm H_{\op}
\le32\left(\log9+\frac54\right)\sigma^2(p+m+\log n) \le 128 \sigma^2 (p + m + \log n)
\]
with probability at least $1-e^{-(p+m)}$.  The desired result is proved because $\norm{E \hat{S}}_\op = \norm{H}_\op^{1/2}$.
\end{proof}

\end{document}